\pdfoutput=1

\documentclass[10pt,twocolumn,letterpaper]{article}

\usepackage[pagenumbers]{cvpr}

\usepackage{multirow}
\usepackage{pifont}

\usepackage{microtype}

\makeatletter
\renewcommand\paragraph{\@startsection{paragraph}{4}{\z@}{6pt \@plus2pt \@minus2pt}{-1em}{\normalfont\normalsize\bfseries}}
\makeatother

\newcommand{\cisyn}{\textsc{CISyn}}

\newcommand{\Y}{{\color{ForestGreen}\checkmark}}
\newcommand{\y}{{\color{Goldenrod}$\bullet$}}
\newcommand{\N}{{\color{Red}$\times$}}

\newcommand{\maydelete}[1]{}

\definecolor{cvprblue}{rgb}{0.21,0.49,0.74}
\usepackage[pagebackref,breaklinks,colorlinks,allcolors=cvprblue]{hyperref}

\def\paperID{411}
\def\confName{3DV\xspace}
\def\confYear{2027\xspace}

\title{\cisyn{}: Human Mesh Recovery in Close Interactions with \\ Physics-Simulated Synthetic Data}

\author{Yufan Deng$^{1}$ \quad Lorenza Prospero$^{1,2}$ \quad Christian Rupprecht$^{1}$\\
{\fontsize{10.5}{12.5}\selectfont $^{1}$VGG, University of Oxford \quad $^{2}$The Podium Institute for Sports Medicine and Technology, University of Oxford}\\
{\tt\small \{yufan,lorenza,chrisr\}@robots.ox.ac.uk}
}

\begin{document}
\maketitle

\begin{abstract}
  3D human mesh recovery (HMR) in close-interaction scenarios is hindered by the lack of large-scale, accurately annotated training data: real capture is expensive, and existing synthetic datasets rarely include physical contact. We introduce \cisyn{}, a large-scale synthetic dataset purpose-built for close human-human interactions, generated via physics simulation and Blender-based rendering. By simulating collision dynamics between SMPL-X-rigged humanoids, \cisyn{} naturally produces the contact, mutual occlusion, and pose entanglement that characterize close-interaction scenes, with no manual annotation effort. The released set provides 25{,}000 sequences with accurate 3D annotations. We further propose an uncertainty-aware loss that allows models to learn more effectively from the heavily occluded examples in our data. Augmenting state-of-the-art HMR training with \cisyn{} and replacing the regression loss with our uncertainty-aware variant yields consistent improvements on close-interaction benchmarks, validating \cisyn{} as a scalable complement to real-world data. The dataset, generation pipeline, and code are available at \url{https://github.com/yfde/CISyn}.%
  \par
\end{abstract}

\begin{figure}[t]
  \centering
  \includegraphics[width=\linewidth]{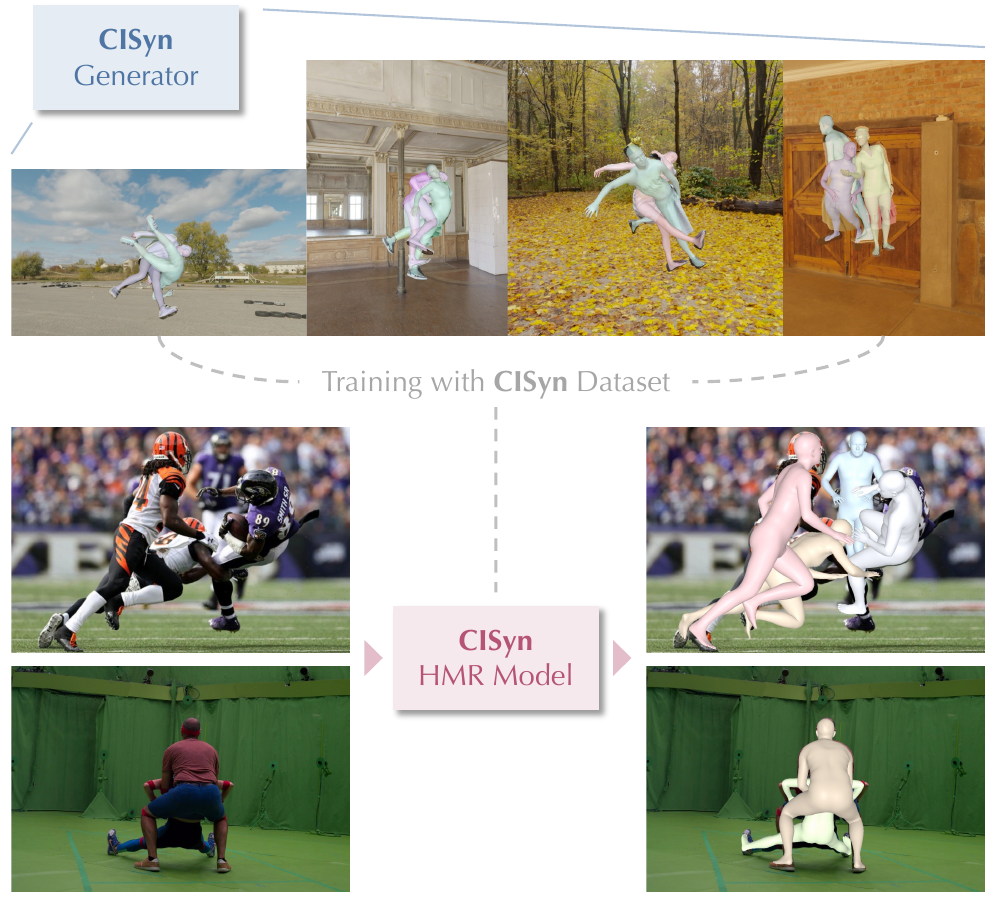}
  \caption{
    \textbf{\cisyn{} overview.}
    We introduce \cisyn{}, a large-scale synthetic dataset of close human-human interactions generated via physics simulation. \emph{Top:} sample rendered frames of the \cisyn{} dataset. \emph{Bottom:} improved human mesh recovery results on close and complex samples.
  }
  \label{fig:teaser}
\end{figure}

\section{Introduction}
\label{sec:intro}

Recovering 3D human pose and shape from a single image is a long-standing problem in computer vision, with applications ranging from sports analysis and rehabilitation to social robotics and embodied AI. While modern human mesh recovery (HMR) methods produce accurate estimates for isolated, well-separated subjects, performance degrades sharply when two or more people interact at close range. Hugging, grappling, and partner dancing introduce mutual occlusion, deeply entangled limbs, and depth ambiguities between bodies that are within centimeters of each other. In this regime, both crop-based and single-stage regressors~\cite{goel20234dhumans,baradel2024multi,su2025sathmr,wang2025mahmr,wang2025prompthmr} still struggle.

A central reason is the lack of suitable training data. Studio-captured close-interaction datasets such as Hi4D~\cite{yin2023hi4d} and CHI3D~\cite{fieraru2025chi3d} provide accurate SMPL(-X)~\cite{2015smpl,pavlakos2019smplx} annotations but are limited in scale and cover only a small set of canonical interactions. In-the-wild pseudo-ground-truth pipelines~\cite{goel20234dhumans,wang2025mahmr,patel2025camerahmr} cover diverse appearances at scale, yet contain few close-contact frames and produce noisy labels where bodies overlap. Synthetic datasets~\cite{patel2021agora,black2023bedlam,yang2023synbody,tesch2025bedlam2} render large numbers of subjects with accurate ground truth, but place each person independently in the scene; the contact, collision response, and pose entanglement that define close interaction are absent by construction.

We argue that physics simulation is a natural way to fill this gap. A rigid-body simulator with collision detection produces contact, mutual occlusion, and physically plausible pose coupling between bodies, with no manual labeling. The resulting dataset includes interactions rarely seen in existing capture datasets, sampling from a wider distribution than the target domain, so models trained on it can generalize better. Building on this, we introduce \cisyn{}, a large-scale synthetic dataset for close human-human interaction HMR. Our pipeline launches pairs of SMPL-X-rigged humanoids toward each other in the Genesis physics engine~\cite{2024genesis}, simulates the high-speed collisions, and renders the contact-rich motion with a Blender-based~\cite{blender} renderer using diverse HDR environments, body assets, and camera placements. The pipeline scales to arbitrarily many sequences and can easily be extended to more than two interactors (Supp.\ Figure~\ref{fig:supp_datasample}).
The released set contains 25{,}000 sequences with accurate SMPL-X~\cite{pavlakos2019smplx}, SMPL~\cite{2015smpl}, and camera annotations.

Even with exact ground truth, supervision for heavily occluded joints is intrinsically ambiguous in a single image. We therefore complement \cisyn{} with an occlusion-weighted, uncertainty-aware Laplacian NLL loss that allows the model to down-weight occluded joints and learn per-joint confidence, which we find important for exploiting the harder examples that \cisyn{} provides.

In summary, our contributions are as follows:
\begin{itemize}
  \item We introduce \cisyn{}, a large-scale synthetic dataset of close human-human interactions generated by physics-simulated collisions between SMPL-X humanoids, providing 25{,}000 rendered sequences with accurate SMPL(-X) and camera annotations.
  \item We release a fully automated generation pipeline that requires no manual interaction or design, and scales to arbitrary numbers of sequences and to more interactors, making physics-based close-interaction data a practical resource for the community.
  \item We propose an occlusion-weighted, uncertainty-aware Laplacian NLL loss that enables a single-stage regressor model to estimate per-joint reliability and show that it is important for extracting the benefit of synthetic close-interaction data.
  \item We conduct comprehensive experiments that benchmark \cisyn{} against existing close-interaction training sets and ablate each component of our framework, validating the effectiveness of both the dataset and the proposed loss.
\end{itemize}

\section{Related Work}
\label{sec:related}

\subsection{Multi-Person HMR Datasets}
\label{sec:related:datasets}
\begin{table}[t]
  \caption{
    \textbf{Comparison with existing multi-person HMR datasets.}
    \emph{\#Seq}/\emph{\#Frm}: number of sequences/frames; \emph{Syn.}: synthetically generated; \emph{SMPL}: SMPL or SMPL-X body annotations; \emph{CC}: close contact; \emph{CP}: complex pose.
    For the \emph{CC} and \emph{CP} columns, \Y\ marks a primary focus of the dataset and \y\ marks incidental coverage.
    \cisyn{} is the only entry that is simultaneously SMPL-X-annotated, close-contact, and complex-pose.
  }
  \label{tab:dataset_comparison}
  \centering
  \small
  \setlength{\tabcolsep}{3pt}
  \begin{tabular}{lcccccc}
    \toprule
    Dataset & \#Seq & \#Frm & Syn. & SMPL & CC & CP \\
    \midrule
    CMU Panoptic~\cite{joo2015panoptic}         & 65    & 1.5M     & \N & \N & \Y & \y \\
    MuPoTS-3D~\cite{mehta2018mupots}            &  20   & 8K     & \N & \N & \Y & \y \\
    ExPI~\cite{guo2022expi}                     &  115   & 30K     & \N & \N & \Y & \Y \\
    3DPW~\cite{von20183dpw}                     &   60  &  51K    & \N & \Y & \y & \y \\
    4D-Humans~\cite{goel20234dhumans}           &  -   &  3.2M    & \N & \Y & \y & \y \\
    CameraHMR~\cite{patel2025camerahmr}         &  -   &  3.2M    & \N & \Y & \y & \y \\
    DTO-Humans~\cite{wang2025mahmr}             &  -   &  560K    & \N & \Y & \y & \y \\
    Embody3D~\cite{mclean2025embody3d}          &  9K   & 54M     & \N & \Y & \y & \y \\
    Hi4D~\cite{yin2023hi4d}                     &  100   & 95K     & \N & \Y & \Y & \y \\
    CHI3D~\cite{fieraru2025chi3d}               &  631   & 259K     & \N & \Y & \Y & \y \\
    Harmony4D~\cite{khirodkar2024harmony4d}     & 208    &  1.66M    & \N & \Y & \Y & \y \\
    AGORA~\cite{patel2021agora}                 &  -   & 17K     & \Y & \Y & \N & \N \\
    BEDLAM~\cite{black2023bedlam}               & 10K & 380K & \Y & \Y & \N & \N \\
    SynBody~\cite{yang2023synbody}              &  27K   &  1.2M    & \Y & \Y & \N & \N \\
    BEDLAM2~\cite{tesch2025bedlam2}             & 27K & 8M   & \Y & \Y & \N & \N \\
    \midrule
    \textbf{\cisyn{} (ours)}                    & 25K & 300K & \Y & \Y & \Y & \Y \\
    \bottomrule
  \end{tabular}
\vspace{-3mm}
\end{table}

\begin{figure*}[t]
  \centering
  \includegraphics[width=0.99\textwidth]{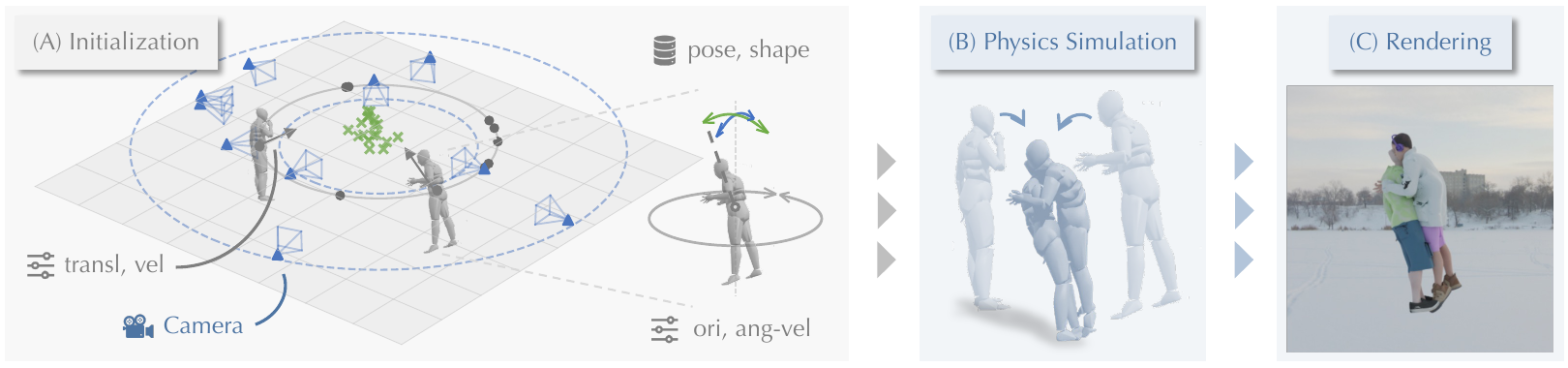}
  \caption{
    \textbf{\cisyn{} generation pipeline.}
    (A)~Randomized initialization sets each actor's body shape and pose (sampled from a database), translation on the gray circle, linear velocity directed at the green~$\times$ collision target near the scene center, mostly upright orientation, and small random angular velocity.
    (B)~The physics simulator advances the state, producing contact-rich pose sequences.
    (C)~The renderer turns the simulated motion into RGB images paired with ground-truth annotations.
  }
  \label{fig:pipeline}
\vspace{-2mm}
\end{figure*}

Human Mesh Recovery (HMR) datasets are obtained in two ways: (1) annotating real images with 3D ground truth, or (2) rendering synthetic images from captured or generated motion. Table~\ref{tab:dataset_comparison} compares existing multi-person HMR datasets.

\paragraph{Real-world datasets.}
Studio-based datasets such as MuPoTS-3D~\cite{mehta2018mupots} and the CMU Panoptic Studio~\cite{joo2015panoptic} obtain accurate skeleton annotations from multi-camera rigs but are restricted to controlled indoor environments. 3DPW~\cite{von20183dpw} extends coverage to outdoor scenes via IMU-assisted annotation and has become a standard in-the-wild benchmark; 
its annotations, however, are optimization-derived rather than marker-based, and are less precise than indoor mocap ground truth.
Recent capture systems~\cite{yin2023hi4d,fieraru2025chi3d,khirodkar2024harmony4d}
provide high-fidelity SMPL(-X) annotations for tightly interacting pairs, yet remain restricted to a small set of canonical interactions such as hugging or partner dancing. ExPI~\cite{guo2022expi} captures extreme acrobatic two-person motion but provides no body-mesh annotations, and Embody3D~\cite{mclean2025embody3d} offers very large-scale capture but does not target close interaction.
When studio capture is too costly, pseudo-ground-truth (pGT) is generated from in-the-wild images~\cite{goel20234dhumans,wang2025mahmr,patel2025camerahmr}. These pGT datasets cover diverse appearances at scale, but their 3D accuracy is bounded by the optimizer used to produce them, and they contain few close-contact frames. 
Overall, real-capture datasets are constrained by acquisition costs and capture setup: studio rigs are expensive to operate, and even markerless pipelines~\cite{mamma} that reduce annotation effort cannot broaden interaction diversity beyond what fits in a studio.

\paragraph{Synthetic HMR datasets.}
Synthetic data has long served as a scalable alternative to real capture for HMR. SURREAL~\cite{varol17surreal} pioneered this direction by rendering SMPL bodies on randomized backgrounds. AGORA~\cite{patel2021agora} raised photorealism by compositing SMPL-X avatars into HDR outdoor scenes with metric-scale 3D annotations, and BEDLAM~\cite{black2023bedlam} subsequently demonstrated that carefully rendered synthetic data \emph{alone}
can match or exceed methods trained on costly real datasets. GTAHuman~\cite{cai2024gtahuman} and SynBody~\cite{yang2023synbody} are other examples of synthetic HMR datasets.
A consistent limitation is that humans are placed in scenes \emph{independently}, with no physical coupling between bodies. Inter-person contact, collision response, and pose entanglement (the defining features of close interaction) are therefore absent. As a result, models trained on existing synthetic corpora are not exposed to the occlusion patterns and body configurations that characterize real hugging, dancing, or contact sport, precisely where real-capture datasets are also most scarce. \cisyn{} addresses this gap by generating interactions through rigid-body collision simulation, so that contact, mutual occlusion, and pose coupling emerge organically from the dynamics of interacting bodies.

\subsection{Multi-Person Human Mesh Recovery}
\label{sec:related:hmr}
Human mesh recovery (HMR) takes a single image as input and produces a fitted SMPL~\cite{2015smpl} or SMPL-X~\cite{pavlakos2019smplx} parametric body model. Early approaches solve this iteratively with rendering and optimization in the loop~\cite{bogo2016smplify,kolotouros2019spin}, while modern HMR methods regress body parameters in a single feed-forward pass.
Multi-person HMR follows two paradigms. \emph{Crop-based} methods first detect each person and then apply a single-person regressor to the crop~\cite{kanazawa2018hmr,li2022cliff,goel20234dhumans,cai2023smpler,yin2025smplest,patel2025camerahmr}; they yield accurate per-person estimates but discard context that can be useful to resolve occlusion and inter-person interaction. SAM 3D Body~\cite{yang2026sam3dbody} scales this paradigm up, but predicts MHR rather than SMPL meshes and trains on Harmony4D, placing it outside our SMPL-based cross-dataset protocol.
\emph{Single-stage} methods~\cite{jiang2020crmh,sun2021romp,sun2022bev,baradel2024multi,su2025sathmr,wang2025mahmr,wang2025prompthmr} instead process the entire image to jointly detect and reconstruct all persons, naturally capturing the scene-level cues that help under occlusion and contact.
Diffusion-based methods~\cite{cho2023diffhmr,muller2024buddi, xia2026socialmirror} recover bodies by denoising SMPL(-X) parameters conditioned on the image; their probabilistic formulation provides a useful prior under heavy occlusion but typically trails regression-based methods in accuracy.
Despite this progress, the close-interaction regime, where occlusion is severe and poses are deeply entangled, remains an open challenge. We propose an uncertainty-aware loss modification that enables regression-based methods to better handle occlusion in close-interaction scenarios.

\subsection{Human Motion Synthesis}
\label{sec:related:motion}
Beyond direct capture, prior work synthesizes human motion to expand training data along two complementary lines. \emph{Generative motion models}, including diffusion-based single-person motion generators~\cite{tevet2022mdm, chi2024m2d2m,shafir2023priormd}
and recent two-person extensions~\cite{liang2024interhuman,fang2024dualhuman}%
, produce diverse motion sequences but inherit the coverage bias of the interaction datasets they are trained on, and rarely produce the highly entangled contact configurations that arise in genuine close interaction.
\emph{Physics-based motion synthesis} produces physically plausible motion via RL controllers that track reference motions in a simulator~\cite{yuan2021simpoe,Luo2023PerpetualHC,Luo2022EmbodiedSH,ugrinovic2024multiphys}, environment-contact models~\cite{hassan2021samp,zhang2022couch,shimada2020physcap}, or physics-guided diffusion~\cite{yuan2023physdiff}.
Even multi-person variants still follow reference motions, using the simulator to prevent interpenetration rather than to originate close-contact interactions.
\cisyn{} instead drives SMPL-X humanoids directly via rigid-body collision dynamics, generating contact-rich interaction sequences without pre-existing interaction references.

\section{Method}
\label{sec:method}
\begin{figure*}[t]
  \centering
  \begin{subfigure}[b]{0.21\linewidth}
    \begin{minipage}[c]{0.08\linewidth}
      \centering\rotatebox{90}{\small Count}
    \end{minipage}%
    \begin{minipage}[c]{0.92\linewidth}
      \includegraphics[width=\linewidth]{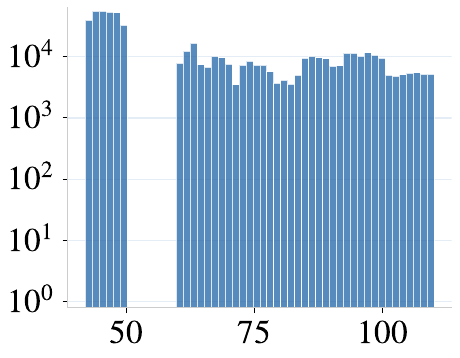}
    \end{minipage}
    \caption{FoV ($^\circ$)}
    \label{fig:diversity:fov}
  \end{subfigure}
  \hfill
  \begin{subfigure}[b]{0.21\linewidth}
    \begin{minipage}[c]{0.92\linewidth}
      \includegraphics[width=\linewidth]{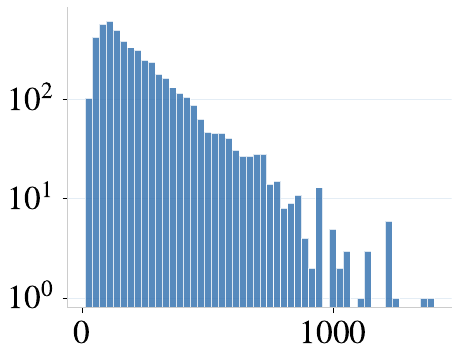}
    \end{minipage}
    \caption{bbox area ($10^3$ px$^2$)}
    \label{fig:diversity:bbox}
  \end{subfigure}
  \hfill
  \begin{subfigure}[b]{0.56\linewidth}
    \includegraphics[width=\linewidth]{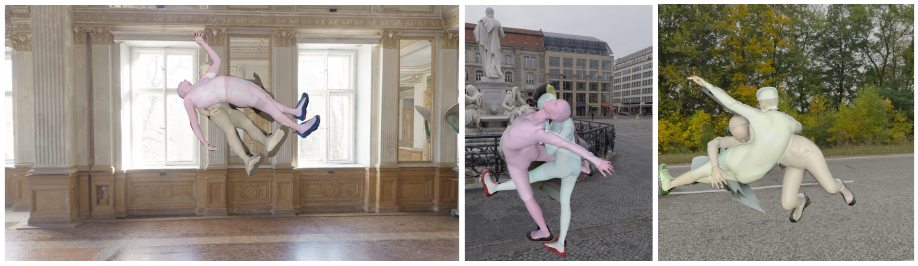}
    \caption{Sample frames.}
    \label{fig:diversity:grid}
  \end{subfigure}
  \caption{
    \textbf{\cisyn{} dataset diversity.}
    (a)~Distribution of camera field of view (FoV) across rendered frames.
    (b)~Distribution of per-person bounding-box area.
    (c)~Sample frames.
  }
  \label{fig:diversity}
\vspace{-2mm}
\end{figure*}

We address human mesh recovery in close interactions along two complementary axes: data and training objective. On the data side, we introduce the \cisyn{} Dataset, a large-scale synthetic dataset for close-interaction HMR (Section~\ref{sec:method:dataset}). On the objective side, we propose an uncertainty-aware loss function tailored to this regime (Section~\ref{sec:method:model}).

\subsection{The \cisyn{} Dataset}
\label{sec:method:dataset}

Figure~\ref{fig:pipeline} illustrates the \cisyn{} generation pipeline, which consists of two stages: (1) physics simulation of high-speed collisions between humans to obtain close-contact, physically plausible interaction pose sequences, and (2) Blender-based~\cite{blender} rendering of the simulated motion into RGB images paired with ground-truth annotations.

\subsubsection{Human Motion Simulation}
\label{sec:method:dataset:simulate}
After randomly sampling the initial state of the humans, we simulate motion using the Genesis~\cite{2024genesis} physics engine with standard gravity and rigid-body collision detection. The simulator runs at 480 steps per second; we export 12 frames per sequence at 30\,FPS, yielding 0.4\,s sequences centered on the collision window (Supp. Sec.~\ref{app:pipeline:sim}).

\paragraph{SMPL-X Simulator Model.}
We adopt the approximated SMPL-X simulator model from~\cite{Luo2023PerpetualHC}.
This conversion makes SMPL-X compatible with rigid-body physics engines such as MuJoCo~\cite{todorov2012mujoco} and Genesis~\cite{2024genesis} and enables more efficient simulation.
We additionally specify per-joint angle limits and disable self-collision for body-part pairs that would otherwise produce artifacts, detailed in Supp. Sec.~\ref{app:pipeline:sim:model}. The body density is set to $1000\,\mathrm{kg/m^3}$, consistent with the physiological range of $900$–$1100\,\mathrm{kg/m^3}$ observed across most human tissues.

To maximize diversity while preserving physical plausibility, we randomize six aspects of each simulated sequence:

\begin{itemize}
  \item \textbf{Body shape} $\boldsymbol{\beta} \in \mathbb{R}^{10}$: each actor is assigned a distinct shape sampled without replacement from a pool of 1100 simulator-compatible SMPL-X assets, matched to the dressed rendering assets in Section~\ref{sec:method:dataset:render:assets}.
  \item \textbf{Initial pose} $\boldsymbol{\theta} \in \mathbb{R}^{55 \times 3}$: each actor is assigned an initial body pose sampled without replacement from the AMASS library~\cite{mahmood2019amass}.
  \item \textbf{Initial translation} $\boldsymbol{t} \in \mathbb{R}^3$: actors are placed on a circle around the scene origin at uniformly drawn angular positions, subject to a minimum pairwise separation constraint.
  \item \textbf{Initial orientation} $\boldsymbol{\theta}_0 \in \mathbb{R}^3$: yaw is sampled uniformly over the full range; pitch and roll use small zero-mean Gaussian noise to keep poses roughly upright.
  \item \textbf{Linear velocity} $\boldsymbol{v} \in \mathbb{R}^3$: each actor is launched toward the scene center with a small directional perturbation; horizontal and vertical speed magnitudes are independently sampled from bounded ranges.
  \item \textbf{Angular velocity} $\boldsymbol{\omega} \in \mathbb{R}^3$: initial spin is sampled uniformly per axis.
\end{itemize}

We generate sequences under two parameter regimes to cover a broader range of contact dynamics: a \emph{standard} batch and a \emph{high-speed} batch with higher linear and angular velocities that produce more forceful collisions. Full parameter ranges are listed in Supp. Sec.~\ref{app:pipeline:sim:params}.
\subsubsection{Human Motion Rendering}
\label{sec:method:dataset:rendering}
We render the simulated motion into RGB images using XRFeitoria~\cite{2023xrfeitoria}, a Blender-based~\cite{blender}
rendering engine previously used for the SynBody~\cite{yang2023synbody} HMR dataset.

\paragraph{SMPL-X assets.}
\label{sec:method:dataset:render:assets}
We employ 1100 SMPL-X assets dressed with diverse clothing from SynBody~\cite{yang2023synbody}. Each asset is provided as a pre-skinned FBX file; XRFeitoria drives it with standard linear blend skinning (LBS) to reproduce the per-frame joint rotations exported from the physics simulation. The body shape parameters $\boldsymbol{\beta}$ of these 1100 assets also seed the matched simulator-compatible models described in Section~\ref{sec:method:dataset:simulate}, ensuring the rendered appearance is geometrically consistent with the simulated motion.

\begin{figure*}[t]
  \centering
  \includegraphics[width=0.99\textwidth]{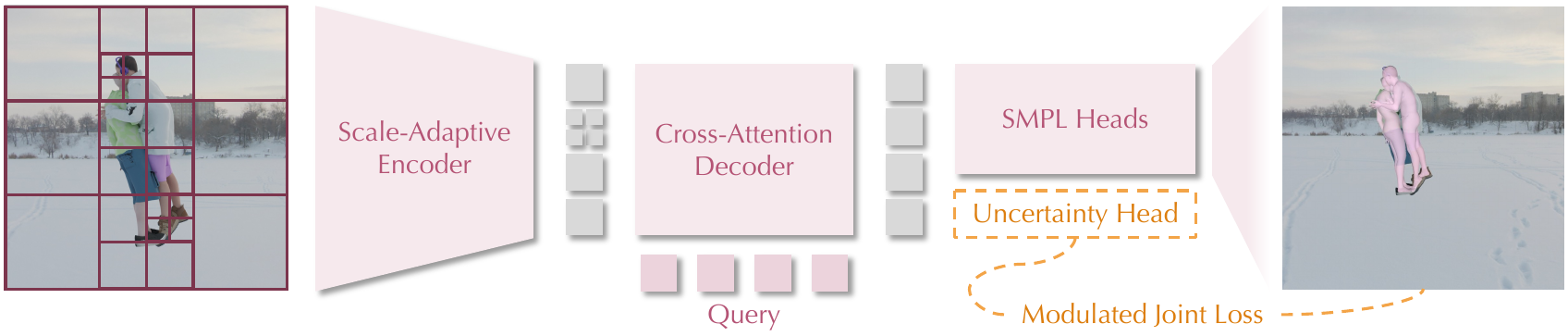}
  \caption{
    \textbf{\cisyn{}-HMR architecture.}
    Our model builds on MA-HMR~\cite{wang2025mahmr}: a DINOv2~\cite{oquab2023dinov2} ViT encoder with scale-adaptive tokens (SAT) feeds a DAB-DETR~\cite{liu2022dabdetr} decoder that predicts SMPL parameters per query. Beyond the standard heads, we add an uncertainty head that predicts a per-joint $\sigma$, used in the modified 3D/2D joint losses shown with dashed lines.
  }
  \label{fig:model}
\vspace{-3mm}
\end{figure*}

We randomize two aspects of the rendering to improve appearance diversity:

\begin{itemize}
  \item \textbf{Randomized Lighting and Background}: each sequence is lit by a randomly sampled HDR environment map used as image-based lighting (IBL). Outdoor maps are sourced from the BEDLAM2~\cite{tesch2025bedlam2} HDR collection; indoor maps from Poly Haven~\cite{polyhaven}.
  \item \textbf{Randomized Camera}: the camera is placed at a distance of $2$–$5$\,m from the interaction center at a height of ca.~$1.4$\,m, with FoV ranging from $42^\circ$ to $110^\circ$ across batches, which is shown in Figure~\ref{fig:diversity:fov}. We use a single static camera for each sequence. Three resolution formats are used: $1280\times720$ (landscape), $940\times1280$ (portrait), and $1280\times1280$ (square).
\end{itemize}

\subsubsection{Annotations}
\label{sec:method:dataset:annotations}
Each image is annotated with SMPL-X~\cite{pavlakos2019smplx} parameters (pose $\boldsymbol{\theta}\in \mathbb{R}^{55 \times 3}$, shape $\boldsymbol{\beta}\in \mathbb{R}^{10}$, translation $\boldsymbol{t}\in \mathbb{R}^3$), as well as SMPL~\cite{2015smpl} parameters ($\boldsymbol{\theta}\in \mathbb{R}^{24 \times 3}$, $\boldsymbol{\beta}\in \mathbb{R}^{10}$, $\boldsymbol{t}\in \mathbb{R}^3$) obtained via the official SMPL-X-to-SMPL optimization-based conversion~\cite{pavlakos2019smplx}. We also provide camera intrinsics and extrinsics to place bodies in camera space. The renderer additionally produces depth maps and segmentation masks, which we release but do not use for training.

\subsubsection{Dataset Statistics}
\label{sec:method:dataset:stats}

We set the number of humans per scene to 2 and use the pipeline to generate \cisyn{}, which contains 24{,}000 training sequences and 1{,}000 test sequences, each 12 frames long, for a total of 300{,}000 frames. Figure~\ref{fig:diversity} shows the distribution of camera field of view (FoV) and per-person bounding-box area computed from the 3D annotations, along with sample frames.
Supp. Sec.~\ref{app:dataset_analysis} compares the pose distribution of \cisyn{} against real close-interaction captures and measures inter-person mesh penetration.
The pipeline is computationally lightweight and scales easily: sequences are independent and can be parallelized across GPUs, and extending to more interacting people requires minimal effort (Supp. Sec.~\ref{app:pipeline:cost}).

\subsection{Uncertainty-aware HMR Model: \cisyn-HMR}
\label{sec:method:model}
\begin{table*}[t]
  \caption{
    \textbf{3D human mesh recovery on close-interaction benchmarks.}
    \cisyn{}-HMR vs.\ state-of-the-art monocular multi-person methods; all metrics in mm.
    First block: PromptHMR numbers from its paper, the rest from released checkpoints.
    The last two blocks are trained by us under an identical schedule and differ only in whether \cisyn{} is added to Base, the original MA-HMR~\cite{wang2025mahmr} training mix (AGORA~\cite{patel2021agora} + BEDLAM~\cite{black2023bedlam} + DTO-Humans~\cite{wang2025mahmr}); for a fair comparison, MA-HMR~(Base) continues training the released checkpoint on Base for the same number of epochs, by which point all four models have converged.
    No model is trained or fine-tuned on any benchmark dataset.
  }
  \label{tab:main_results}
  \centering
  \small
  \setlength{\tabcolsep}{3.5pt}
  \begin{tabular}{lccccccccc}
    \toprule
    \multirow{2}{*}{Method} & \multicolumn{3}{c}{Hi4D} & \multicolumn{3}{c}{CHI3D} & \multicolumn{3}{c}{Harmony4D} \\
    \cmidrule(lr){2-4} \cmidrule(lr){5-7} \cmidrule(lr){8-10}
    & MPJPE $\downarrow$ & PA-MPJPE $\downarrow$ & PVE $\downarrow$ & MPJPE $\downarrow$ & PA-MPJPE $\downarrow$ & PVE $\downarrow$ & MPJPE $\downarrow$ & PA-MPJPE $\downarrow$ & PVE $\downarrow$ \\
    \midrule
    Multi-HMR~\cite{baradel2024multi} & 86.7 & 52.8 & 106.2 & 63.0 & 35.7 & 76.2 & 117.9 & 65.2 & 135.3 \\
    PromptHMR~\cite{wang2025prompthmr} & 63.9 & \textbf{39.2} & -- & 48.0 & \textbf{27.2} & -- & -- & -- & -- \\
    MA-HMR~\cite{wang2025mahmr}       & 61.2 & 45.9 & 77.2 & 47.2 & 31.4 & 55.7 & 77.5 & 51.6 & 91.9 \\
    \midrule
    MA-HMR (Base)                       & 61.2 & 46.2 & 77.1 & 47.1 & 31.7 & 55.7 & 77.8 & 51.3 & 92.4 \\
    \cisyn{}-HMR (Base)                 & \textbf{59.4} & \underline{45.7} & \textbf{75.1} & \textbf{46.4} & \underline{31.2} & \textbf{54.9} & \textbf{77.0} & \textbf{51.2} & \textbf{91.7} \\
    \midrule
    MA-HMR (Base+\cisyn{})              & 54.8 & 39.4 & 68.0 & 46.4 & 31.0 & 55.1 & 77.6 & 50.9 & 91.8 \\
    \cisyn{}-HMR (Base+\cisyn{})        & \textbf{52.3} & \underline{39.3} & \textbf{65.6} & \textbf{44.9} & \underline{29.9} & \textbf{53.4} & \textbf{74.4} & \textbf{50.4} & \textbf{88.5} \\
    \bottomrule
  \end{tabular}
\end{table*}

\begin{table*}[t]
  \caption{
    \textbf{Comparison of close-interaction training datasets.}
    Each row augments its backbone's base training mix (``Base'', given in the block header) with one close-interaction dataset; ``--'' marks the in-domain split excluded from evaluation.
    All metrics in mm.
  }
  \label{tab:dataset_results}
  \centering
  \small
  \setlength{\tabcolsep}{4pt}
  \begin{tabular}{lccccccccc}
    \toprule
    \multirow{2}{*}{Training data} & \multicolumn{3}{c}{Hi4D} & \multicolumn{3}{c}{CHI3D} & \multicolumn{3}{c}{Harmony4D} \\
    \cmidrule(lr){2-4} \cmidrule(lr){5-7} \cmidrule(lr){8-10}
    & MPJPE $\downarrow$ & PA-MPJPE $\downarrow$ & PVE $\downarrow$ & MPJPE $\downarrow$ & PA-MPJPE $\downarrow$ & PVE $\downarrow$ & MPJPE $\downarrow$ & PA-MPJPE $\downarrow$ & PVE $\downarrow$ \\
    \midrule
    \multicolumn{10}{l}{\emph{\cisyn{}-HMR \quad (Base = AGORA + BEDLAM + DTO-Humans)}} \\
    Base+Hi4D      & --   & --   & --   & 45.1 & 30.2 & \textbf{53.4} & 75.4 & \underline{50.4} & 90.0 \\
    Base+CHI3D     & 56.2 & 42.8 & 70.9 & --   & --   & --   & 75.5 & \textbf{50.2} & 89.9 \\
    Base+Harmony4D & 58.1 & 44.4 & 73.2 & 45.9 & 30.5 & 54.2 & --   & --   & --   \\
    Base+\cisyn{}  & \textbf{52.3} & \textbf{39.3} & \textbf{65.6} & \textbf{44.9} & \textbf{29.9} & \textbf{53.4} & \textbf{74.4} & \underline{50.4} & \textbf{88.5} \\
    \midrule
    \multicolumn{10}{l}{\emph{Multi-HMR~\cite{baradel2024multi} \quad (Base = BEDLAM + DTO-Humans)}} \\
    Base+Hi4D      & --   & --   & --   & 62.9 & 35.1 & 75.7 & 99.4 & \textbf{59.8} & 115.5 \\
    Base+CHI3D     & 81.3 & 48.4 & 98.8 & --   & --   & --   & 103.0 & 62.0 & 118.5 \\
    Base+Harmony4D & 81.4 & 48.5 & 99.3 & 64.2 & 37.9 & 77.7 & --   & --   & --   \\
    Base+\cisyn{}  & \textbf{74.2} & \textbf{46.4} & \textbf{91.3} & \textbf{60.3} & \textbf{34.7} & \textbf{74.3} & \textbf{98.9} & \underline{60.4} & \textbf{114.0} \\
    \bottomrule
  \end{tabular}
  \vspace{-3mm}
\end{table*}

Our model builds on MA-HMR~\cite{wang2025mahmr}, a single-stage multi-person HMR model that itself builds on SAT-HMR~\cite{su2025sathmr}. We prefer MA-HMR over head-anchored alternatives such as Multi-HMR~\cite{baradel2024multi}, whose detection is unreliable in close interactions where heads are often occluded. Figure~\ref{fig:model} illustrates the architecture. Section~\ref{sec:method:model:baseline} reviews the baseline, and Section~\ref{sec:method:model:loss} describes our uncertainty-aware extensions.

\subsubsection{Baseline: MA-HMR}
\label{sec:method:model:baseline}

The backbone is a DINOv2~\cite{oquab2023dinov2} ViT-B/14 with a Scale-Adaptive Token (SAT) mechanism that extracts features at three patch scales (14, 28, 56) and adaptively selects tokens via per-token confidence scores, enabling efficient high-resolution processing of dense multi-person scenes.
Image features are decoded by a 6-layer transformer following the DAB-DETR~\cite{liu2022dabdetr} design with 50 learnable object queries; bounding-box reference points modulate the cross-attention at each layer via sinusoidal positional encodings.
Lightweight MLP heads predict SMPL~\cite{2015smpl} body parameters, a bounding box, and detection confidence from per-query decoder features, and the camera field of view from the CLS token; architecture details are provided in Supp. Sec.~\ref{app:model}.

\subsubsection{Uncertainty-Aware Extensions}
\label{sec:method:model:loss}

Close interactions introduce heavy mutual occlusion, making supervision unreliable for occluded joints, whereas a fixed-weight L1 loss treats all joints equally. We address this with two lightweight additions.

\paragraph{Per-joint occlusion weight.}
For each person, we compute a per-joint occlusion indicator from the ground truth: we depth-order all other people, project their SMPL meshes onto the image plane, and set $o_j{=}1$ if joint $j$ falls within a closer person's silhouette. This defines a per-joint weight $w_j = 1 - 0.9\,o_j \in \{0.1, 1.0\}$ that replaces the baseline per-joint L1 $\|\hat{p}_j - p_j\|_1$ with $w_j \|\hat{p}_j - p_j\|_1$, down-weighting supervision where it is least reliable.

\paragraph{Uncertainty head and Laplacian NLL loss.}
To make the model aware of its own uncertainty, we add a lightweight MLP head to the decoder that predicts a per-joint log-standard-deviation $\sigma_j \in \mathbb{R}$ from the decoder features (initialization and clamping details in Supp. Sec.~\ref{app:model}).
Following~\cite{kendall2017uncertainties}, we model each joint as an isotropic Laplace vector with standard deviation $\exp(\sigma_j)$ and extend the occlusion-weighted L1 into a Laplacian NLL:
\begin{equation}
\label{eq:uncertainty_loss}
    \mathcal{L}_j \;=\;
    \sqrt{2}\;e^{-\sigma_j}\;w_j\,\bigl\|\hat{p}_j - p_j\bigr\|_1
    \;+\; \sigma_j,
\end{equation}
applied to both 2D-projected and root-aligned 3D joints; the $\sigma_j$ term is the distributional entropy that prevents $\sigma_j \to -\infty$. This replaces the L1 objective for $\mathcal{L}_\text{j3d}$ and $\mathcal{L}_\text{j2d}$; all other losses are unchanged.

\paragraph{Total training objective.}
Combining the joint losses with the inherited MA-HMR terms, the model is trained end-to-end with a weighted sum of nine losses:
\begin{equation}
\label{eq:total_loss}
\begin{aligned}
\mathcal{L} \;=\;\;
&\lambda_\text{j3d}\mathcal{L}_\text{j3d}
+ \lambda_\text{j2d}\mathcal{L}_\text{j2d}
+ \lambda_\text{pose}\mathcal{L}_\text{pose}
+ \lambda_\text{shape}\mathcal{L}_\text{shape}\\
+\;
&\lambda_\text{box}\mathcal{L}_\text{box}
+ \lambda_\text{detect}\mathcal{L}_\text{detect}
+ \lambda_\text{depth}\mathcal{L}_\text{depth}\\
+\;
&\lambda_\text{fov}\mathcal{L}_\text{fov}
+ \lambda_\text{scale}\mathcal{L}_\text{scale}
\end{aligned}
\end{equation}
Here $\mathcal{L}_\text{j3d}$ and $\mathcal{L}_\text{j2d}$ are the per-joint NLLs of Eq.~\ref{eq:uncertainty_loss}; the remaining terms are inherited from MA-HMR~\cite{wang2025mahmr}. Loss weights and full definitions are in Supp. Secs.~\ref{app:model:losses} and~\ref{app:model:lossweights}.

\section{Experiments}
\label{sec:experiments}

We evaluate the \cisyn{} dataset and \cisyn{}-HMR on three close-interaction HMR benchmarks, Hi4D~\cite{yin2023hi4d}, CHI3D~\cite{fieraru2025chi3d}, and Harmony4D~\cite{khirodkar2024harmony4d}, and conduct an ablation study.
\subsection{Experimental Setup}
\label{sec:exp:setup}
\begin{figure*}[t]
  \centering
  \setlength{\tabcolsep}{1pt}
  \renewcommand{\arraystretch}{1.0}
  \newcommand{\qw}{0.19\linewidth}
  \newcommand{\qhead}[1]{\scriptsize #1}
  \newcommand{\qrowlabel}[1]{\rotatebox{90}{\scriptsize #1}}
  \newcommand{\qna}[1]{\raisebox{#1\linewidth}{\textcolor{black!40}{\scriptsize n/a}}}
  \begin{tabular}{@{}c@{\hspace{2pt}}ccccc@{}}
    & \qhead{Input} & \qhead{GT} & \qhead{PromptHMR~\cite{wang2025prompthmr}} & \qhead{MA-HMR~\cite{wang2025mahmr}} & \qhead{\cisyn{}-HMR (Ours)} \\
    \qrowlabel{Hi4D}      & \includegraphics[width=\qw]{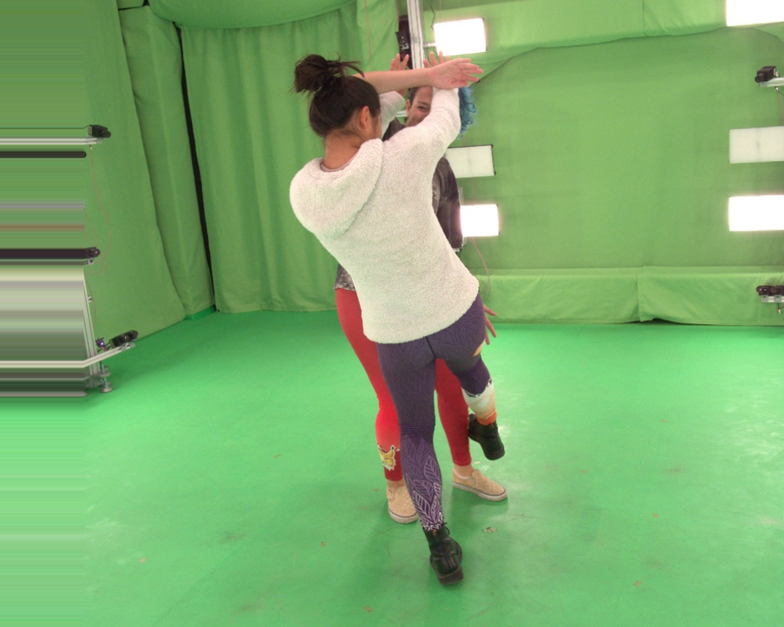}      & \includegraphics[width=\qw]{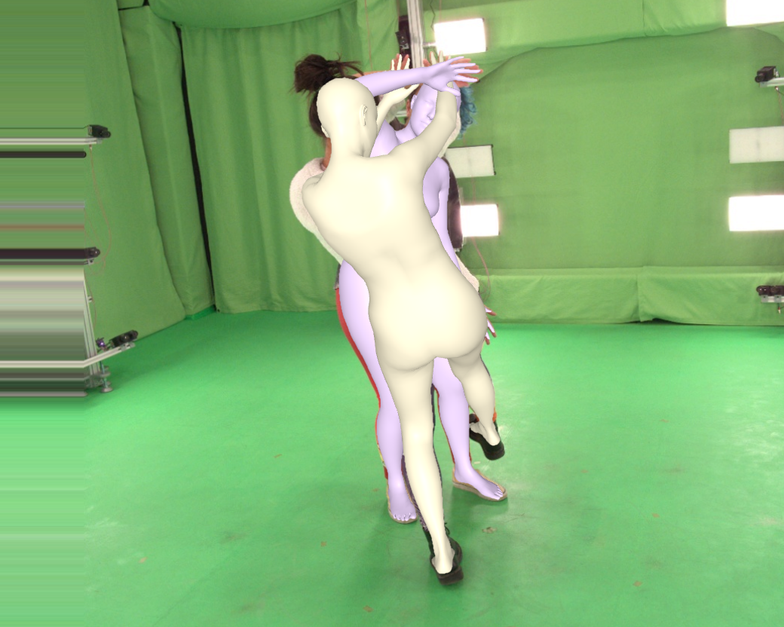}      & \qna{0.072}           & \includegraphics[width=\qw]{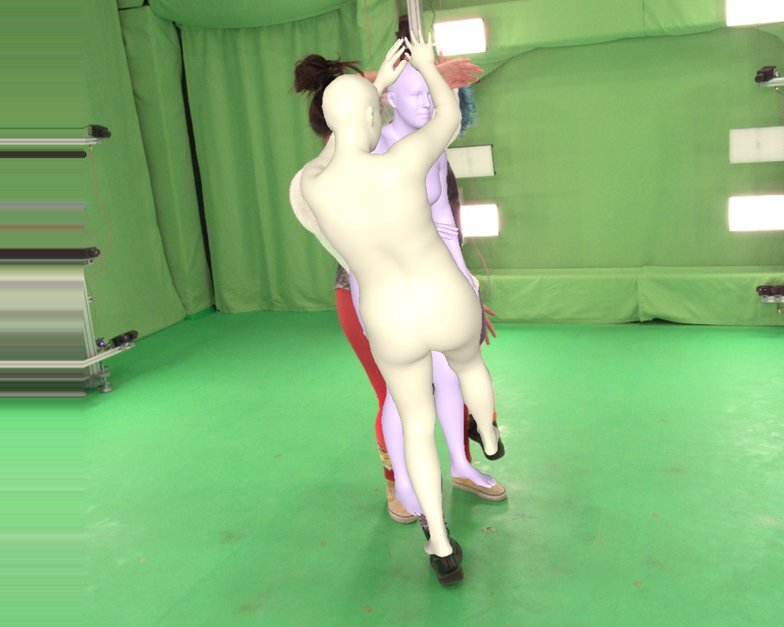}      & \includegraphics[width=\qw]{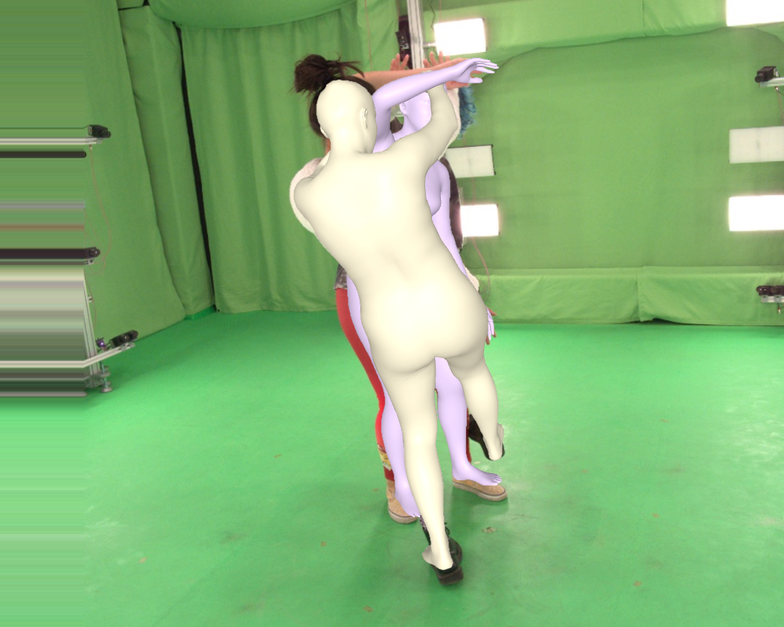}      \\
    \qrowlabel{CHI3D}     & \includegraphics[width=\qw]{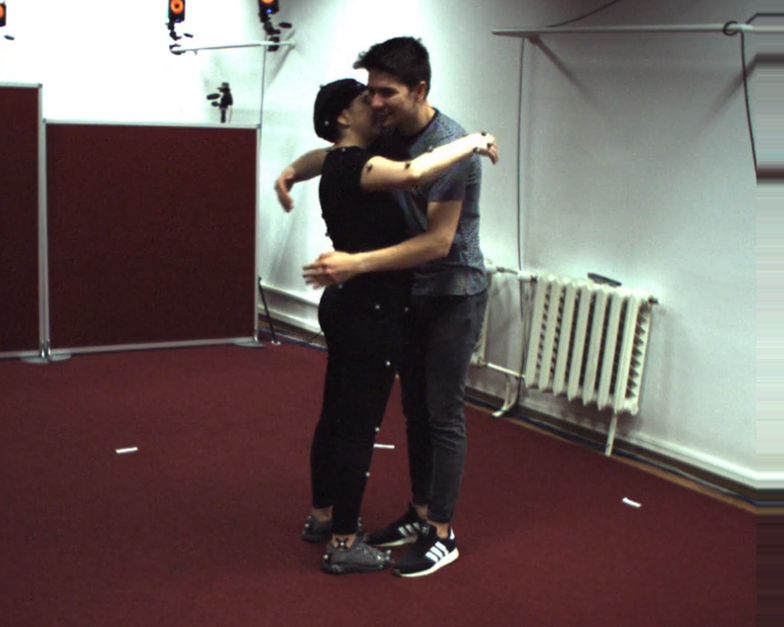}     & \includegraphics[width=\qw]{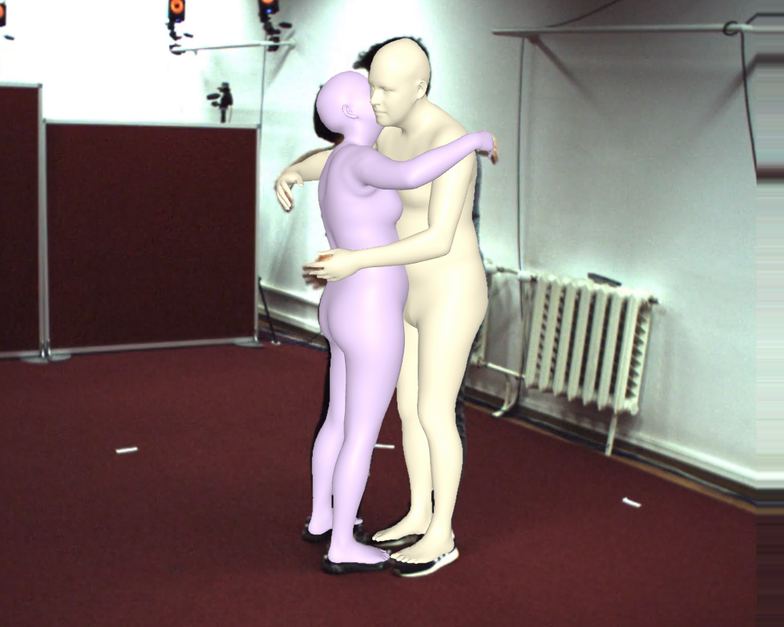}     & \qna{0.072}           & \includegraphics[width=\qw]{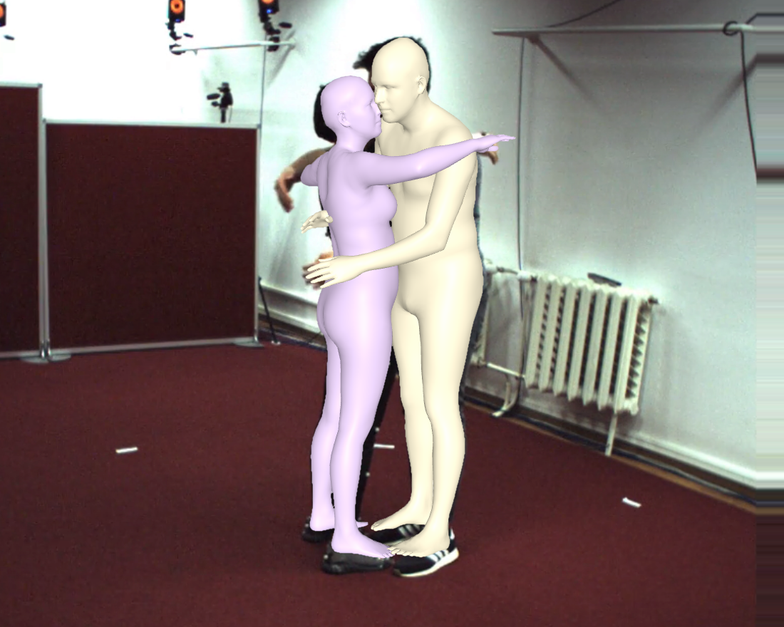}     & \includegraphics[width=\qw]{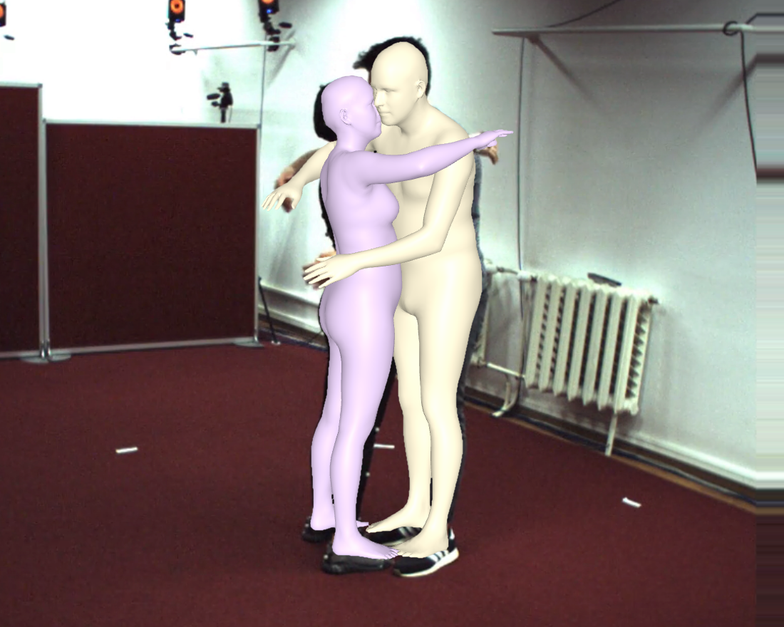}     \\
    \qrowlabel{Harmony4D} & \includegraphics[width=\qw]{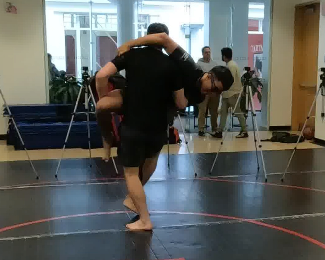} & \includegraphics[width=\qw]{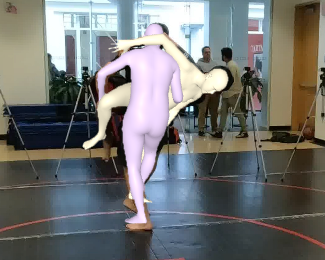} & \includegraphics[width=\qw]{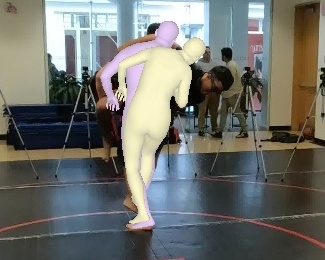} & \includegraphics[width=\qw]{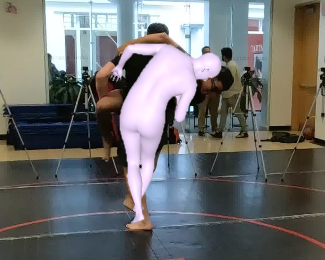} & \includegraphics[width=\qw]{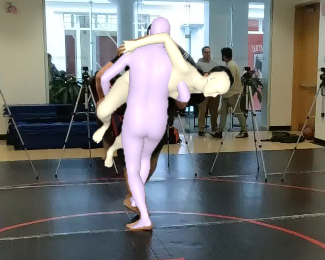} \\
    \qrowlabel{ExPI}      & \includegraphics[width=\qw]{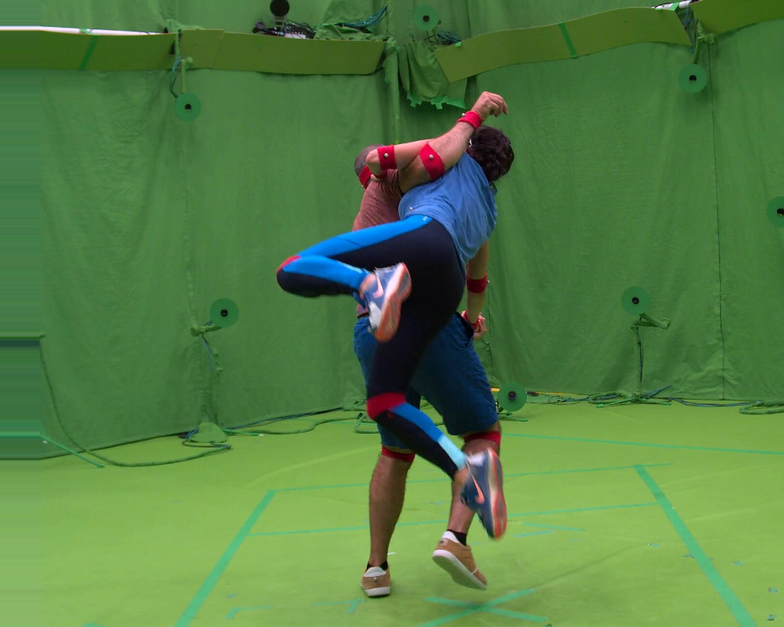}      & \qna{0.072}          & \includegraphics[width=\qw]{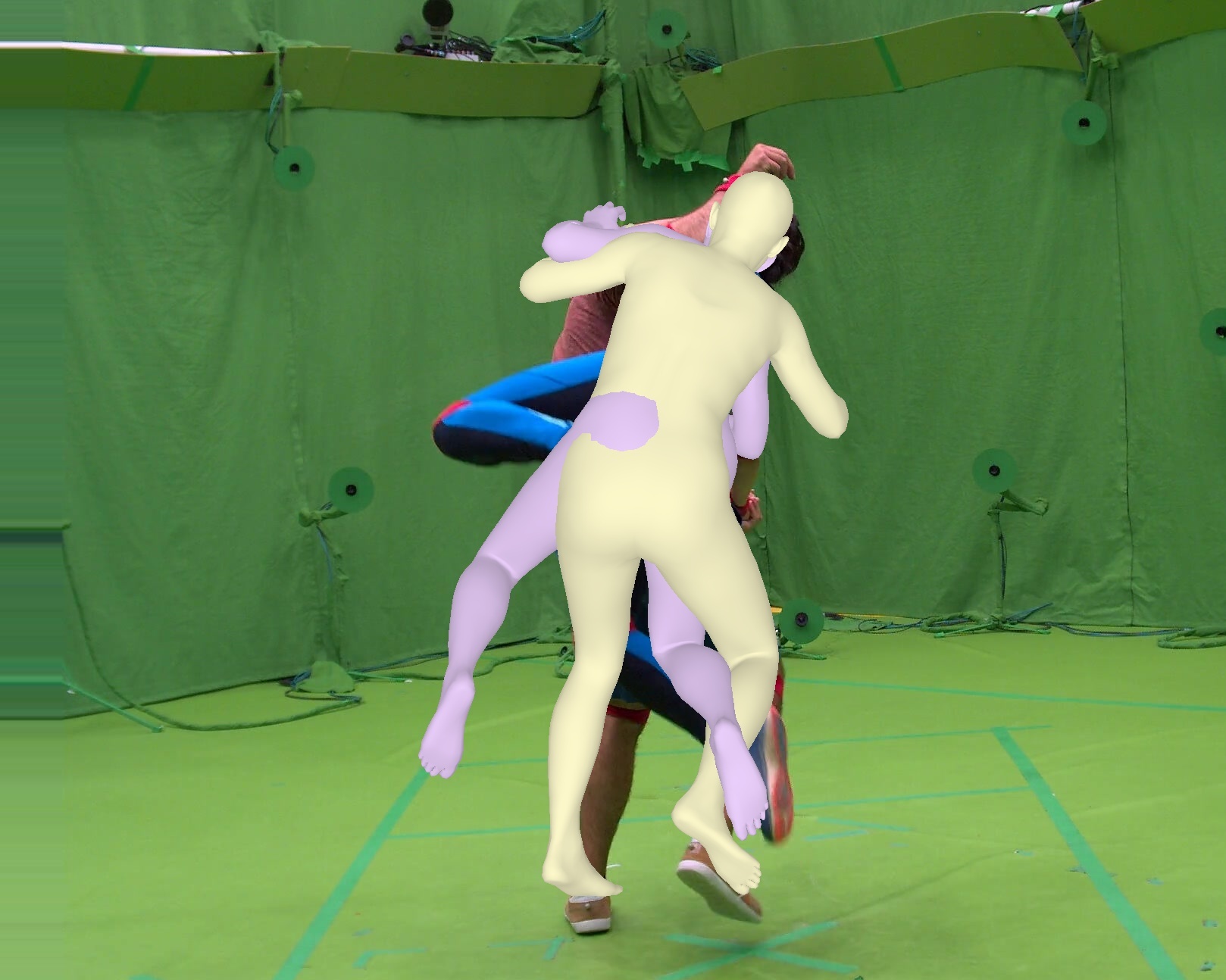}      & \includegraphics[width=\qw]{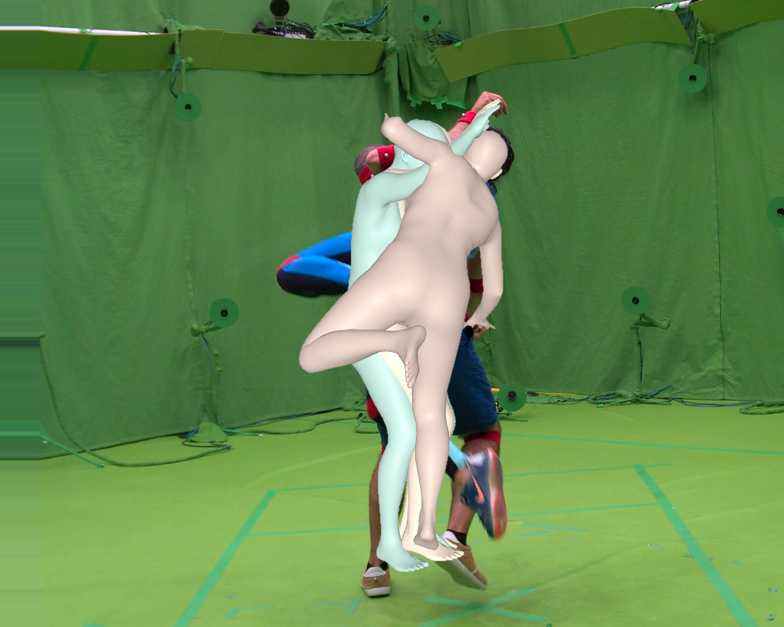}      & \includegraphics[width=\qw]{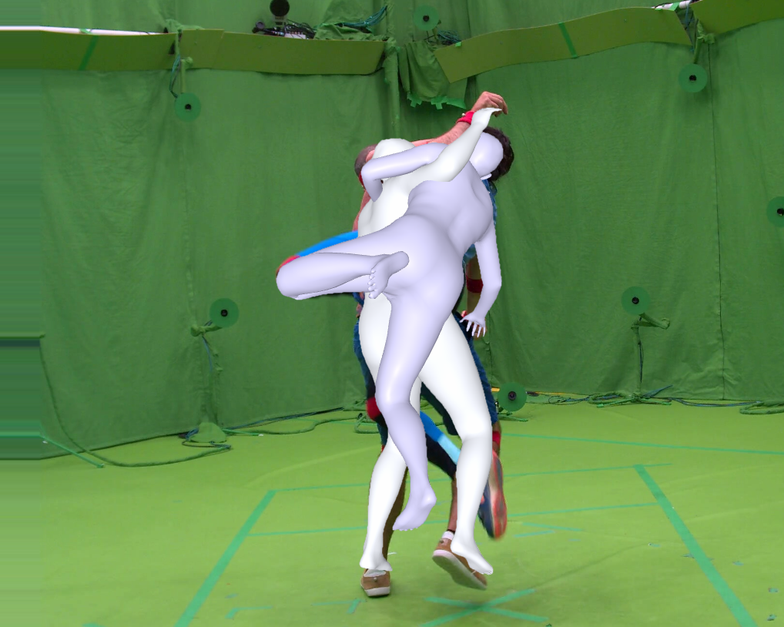}      \\
    \qrowlabel{ITW}       & \includegraphics[width=\qw]{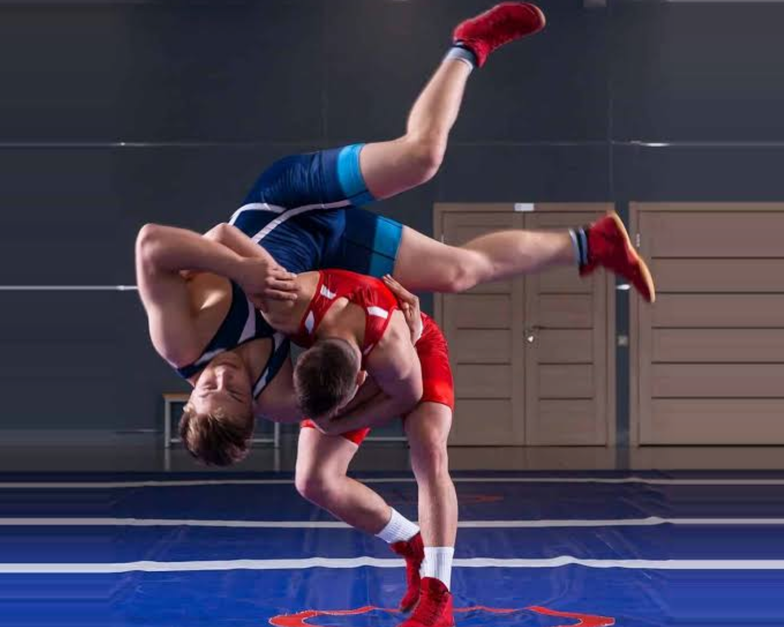}       & \qna{0.072}          & \includegraphics[width=\qw]{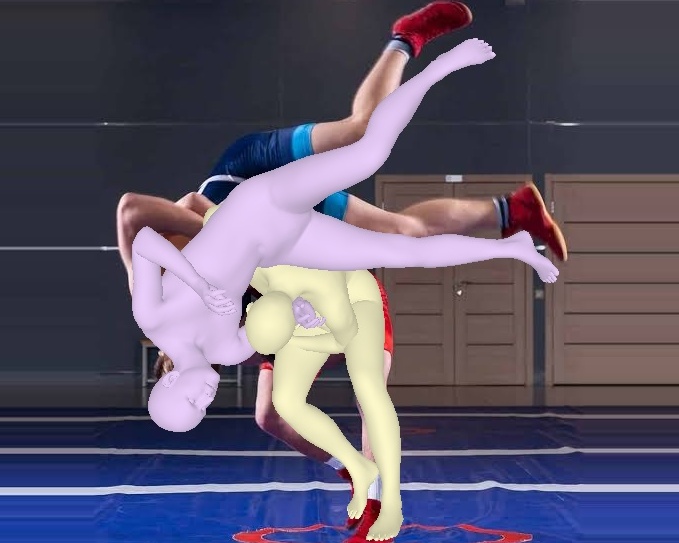}       & \includegraphics[width=\qw]{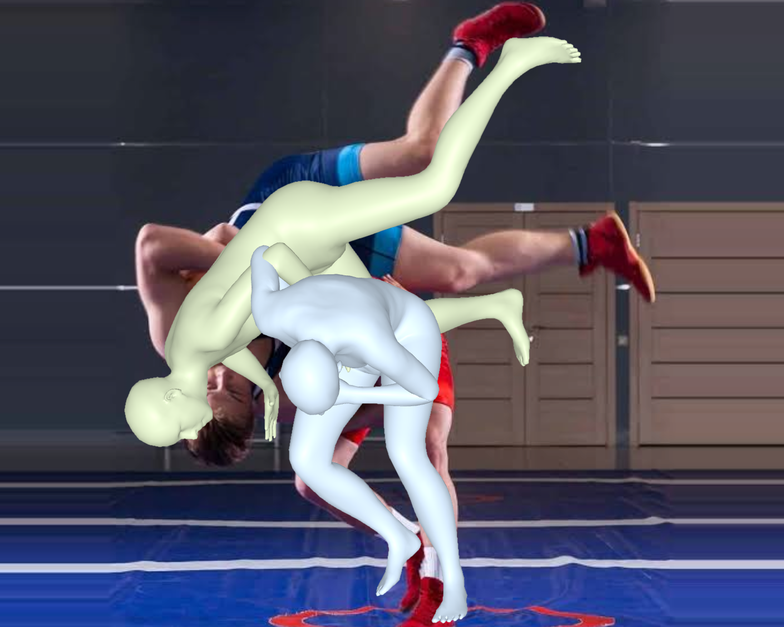}       & \includegraphics[width=\qw]{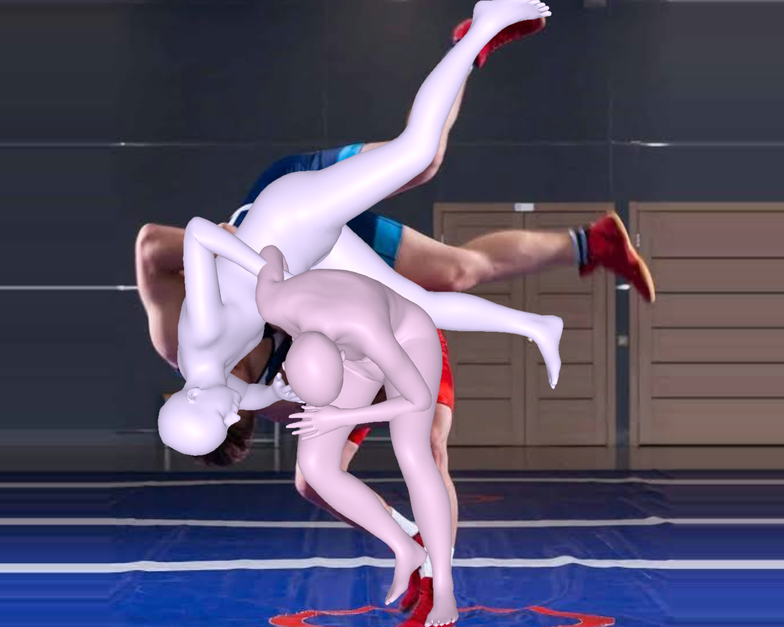}       \\
  \end{tabular}
  \caption{%
    \textbf{Qualitative comparison on close-interaction samples.}
    Top three rows (Hi4D, CHI3D, Harmony4D): ground-truth meshes are available.
    Bottom two rows (ExPI and an in-the-wild image, ITW): no 3D ground truth is available.
    PromptHMR is shown only on Harmony4D, ExPI and ITW, because its released checkpoint is trained on Hi4D and CHI3D, making these the only held-out settings.
  }
  \label{fig:qualitative}
\vspace{-2mm}
\end{figure*}

We adopt the final checkpoint of MA-HMR~\cite{wang2025mahmr}, a state-of-the-art single-stage multi-person SMPL recovery method, as our baseline; it is trained on AGORA~\cite{patel2021agora}, BEDLAM~\cite{black2023bedlam}, and DTO-Humans~\cite{wang2025mahmr}, a mix we refer to as ``Base''. DTO-Humans re-annotates the in-the-wild images of 4D-Humans~\cite{goel20234dhumans} with a higher-quality pseudo-GT pipeline, whereas AGORA and BEDLAM are synthetic. Our model with the uncertainty extension, \cisyn{}-HMR, is initialized from this checkpoint (except for the freshly added uncertainty head) and fine-tuned for 6 epochs on Base augmented with \cisyn{}. Additional training details are in Supp. Sec.~\ref{app:exp}.

\paragraph{Evaluation Benchmarks and Metrics.}
The CHI3D~\cite{fieraru2025chi3d} test set ground-truth annotations are not released, so we evaluate on a 16$\times$ temporally downsampled subset of its training split as our test set. Following PromptHMR~\cite{wang2025prompthmr}, we use pairs 12, 15, and 32 of Hi4D~\cite{yin2023hi4d} as the test set. For Harmony4D~\cite{khirodkar2024harmony4d}, we use the standard test set, 16$\times$ temporally downsampled to a tractable size. Between them, these three benchmarks span casual interaction, social contact, combat, dance, and sports; Supp. Sec.~\ref{app:results:percate} adds the acrobatic interactions of ExPI~\cite{guo2022expi} and reports results per category. We report three standard HMR metrics: Mean Per-Joint Position Error (MPJPE), Procrustes-Aligned MPJPE (PA-MPJPE), and Per-Vertex Error (PVE)~\cite{black2023bedlam}, all in millimeters ($\downarrow$). All three are root-relative and computed over detected people; detection recall and inter-person accuracy are reported in Supp. Sec.~\ref{app:results:pair}. No model is trained or fine-tuned on any benchmark dataset.

\subsection{Main Results}
\label{sec:exp:main}

Table~\ref{tab:main_results} reports close-interaction performance against state-of-the-art monocular multi-person methods.
Within either block, replacing MA-HMR with \cisyn{}-HMR reduces the error on every metric; within either model, adding \cisyn{} yields a further reduction. The model and the dataset thus contribute independent, complementary gains.
PromptHMR remains slightly ahead on PA-MPJPE; however, PA-MPJPE rigidly aligns each prediction to the ground-truth pose before measuring error, an operation not available at deployment time. The alignment-free MPJPE and PVE, on which \cisyn{}-HMR is best on all three benchmarks, are therefore the more representative indicators of real-inference quality.
\begin{table*}[th]
  \caption{
    \textbf{Ablation of the \cisyn{} framework.}
    Each row removes one component from the full framework (last row) and is trained for the same number of epochs, by which point all models have converged.
    \emph{Phys.\ Sim.}\ is our physics simulation (Sec.~\ref{sec:method:dataset:simulate}); \N: motions from DualHuman~\cite{fang2024dualhuman} and rendered with the same pipeline, \Y: physics-simulated motions, i.e.\ \cisyn{}.
    \emph{Occ.}\ is the per-joint occlusion mask and \emph{Unc.}\ is the uncertainty head with Laplacian NLL (Sec.~\ref{sec:method:model:loss}).
    All metrics in mm.
  }
  \label{tab:ablation}
  \centering
  \small
  \setlength{\tabcolsep}{4pt}
  \begin{tabular}{ccc|ccccccccc}
    \toprule
    \multirow{2}{*}{Phys.\ Sim.} & \multirow{2}{*}{Occ.} & \multirow{2}{*}{Unc.} & \multicolumn{3}{c}{Hi4D} & \multicolumn{3}{c}{CHI3D} & \multicolumn{3}{c}{Harmony4D} \\
    \cmidrule(lr){4-6} \cmidrule(lr){7-9} \cmidrule(lr){10-12}
    & & & MPJPE $\downarrow$ & PA-MPJPE $\downarrow$ & PVE $\downarrow$ & MPJPE $\downarrow$ & PA-MPJPE $\downarrow$ & PVE $\downarrow$ & MPJPE $\downarrow$ & PA-MPJPE $\downarrow$ & PVE $\downarrow$ \\
    \midrule
    \N & \Y & \Y & 55.4 & 41.6 & 69.3 & 45.8 & 30.8 & 54.2 & 75.8 & 50.6 & 90.4 \\
    \Y & \N & \Y & 52.6 & 39.4 & 65.8 & 46.2 & 30.5 & 54.5 & 74.6 & 50.5 & 88.7 \\
    \Y & \Y & \N & 54.6 & \textbf{39.3} & 68.2 & 46.4 & 30.9 & 54.9 & 76.6 & 50.9 & 90.8 \\
    \Y & \Y & \Y & \textbf{52.3} & \textbf{39.3} & \textbf{65.6} & \textbf{44.9} & \textbf{29.9} & \textbf{53.4} & \textbf{74.4} & \textbf{50.4} & \textbf{88.5} \\
    \bottomrule
  \end{tabular}
\vspace{-2mm}
\end{table*}

Table~\ref{tab:dataset_results} compares \cisyn{} against the existing real close-interaction datasets Hi4D~\cite{yin2023hi4d}, CHI3D~\cite{fieraru2025chi3d}, and Harmony4D~\cite{khirodkar2024harmony4d} as training resources.
Within each block, only the close-interaction dataset added to ``Base'' varies, so any difference is attributable to the training data alone; the benchmark matching the added dataset is excluded from evaluation, and the comparison is repeated on two backbones, \cisyn{}-HMR and Multi-HMR~\cite{baradel2024multi}.
For the Multi-HMR rows, we initialize from a checkpoint trained on BEDLAM only and continue training for 10 epochs on each mix. Since Multi-HMR is supervised with SMPL-X, the SMPL-only annotations of Hi4D, Harmony4D, and DTO-Humans are converted to SMPL-X beforehand using the official optimization-based SMPL-to-SMPL-X fitting procedure~\cite{pavlakos2019smplx}.
Adding \cisyn{} gives the lowest MPJPE and PVE on every benchmark for both backbones and the lowest PA-MPJPE on Hi4D and CHI3D. The gain is therefore consistent across both backbones and all three benchmarks, indicating that the contact diversity provided by physics simulation generalizes better across capture conditions than additional samples drawn from a single real source.
The benefit extends to goal-directed categories such as dance and casual interaction (Supp.\ Sec.~\ref{app:results:percate}), suggesting that broad coverage of two-body configurations matters more than matching the dynamics of the target interaction.
These gains do not come at the cost of general-domain performance. Adding \cisyn{} to the training mix leaves all three metrics on the single-person, in-the-wild 3DPW~\cite{von20183dpw} benchmark unchanged or slightly improved (Supp.\ Sec.~\ref{app:results:3dpw}).
More quantitative results in Supp. Sec.~\ref{app:results:quantitative}.

\subsection{Qualitative Results}
\label{sec:exp:qualitative}

Figure~\ref{fig:qualitative} shows representative close-interaction reconstructions on Hi4D~\cite{yin2023hi4d}, CHI3D~\cite{fieraru2025chi3d}, Harmony4D~\cite{khirodkar2024harmony4d}, ExPI~\cite{guo2022expi}, and an in-the-wild image, focusing on heavily entangled poses.
In these examples, MA-HMR places the two bodies close together but does not align them accurately, whereas \cisyn{}-HMR matches the ground-truth meshes more closely on all three annotated benchmarks.
PromptHMR is shown on the Harmony4D, ExPI and in-the-wild rows only, since its released checkpoint is trained on Hi4D and CHI3D but not on these data; in all three held-out settings, \cisyn{}-HMR produces more accurate reconstructions than both PromptHMR and MA-HMR.
The in-the-wild result shows that training on \cisyn{} generalizes beyond studio capture.
More results in Supp. Secs.~\ref{app:results:qualitative} and~\ref{app:results:cisyn_qual}.

\subsection{Ablation Study}
\label{sec:exp:ablation}

Table~\ref{tab:ablation} presents the ablation. Each row disables one component of the framework: \emph{Phys.\ Sim.}\ (whether the synthetic training set is our physics-simulated data or DualHuman~\cite{fang2024dualhuman} motions rendered with the same pipeline), the per-joint occlusion mask (\emph{Occ.}), and the uncertainty head (\emph{Unc.}).

Substituting DualHuman motions for our physics-simulated ones degrades all metrics on all three benchmarks: the physically accurate contact produced by the simulator, and not the synthetic imagery alone, is what benefits training.
Removing the occlusion mask likewise degrades every metric, since down-weighting supervision on occluded joints prevents the model from overfitting to a generic prior over occluded geometry and avoids pseudo-GT label noise on those joints in the real portion of the training mix.
Removing the uncertainty head degrades MPJPE and PVE on all three benchmarks and leaves PA-MPJPE roughly flat on Hi4D; it extends the same idea as the mask by replacing the binary weight with a learned one, so supervision is softly down-weighted on any joint whose 3D position the model judges ambiguous, which subsumes mutual occlusion.
The predicted $\sigma_j$ peaks on the occluded and least accurately reconstructed joints (Supp.\ Sec.~\ref{app:results:uncertainty}).
With neither component, the model reduces to MA-HMR (Base+\cisyn{}) of Table~\ref{tab:main_results}, i.e.\ uniform joint weights; each component alone improves most metrics over it, and together they are best on every metric.
We do not search further over weighting schemes or robust losses, since each requires a full training run.

\section{Conclusion}
\label{sec:conclusion}

We presented \cisyn{}, a large-scale synthetic dataset for close human-human interaction HMR generated by physics-simulated collisions between SMPL-X-rigged humanoids, together with an uncertainty-aware loss that lets the model down-weight unreliable supervision on occluded joints. Adding \cisyn{} to the training mix and replacing the regression loss with our uncertainty-aware variant yields consistent improvements over a state-of-the-art baseline on Hi4D, CHI3D, and Harmony4D, with each component contributing independent gains in our ablation. The pipeline requires no manual interaction design and scales straightforwardly to more sequences and more interactors, making physics-based close-interaction data a practical resource for the community.

\paragraph{Future work and limitations.}
Our actors are passive bodies without motor control, so \cisyn{} does not model coordinated, goal-directed interaction, although its gains extend to such categories (Sec.~\ref{sec:exp:main}).
The pipeline opens several directions, from action-driven simulation and generator-in-the-loop training to more interactors and contact annotations; on the model side, nothing currently enforces non-penetration between the two predicted meshes, which can interpenetrate on difficult interactions (Supp. Sec.~\ref{app:failure}).

\clearpage
\section*{Acknowledgments}
Y.D. is supported by the Jardine Foundation. L.P. is funded by the Podium Institute for Sports Medicine and Technology, at the University of Oxford. C.R. is partially supported by the European Union (ERC, Volute, 101222037). Views and opinions expressed are however those of the author(s) only and do not necessarily reflect those of the European Union or the European Research Council. Neither the European Union nor the granting authority can be held responsible for them.

The authors acknowledge the use of resources provided by the Isambard-AI National AI Research Resource (AIRR) \cite{mcintoshsmith2024isambard}. Isambard-AI is operated by the University of Bristol and is funded by the UK Government’s Department for Business, Innovation, Science and Trade (DBIST) via UK Research and Innovation; and the Science and Technology Facilities Council [ST/AIRR/I-A-I/1023].

{
    \small
    \bibliographystyle{ieeenat_fullname}
    \bibliography{refs}
}

\clearpage
\appendix
\maketitlesupplementary

\setcounter{figure}{0}\renewcommand{\thefigure}{A\arabic{figure}}\renewcommand{\theHfigure}{A\arabic{figure}}
\setcounter{table}{0}\renewcommand{\thetable}{A\arabic{table}}\renewcommand{\theHtable}{A\arabic{table}}
\setcounter{equation}{0}\renewcommand{\theequation}{A\arabic{equation}}\renewcommand{\theHequation}{A\arabic{equation}}

\section{Dataset Generation Details}
\label{app:pipeline}
\begin{figure*}[t]
  \centering
  \setlength{\tabcolsep}{0pt}
  \renewcommand{\arraystretch}{0.5}
  \newlength{\dsh}
  \newcommand{\dscolspec}{@{}c@{\hspace{1.5pt}}c@{\hspace{6pt}}c@{\hspace{1.5pt}}c@{\hspace{6pt}}c@{\hspace{1.5pt}}c@{}}

  {\small\textit{Two-person interactions}}\\[4pt]
  \setlength{\dsh}{0.085\linewidth}
  \begin{tabular}{\dscolspec}
    \includegraphics[height=\dsh]{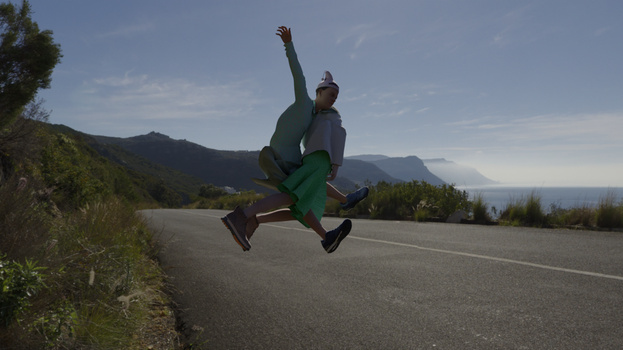} &
    \includegraphics[height=\dsh]{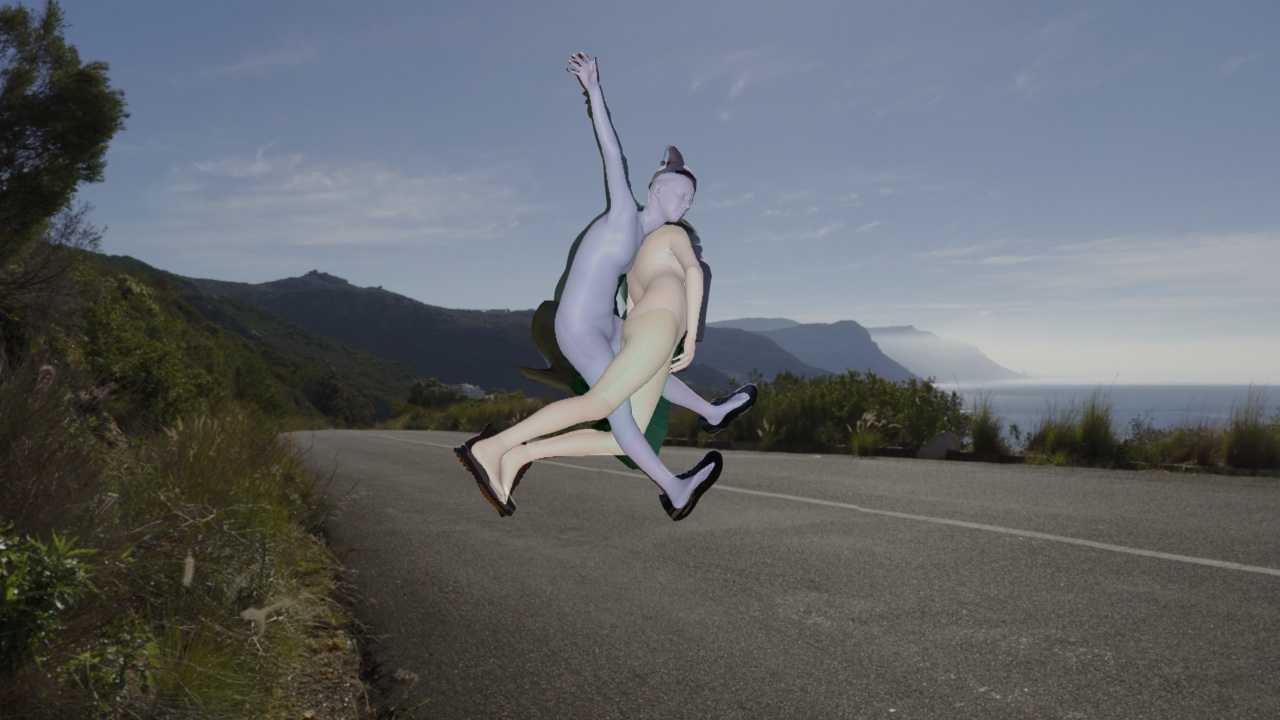} &
    \includegraphics[height=\dsh]{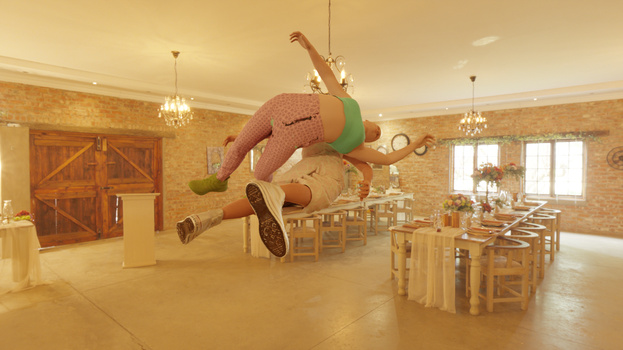} &
    \includegraphics[height=\dsh]{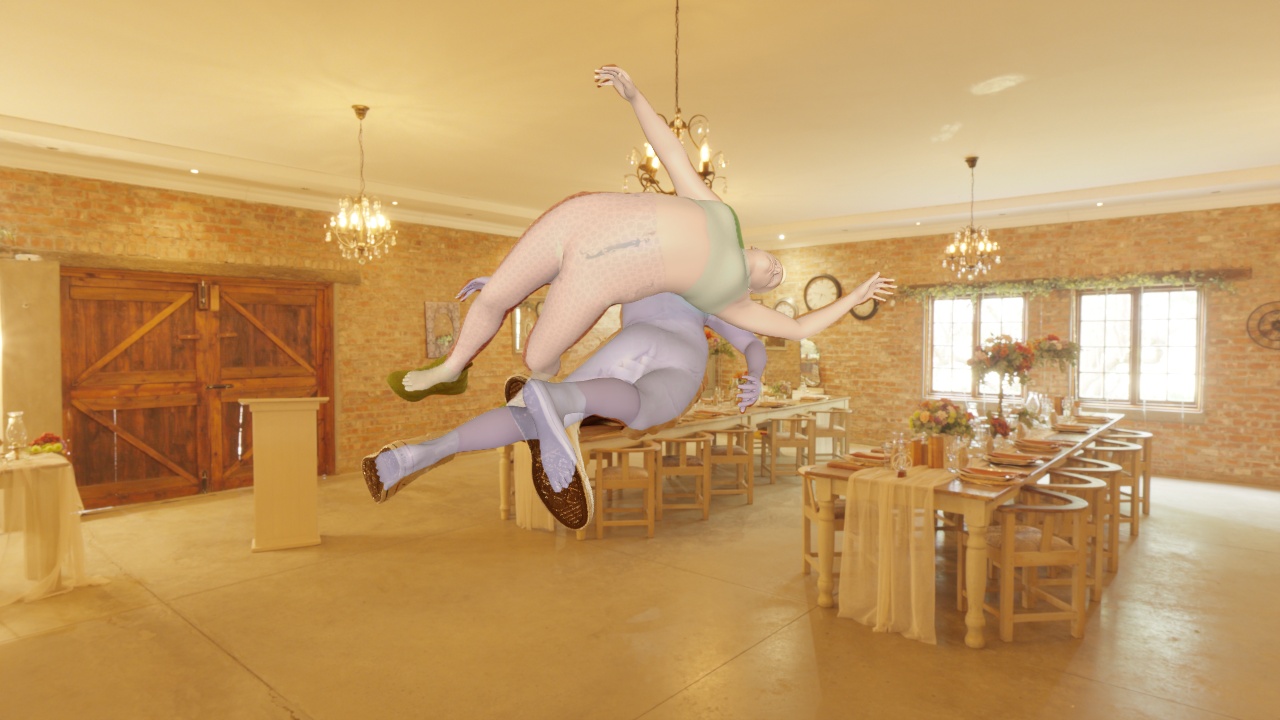} &
    \includegraphics[height=\dsh]{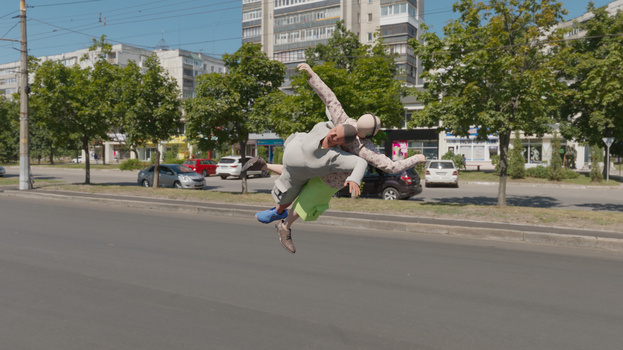} &
    \includegraphics[height=\dsh]{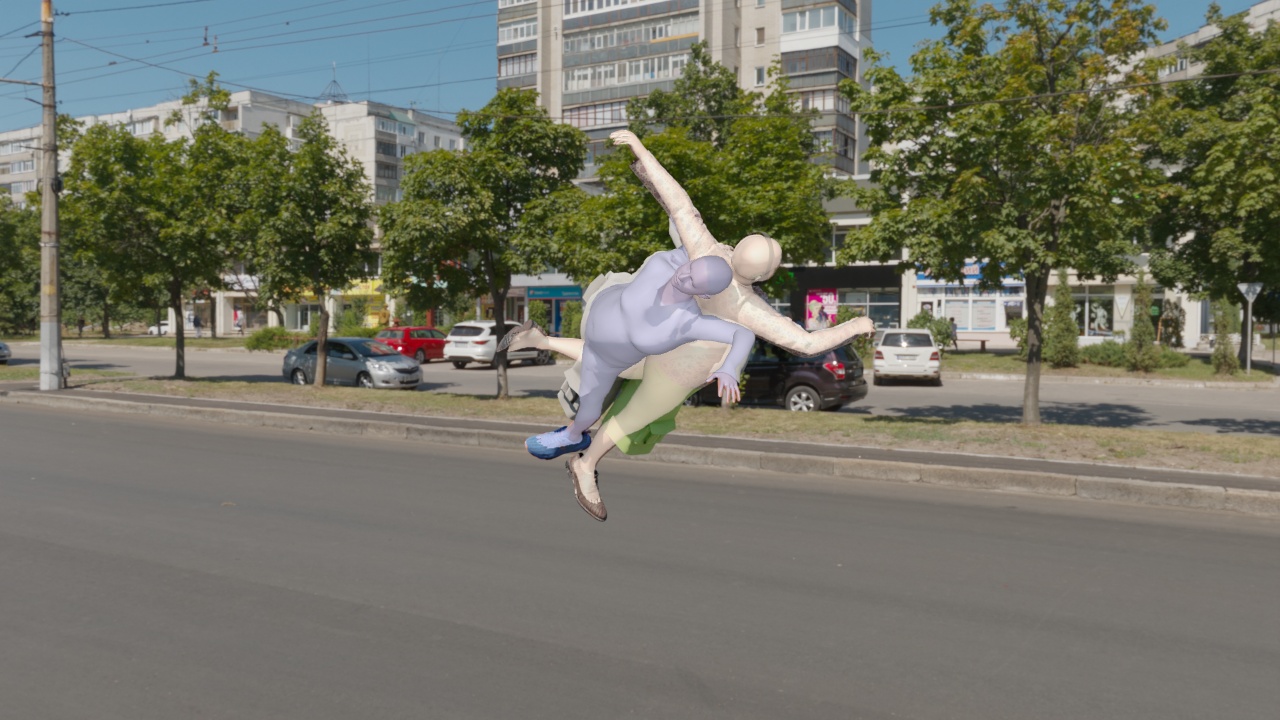} \\[3pt]
  \end{tabular}\\[3pt]
  \setlength{\dsh}{0.183\linewidth}
  \begin{tabular}{\dscolspec}
    \includegraphics[height=\dsh]{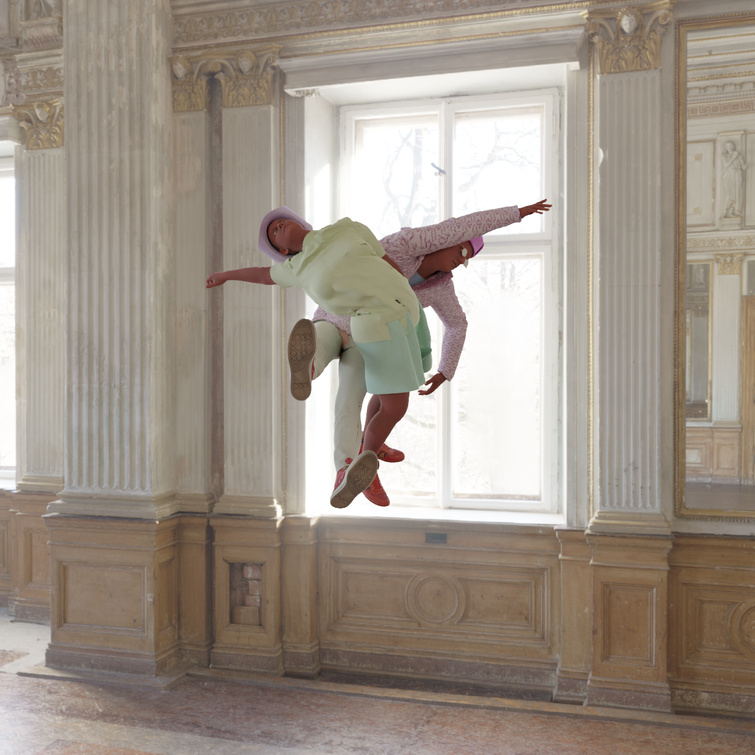} &
    \includegraphics[height=\dsh]{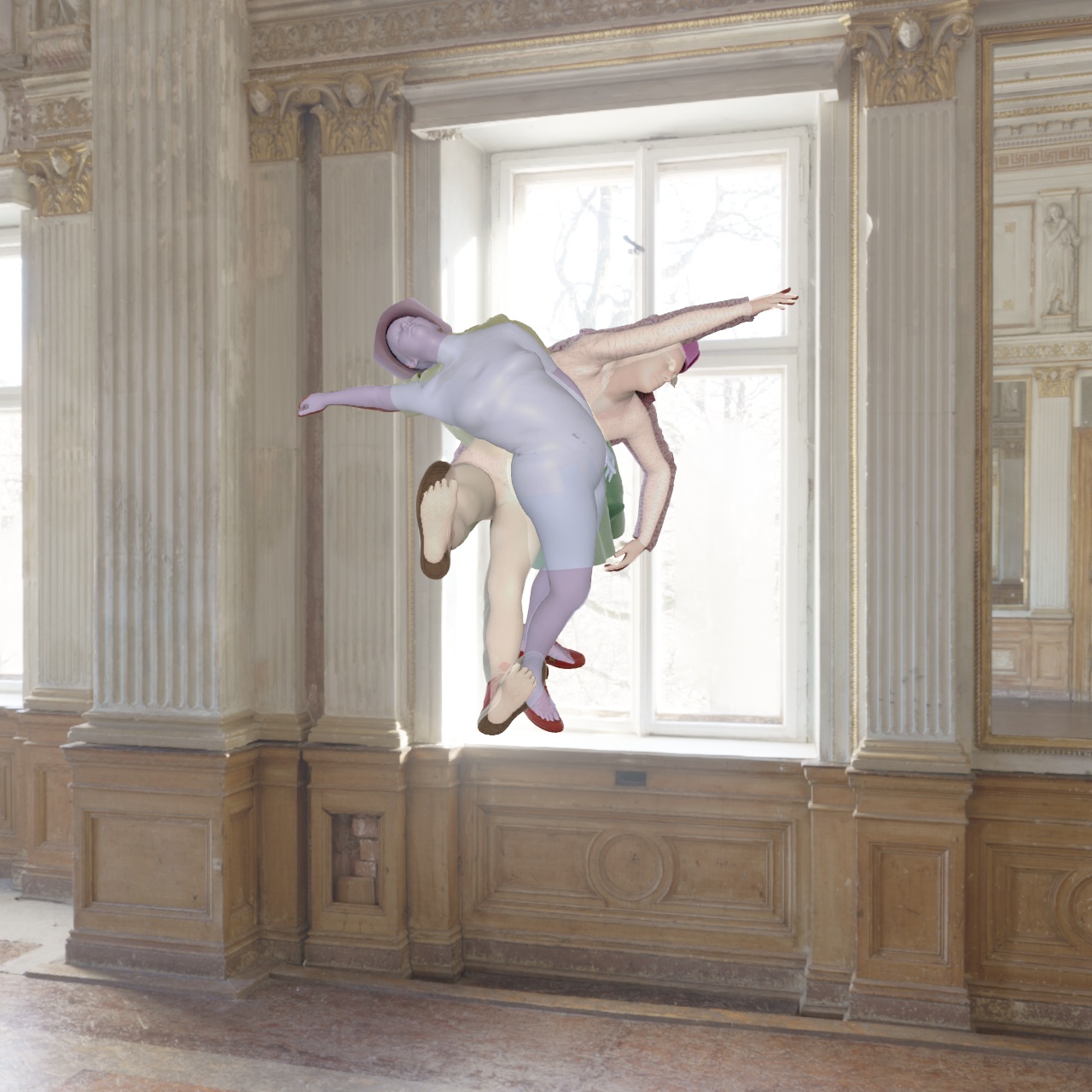} &
    \includegraphics[height=\dsh]{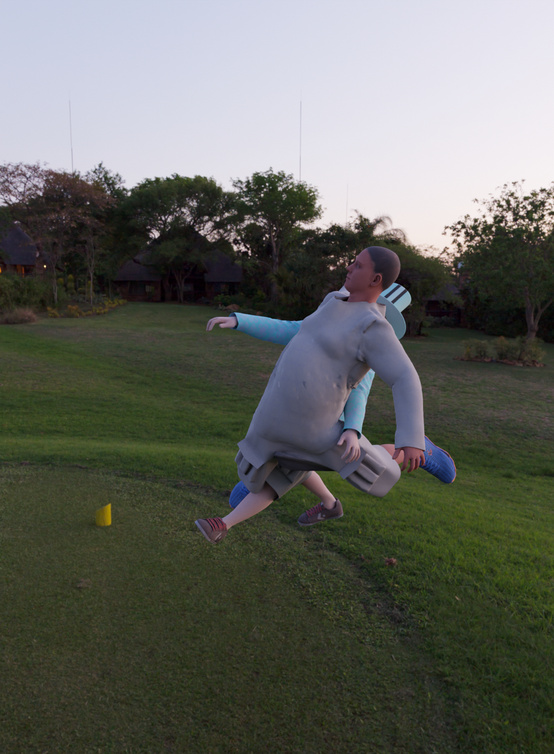} &
    \includegraphics[height=\dsh]{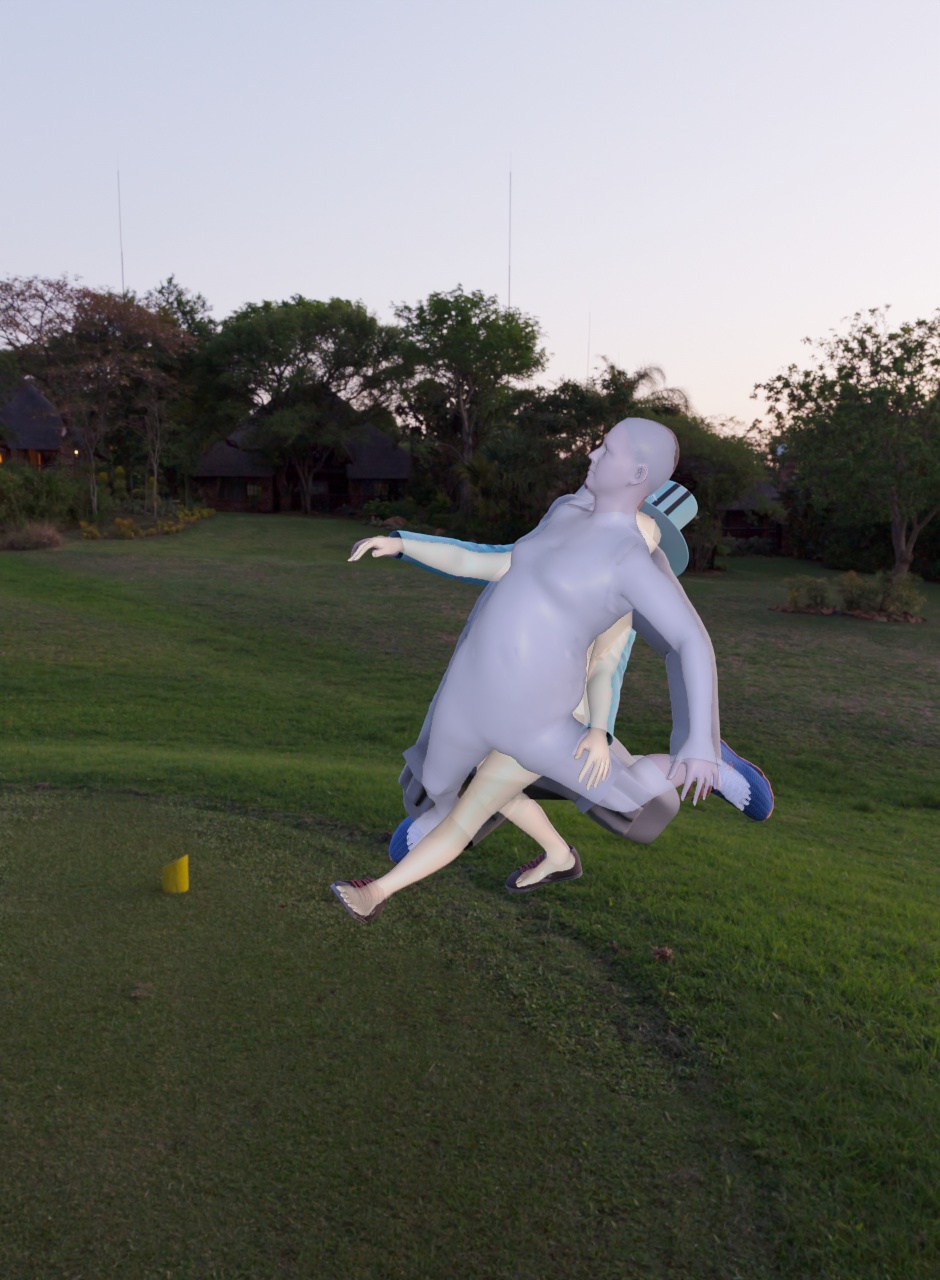} &
    \includegraphics[height=\dsh]{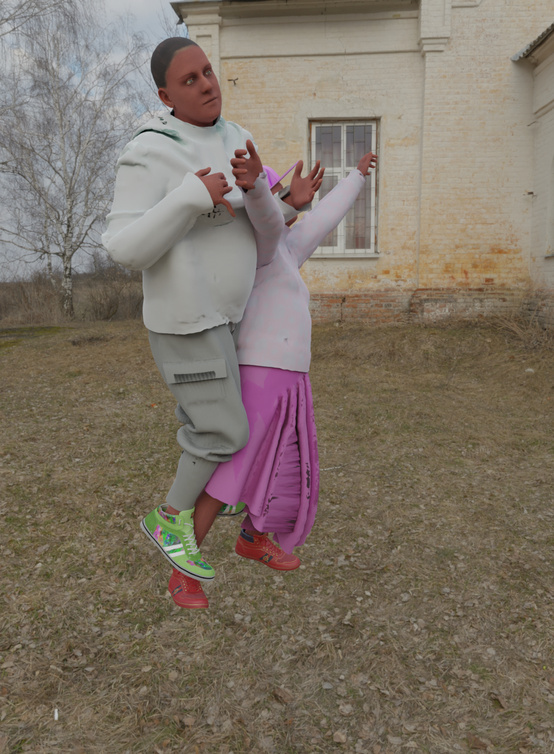} &
    \includegraphics[height=\dsh]{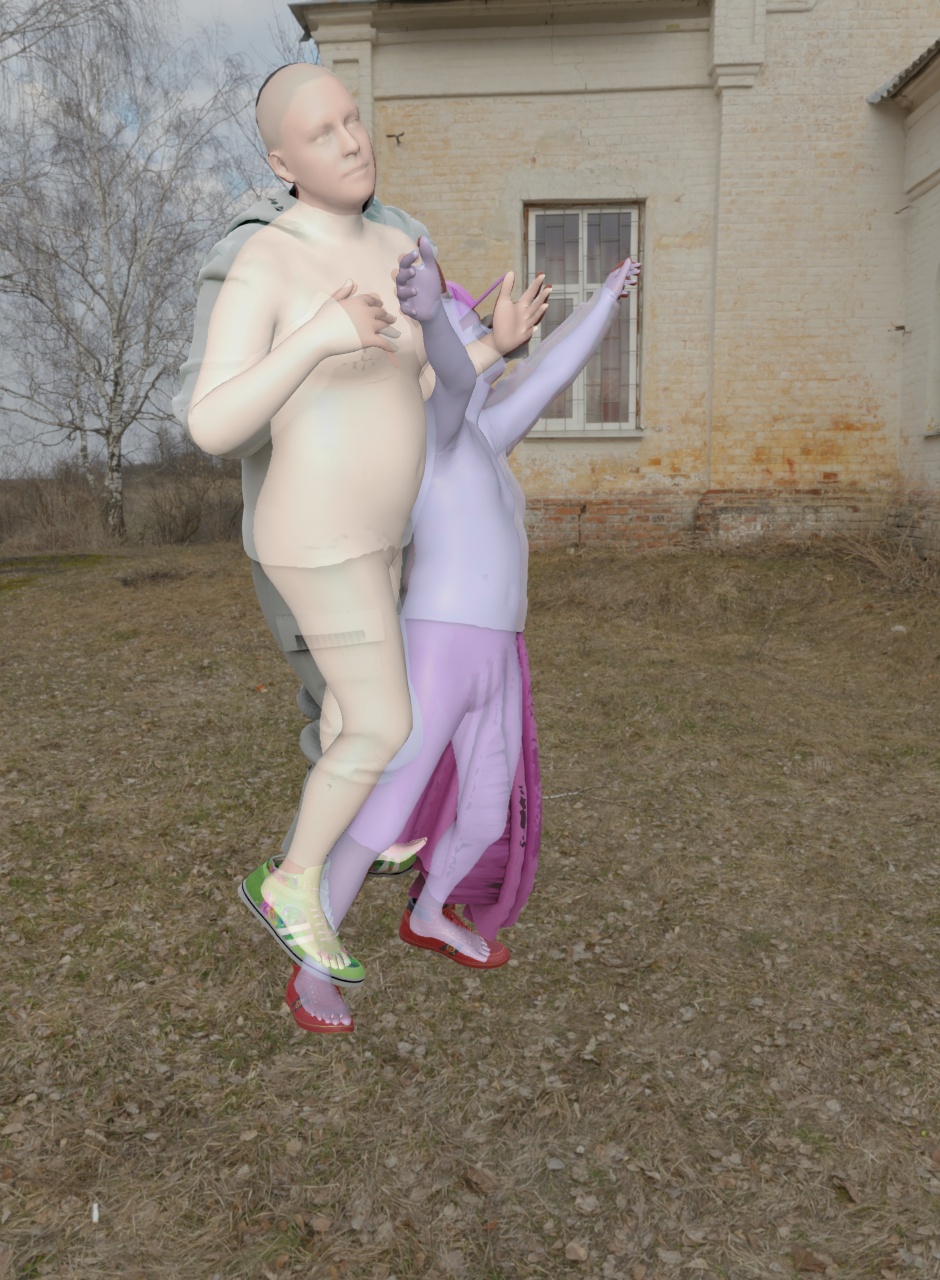} \\
  \end{tabular}

  \vspace{4pt}
  \hrule
  \vspace{6pt}

  {\small\textit{Three- and four-person extensions}}\\[4pt]
  \setlength{\dsh}{0.183\linewidth}
  \begin{tabular}{\dscolspec}
    \includegraphics[height=\dsh]{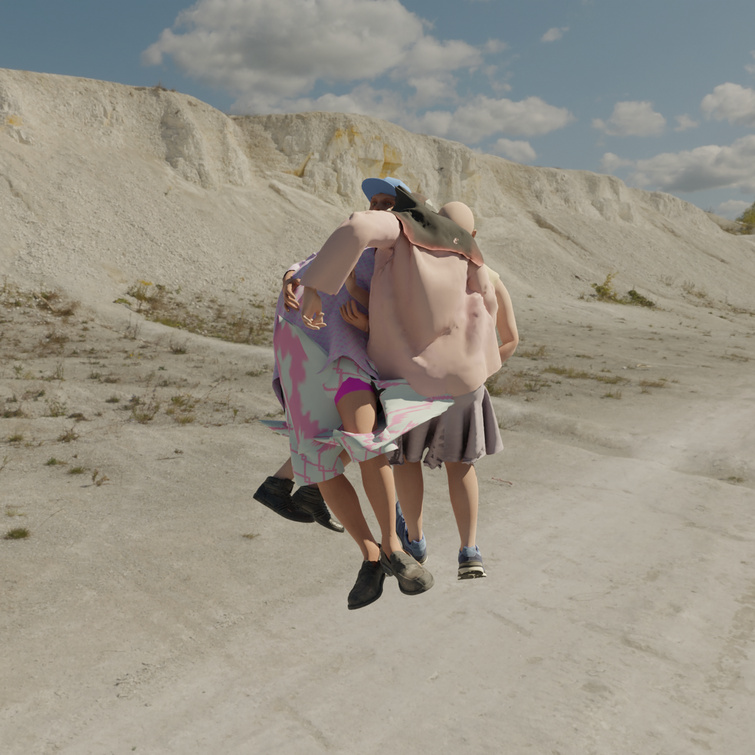} &
    \includegraphics[height=\dsh]{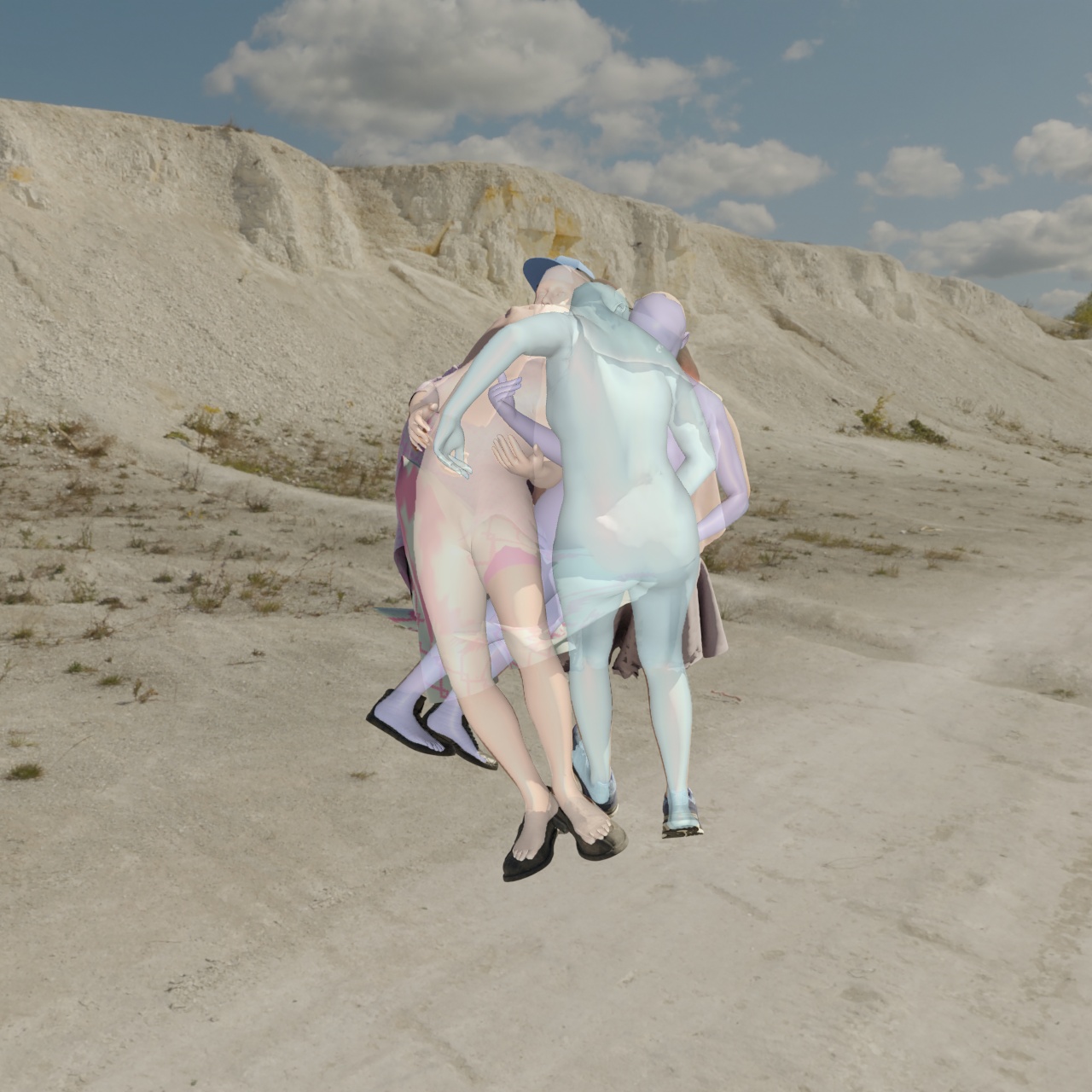} &
    \includegraphics[height=\dsh]{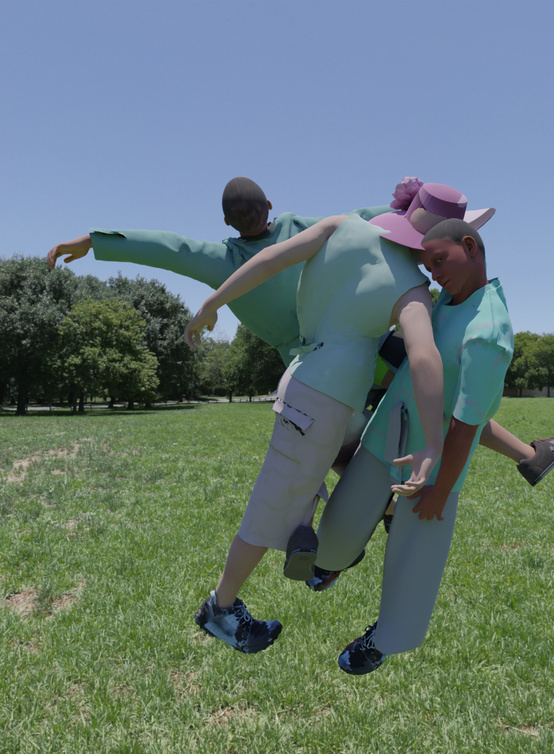} &
    \includegraphics[height=\dsh]{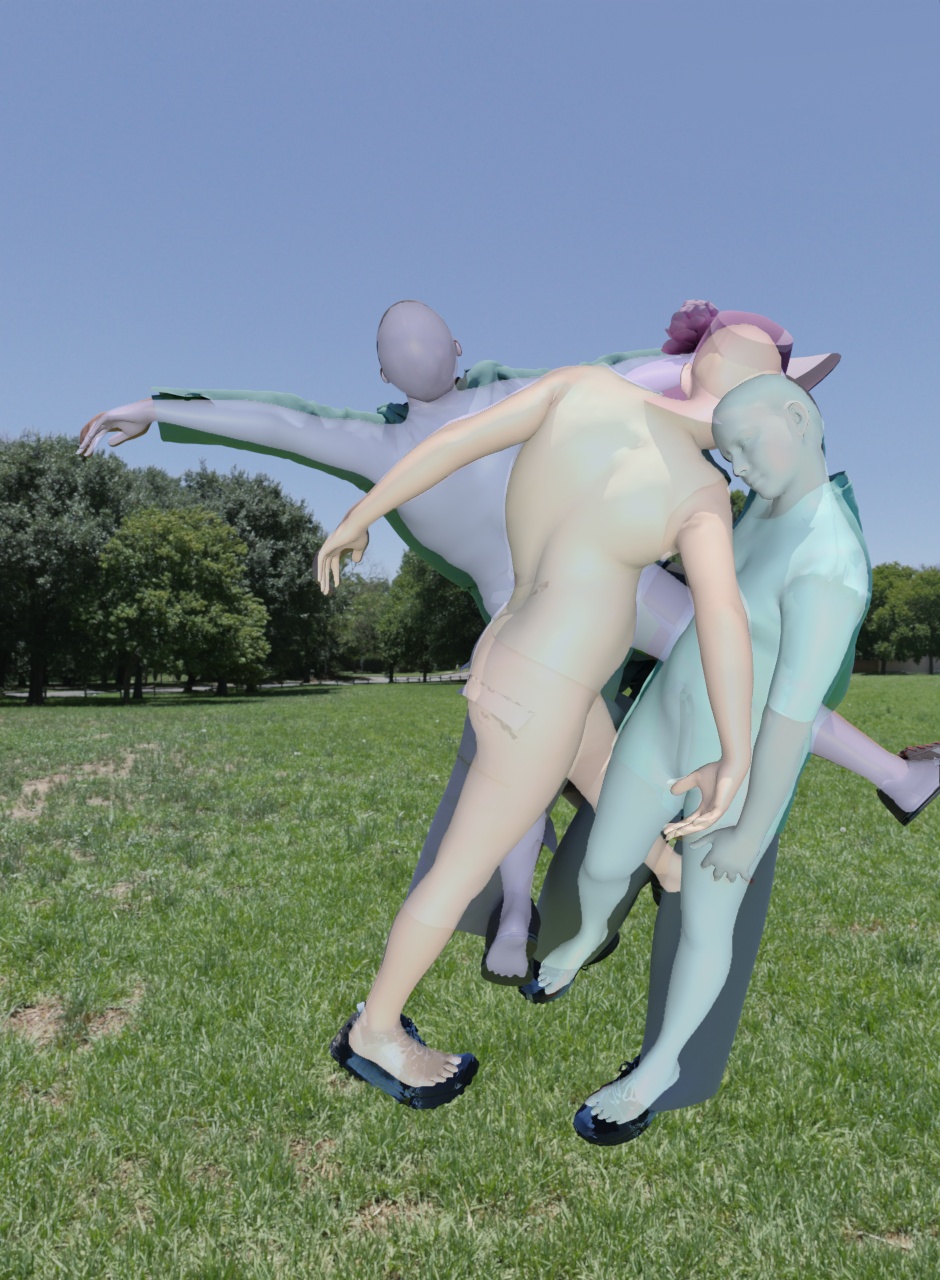} &
    \includegraphics[height=\dsh]{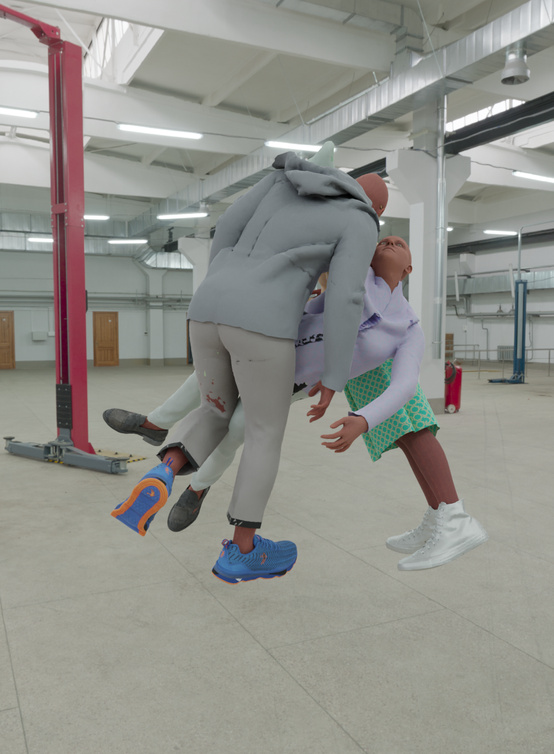} &
    \includegraphics[height=\dsh]{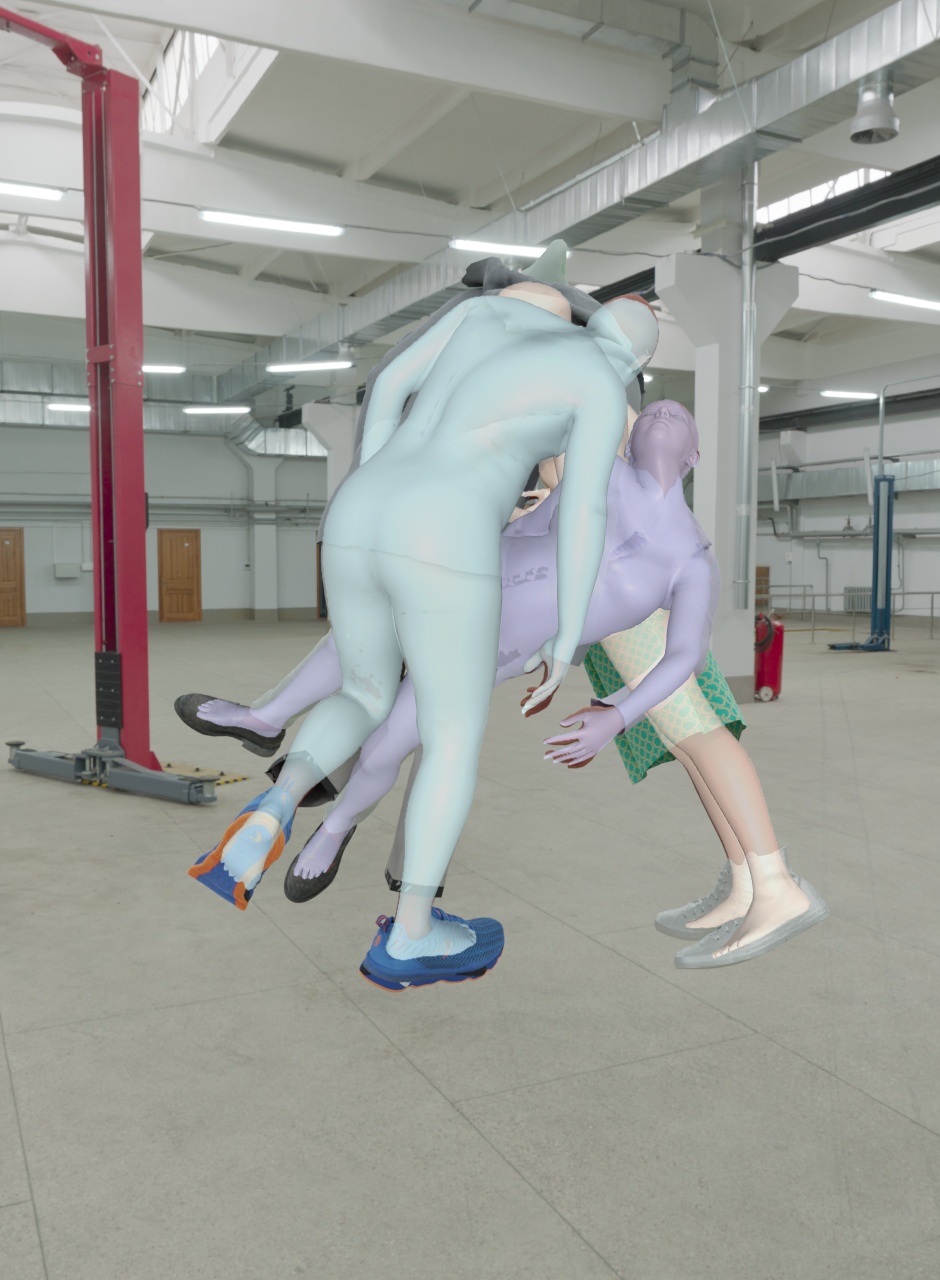} \\
  \end{tabular}\\[3pt]
  \setlength{\dsh}{0.165\linewidth}
  \begin{tabular}{\dscolspec}
    \includegraphics[height=\dsh]{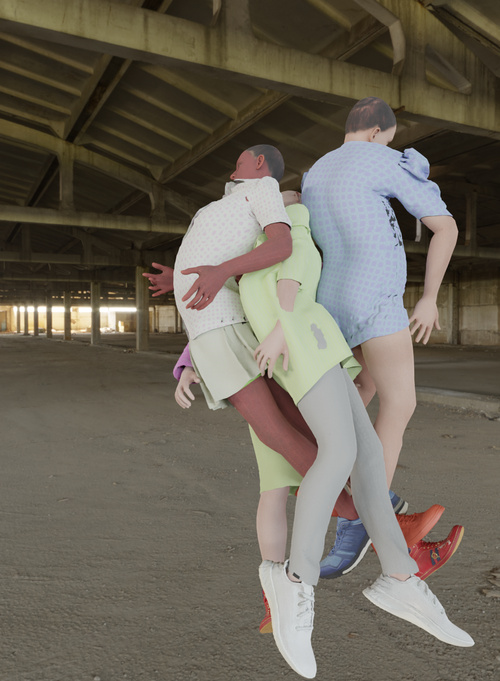} &
    \includegraphics[height=\dsh]{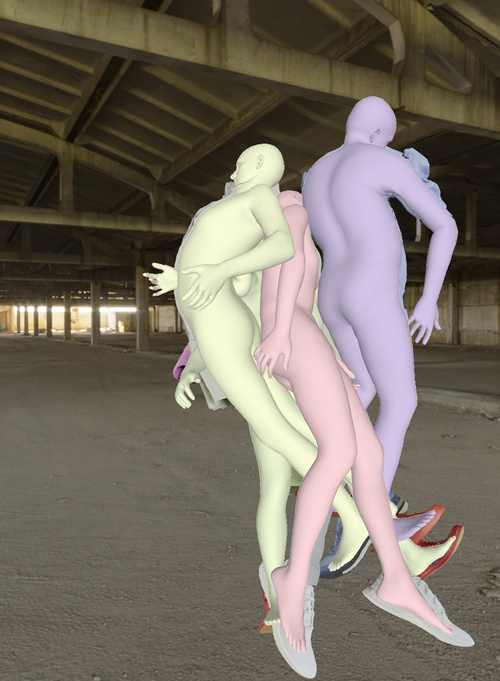} &
    \includegraphics[height=\dsh]{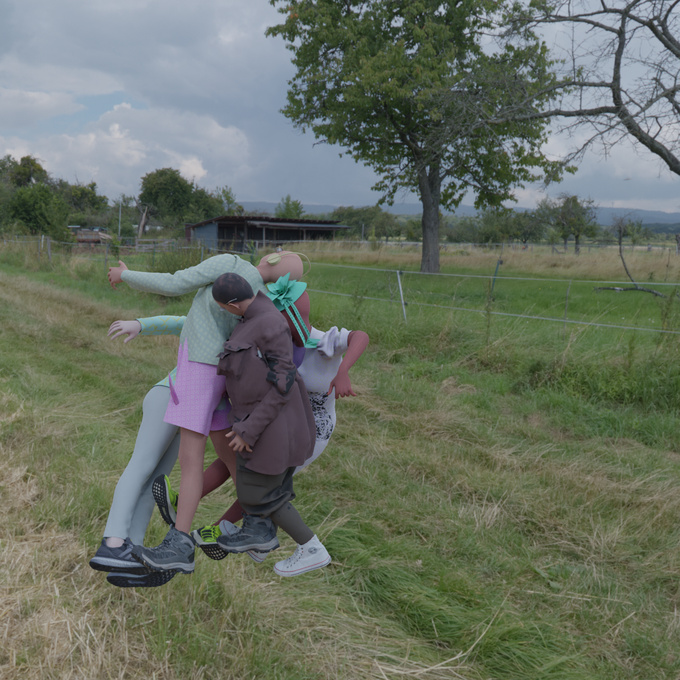} &
    \includegraphics[height=\dsh]{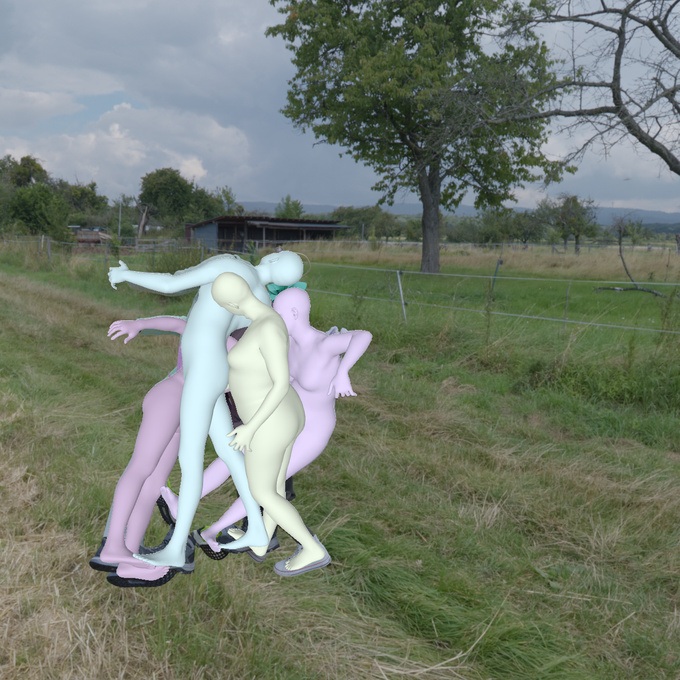} &
    \includegraphics[height=\dsh]{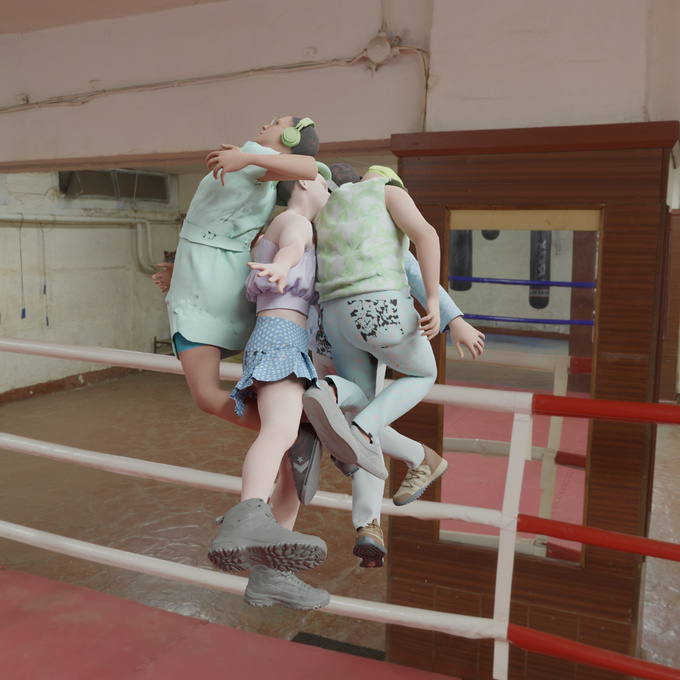} &
    \includegraphics[height=\dsh]{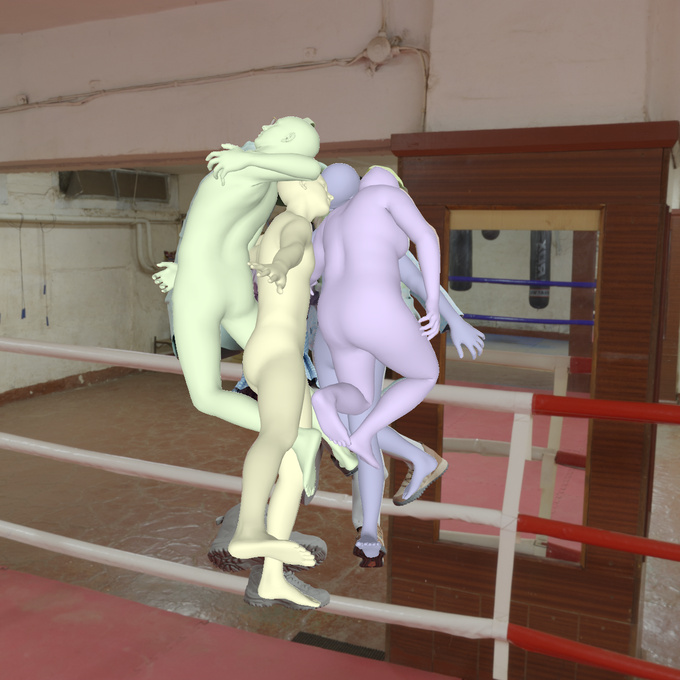} \\
  \end{tabular}

  \caption{%
    \textbf{Additional \cisyn{} dataset samples.}
    Each sample is shown as a pair: the rendered image (left) and the corresponding SMPL-X ground truth overlaid on it (right).
    \emph{Top}: six diverse two-person interaction sequences from the \cisyn{} dataset.
    \emph{Bottom}: three- and four-person sequences generated by the same pipeline with minimal modification, demonstrating scalability beyond two-person interactions.
  }
  \label{fig:supp_datasample}
\end{figure*}

\subsection{Physics Simulation Details}
\label{app:pipeline:sim}

\paragraph{Simulation parameters.}
We configure Genesis~\cite{2024genesis} with a fixed timestep $\mathrm{d}t = 1/480$\,s and standard gravity $g = 9.81$\,m/s$^2$. Rigid-body collision detection is enabled globally (up to 2000 collision pairs per scene); self-collision is selectively disabled for body-part pairs that would otherwise produce artifacts, detailed in the next paragraph. The simulator runs for 480 steps per sequence (1\,s of simulated time). The 12 exported frames start at frame 104 with stride 16 (i.e.\ frames 104, 120, \ldots, 280), spanning the collision window where contact is most pronounced. These 12 frames are saved in AMASS-compatible~\cite{mahmood2019amass} NPZ format at an effective frame rate of $480/16 = 30$\,FPS, resulting in a sequence duration of 0.4\,s. To facilitate inspection of the dataset, the 12 rendered frames of each sequence are compiled into an MP4 video at 5\,fps and released alongside the images.

\paragraph{SMPL-X Simulator Model.}
\label{app:pipeline:sim:model}
The actors are passive articulated rigid bodies (ragdoll physics) with no motion controller or reference trajectory; after launch, all motion is produced by gravity, joint limits, damping, and collision response.
Each humanoid is represented as an SMPL-X~\cite{pavlakos2019smplx} kinematic tree with 52 joints (1 root $+$ 21 body $+$ 15 per hand); facial joints (jaw, eyes) present in the full SMPL-X topology are excluded. Every joint is decomposed into three orthogonal 1-DOF Euler-angle hinges, yielding up to 156 DOFs per character. The model is constructed in the SMPL-X canonical T-pose (all pose parameters set to zero).
SMPL-X uses a right-handed coordinate system with $+y$ pointing upward (foot-to-head), $+z$ pointing forward (facing direction), and $+x$ pointing to the left; all joint angle limits in Table~\ref{tab:joint_limits} are defined in this frame.
Collision geometry is derived from the SMPL-X mesh: vertices are assigned to their highest-weight joint via skinning-weight argmax, a per-joint convex hull is computed from those vertices, and the hull is decimated to at most 50 vertices for simulation efficiency.
\textit{Joint limits} and \textit{damping} are manually specified: per-axis angle ranges are listed in Table~\ref{tab:joint_limits}; viscous damping coefficients are $0.25$--$5.0$\,N\,m\,s/rad for body joints and $0.125$--$0.2$\,N\,m\,s/rad for finger joints.
\textit{Self-collision} is enabled globally but disabled for five anatomically adjacent pairs that would otherwise produce spurious contact forces: Head$\leftrightarrow$Neck, L/R\_Shoulder$\leftrightarrow$L/R\_Thorax, and L/R\_Shoulder$\leftrightarrow$Chest.

\begin{table}[htbp]
\centering
\footnotesize
\setlength{\tabcolsep}{4pt}
\caption{Per-joint angle limits (degrees) for the SMPL-X simulator model.
Axes follow the SMPL-X right-handed frame: $+y$ up (foot-to-head), $+z$ forward (facing), $+x$ left.
Each joint is parameterized as three orthogonal Euler-angle hinges ($x$, $y$, $z$).
For brevity, $\pm\alpha$ denotes $[-\alpha,\,\alpha]$.
Left and right body joints share the same $x$ range; $y$ and $z$ are sign-flipped between sides.
Finger joints follow the same pattern: for right-hand non-thumb joints, replace $z\in[-f,\,5]$ with $z\in[-5,\,f]$;
for right-hand thumb joints, replace $y\in[-5,\,f]$ with $y\in[-f,\,5]$; proximal $x$ widens to $\pm20$ on the right.}
\label{tab:joint_limits}
\resizebox{\columnwidth}{!}{%
\begin{tabular}{llccc}
\toprule
Group & Joint & $x$ [°] & $y$ [°] & $z$ [°] \\
\midrule
\multicolumn{5}{l}{\textit{Body joints (left side shown; right mirrors $y$/$z$)}} \\
\midrule
Root     & Pelvis       & $\pm180$           & $\pm180$           & $\pm180$           \\
\midrule
\multirow{3}{*}{Spine}
         & Torso        & $[-25.8,57.3]$   & $\pm57.3$          & $\pm34.4$          \\
         & Spine        & $[-25.8,57.3]$   & $\pm57.3$          & $\pm34.4$          \\
         & Chest        & $[-25.8,57.3]$   & $\pm57.3$          & $\pm34.4$          \\
\midrule
\multirow{2}{*}{Head}
         & Neck         & $[-28.7,57.3]$   & $\pm40.1$          & $\pm40.1$          \\
         & Head         & $[-28.7,57.3]$   & $\pm40.1$          & $\pm40.1$          \\
\midrule
\multirow{2}{*}{Hip}
         & L Hip        & $[-160,22.9]$    & $\pm70$            & $[-28.7,90]$     \\
         & R Hip        & $[-160,22.9]$    & $\pm70$            & $[-90,28.7]$     \\
\midrule
\multirow{2}{*}{Knee}
         & L Knee       & $[-5.2,171.9]$   & $[-4.6,6.9]$     & $[-6.9,4.6]$     \\
         & R Knee       & $[-5.2,171.9]$   & $[-6.9,4.6]$     & $[-4.6,6.9]$     \\
\midrule
\multirow{2}{*}{Ankle}
         & L Ankle      & $\pm45.8$          & $[-74.5,22.9]$   & $\pm4.6$           \\
         & R Ankle      & $\pm45.8$          & $[-22.9,74.5]$   & $\pm4.6$           \\
\midrule
Toe      & L/R Toe      & $[-91.7,22.9]$   & $\pm4.6$           & $\pm4.6$           \\
\midrule
\multirow{2}{*}{Thorax}
         & L Thorax     & $[-10.9,4.6]$    & $[-28.7,17.2]$   & $[-31.5,40.1]$   \\
         & R Thorax     & $[-10.9,4.6]$    & $[-17.2,28.7]$   & $[-40.1,31.5]$   \\
\midrule
\multirow{2}{*}{Shoulder}
         & L Shoulder   & $\pm97.4$          & $[-108.9,80.2]$  & $[-97.4,85.9]$   \\
         & R Shoulder   & $\pm97.4$          & $[-80.2,108.9]$  & $[-85.9,97.4]$   \\
\midrule
\multirow{2}{*}{Elbow}
         & L Elbow      & $[-17.2,4.6]$    & $[-180,17.2]$    & $[-100.8,4.6]$   \\
         & R Elbow      & $[-17.2,4.6]$    & $[-17.2,180]$    & $[-4.6,100.8]$   \\
\midrule
Wrist    & L/R Wrist    & $[-180,22.9]$    & $\pm17.2$          & $\pm22.9$          \\
\midrule
\multicolumn{5}{l}{\textit{Finger joints (left hand shown; see caption for right-hand mapping)}} \\
\midrule
\multirow{3}{*}{Proximal}
         & Idx/Mid/Ring & $\pm5$          & $\pm20$            & $[-90,5]$        \\
         & Pinky             & $\pm5$          & $[-30,5]$        & $[-90,5]$        \\
         & Thumb             & $[-10,5]$     & $[-5,30]$        & $\pm5$             \\
\midrule
\multirow{3}{*}{Middle}
         & Idx/Mid/Ring & $\pm5$          & $\pm5$             & $[-100,5]$       \\
         & Pinky             & $\pm10$         & $\pm10$            & $[-100,5]$       \\
         & Thumb             & $\pm5$          & $[-5,80]$        & $\pm5$             \\
\midrule
\multirow{3}{*}{Distal}
         & Idx/Mid/Ring & $\pm5$          & $\pm5$             & $[-80,5]$        \\
         & Pinky             & $\pm10$         & $\pm10$            & $[-80,5]$        \\
         & Thumb             & $\pm5$          & $[-5,80]$        & $\pm5$             \\
\bottomrule
\end{tabular}}
\end{table}

\paragraph{SMPL-X pose to simulator DOF conversion.}
The Genesis humanoid is driven by Euler-angle DOFs, whereas SMPL-X~\cite{pavlakos2019smplx} poses (sampled from the AMASS~\cite{mahmood2019amass} library) are stored as axis-angle rotations. For each joint, we convert the axis-angle rotation via L-BFGS-B constrained optimization over the joint's allowed Euler range, minimizing the Frobenius norm between the reconstructed and target rotation matrices. Four random restarts are used to escape local minima introduced by the non-unique mapping between axis-angle and Euler representations.

\paragraph{SMPL annotations.}
SMPL labels are fitted to the SMPL-X meshes independently per frame, as suits single-frame HMR, so $\boldsymbol{\beta}$ may vary slightly within a sequence, whereas the SMPL-X $\boldsymbol{\beta}$ is constant per actor. For sequence-level tasks, the per-sequence mean of $\boldsymbol{\beta}$ can be used.

\subsection{Randomization Parameter Ranges}
\label{app:pipeline:sim:params}

Table~\ref{tab:sim_params} lists all randomization parameters for both generation batches. Body shape and initial pose are sampled identically across batches; the two regimes differ in orientation noise, collision speed, and angular velocity, which together control the intensity of the resulting contact.
Each batch contains 12{,}000 training sequences (144{,}000 frames). The 1{,}000-sequence test set is generated under the high-speed regime for more challenging evaluation. Its body assets, source poses, and HDRIs are not held out from training, so it serves only as an in-domain reference; all main-paper evaluations are cross-dataset.

\begin{table}[htbp]
  \centering
  \caption{Randomization parameter ranges for the two \cisyn{} generation batches. Parameters shared across batches are listed in a merged row. $\mathcal{U}(a, b)$ denotes a uniform distribution on $[a, b]$.}
  \label{tab:sim_params}
  \small
  \setlength{\tabcolsep}{4pt}
  \resizebox{\columnwidth}{!}{%
  \begin{tabular}{lll}
    \toprule
    Parameter & Standard & High-speed \\
    \midrule
    Body shape $\boldsymbol{\beta}$
      & \multicolumn{2}{l}{sampled w/o replacement from 1100 assets} \\
    Initial pose $\boldsymbol{\theta}$
      & \multicolumn{2}{l}{sampled w/o replacement from AMASS} \\
    \midrule
    Spawn radius / height
      & \multicolumn{2}{l}{$r{=}2.4$\,m, $h{=}1.5$\,m} \\
    Min.\ angular separation
      & \multicolumn{2}{l}{$48^\circ$} \\
    \midrule
    Yaw (root orientation)
      & \multicolumn{2}{l}{$\mathcal{U}(-180^\circ, 180^\circ)$} \\
    Pitch / roll noise $\sigma$
      & $10^\circ$ (clipped $\pm10^\circ$) & $20^\circ$ (clipped $\pm20^\circ$) \\
    \midrule
    Directional perturbation
      & \multicolumn{2}{l}{$\pm6^\circ$} \\
    Horizontal speed
      & $\mathcal{U}(5.5, 7.0)$\,m/s & $\mathcal{U}(7.5, 8.5)$\,m/s \\
    Vertical speed
      & \multicolumn{2}{l}{$\mathcal{U}(1.5, 2.5)$\,m/s} \\
    Angular velocity per axis
      & $\mathcal{U}(-1.0, 1.0)$\,rad/s & $\mathcal{U}(-3.0, 3.0)$\,rad/s \\
    \bottomrule
  \end{tabular}}
\end{table}

\subsection{Compute Cost and Scalability}
\label{app:pipeline:cost}
\begin{figure*}[t]
  \centering
  \includegraphics[width=\linewidth]{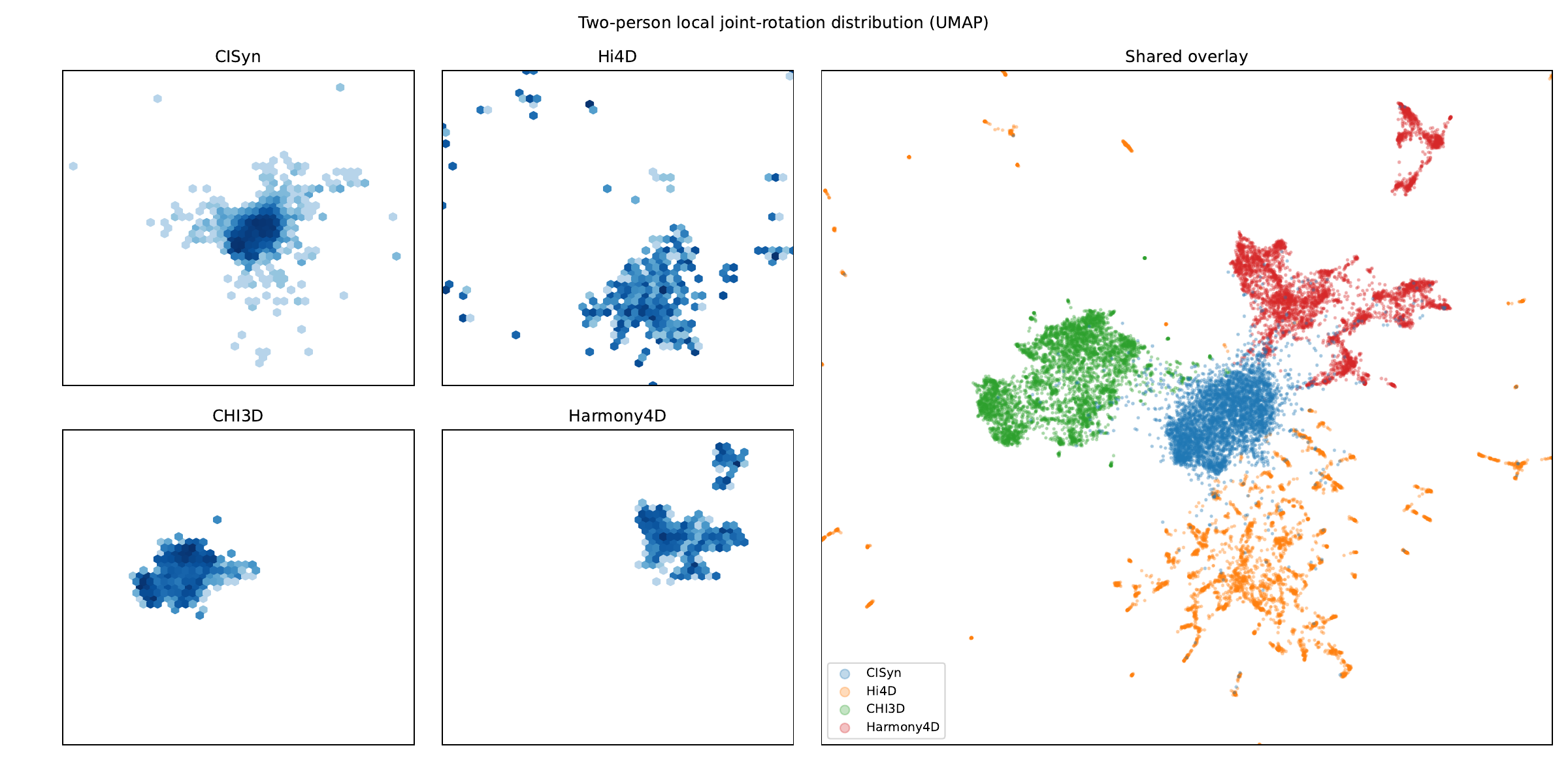}
  \caption{%
    \textbf{UMAP of the two-person pose distribution.}
    A shared embedding fitted on 5K pair descriptors each from \cisyn{}, Hi4D~\cite{yin2023hi4d}, CHI3D~\cite{fieraru2025chi3d}, and Harmony4D~\cite{khirodkar2024harmony4d}.
    \emph{Left}: per-dataset density. \emph{Right}: all four overlaid.
    The three real datasets do not completely overlap; \cisyn{} overlaps portions of the captured support and also covers additional regions.
  }
  \label{fig:pose_umap}
\end{figure*}

Physics simulation runs entirely on CPU; rendering requires a commodity GPU with $>$3\,GB of VRAM and runs on almost any modern GPU.
Each sequence takes approximately 150\,s end-to-end on a single GPU, with rendering dominating the cost.
Because sequences are independent, they can be trivially distributed across many GPUs: we generated the full \cisyn{} dataset in roughly 2 days using $\sim$20 GPUs in parallel.
The same pipeline can scale to more sequences or extend to interactions involving more than two people with minimal additional engineering effort. 
Figure~\ref{fig:supp_datasample} shows additional rendered samples from \cisyn{} together with their SMPL-X ground truth, including three- and four-person extensions generated by the same pipeline with minimal modification.

\section{Additional Dataset Analysis}
\label{app:dataset_analysis}

We further analyze \cisyn{} against the three real close-interaction datasets Hi4D~\cite{yin2023hi4d}, CHI3D~\cite{fieraru2025chi3d}, and Harmony4D~\cite{khirodkar2024harmony4d}, comparing two-person pose distributions (Sec.~\ref{app:dataset_analysis:pose}) and inter-person mesh penetration (Sec.~\ref{app:dataset_analysis:penetration}).
\cisyn{} shares substantial support with captured interactions while extending it toward uncommon collision configurations, and does not produce unusually severe penetration relative to captured datasets with fitted meshes.

\subsection{Pose Distribution}
\label{app:dataset_analysis:pose}

\begin{table}[htbp]
  \caption{%
    \textbf{Pose-distribution distances.}
    Nonparametric distances ($\downarrow$) between two-person pose distributions, computed on 10K unique frames per dataset (Sec.~\ref{app:dataset_analysis:pose}).
    Parentheses give the ratio to the median real-to-real distance.
    \emph{low-speed} is the 50\% subset of \cisyn{} generated with lower collision speeds, not a separate dataset.
  }
  \label{tab:pose_distribution}
  \centering
  \small
  \setlength{\tabcolsep}{3pt}
  \begin{tabular}{lcc}
    \toprule
    & Energy Distance$\downarrow$ & RBF-MMD$^2\downarrow$ \\
    \midrule
    \multicolumn{3}{@{}l}{\emph{Real vs.\ real}} \\
    Hi4D -- CHI3D          & 0.1918 & 0.0830 \\
    Hi4D -- Harmony4D      & 0.1395 & 0.0630 \\
    CHI3D -- Harmony4D     & 0.1677 & 0.0716 \\
    \midrule
    \multicolumn{3}{@{}l}{\emph{Synthetic vs.\ pooled real}} \\
    \cisyn{}               & 0.2350 (1.40$\times$) & 0.1101 (1.54$\times$) \\
    \cisyn{} (low-speed)   & \textbf{0.1559} (0.93$\times$) & \textbf{0.0724} (1.01$\times$) \\
    \bottomrule
  \end{tabular}
\end{table}

We sample 10K unique two-person frames from each dataset after removing duplicate camera views and limiting each sequence's contribution.
We represent each person using the local rotations of the 21 shared SMPL body joints in continuous 6D form, project them with a shared 32D PCA, and form a permutation-invariant pair descriptor.
Table~\ref{tab:pose_distribution} reports the energy distance and RBF-MMD$^2$, two nonparametric distribution distances, between all pairs of distributions; \emph{pooled real} is a balanced 10K-frame reference sampled from the three real datasets, rather than an average of the three real-to-real distances.
The low-speed subset is evaluated using the same frozen PCA, normalization, real reference, and MMD bandwidth as full \cisyn{}.
Full \cisyn{} has a moderate additional gap, consistent with its harder collisions, while the low-speed subset lies within the observed real-to-real ranges.

Figure~\ref{fig:pose_umap} shows a shared UMAP~\cite{mcinnes2018umap} embedding using 5K descriptors each from full \cisyn{} and the three real datasets.
The three real datasets do not completely overlap, indicating substantial real-to-real domain differences.
\cisyn{} overlaps portions of the captured support and also covers additional regions, showing that it shares substantial support with captured interactions rather than forming an unrelated pose cluster, while extending that support toward uncommon collision configurations.

\subsection{Inter-Person Penetration}
\label{app:dataset_analysis:penetration}

\begin{table}[htbp]
  \caption{%
    \textbf{Inter-person penetration.}
    Measured on the annotated gendered SMPL meshes of each dataset (Sec.~\ref{app:dataset_analysis:penetration}); lower is better.
    The mean depth is reported over all frames and, separately, conditional on penetration being present (\emph{w/ pen.}).
    \emph{Frames} is the fraction of frames with any penetration.
  }
  \label{tab:penetration}
  \centering
  \small
  \setlength{\tabcolsep}{3pt}
  \begin{tabular}{lcccc}
    \toprule
    \multirow{2}{*}{Dataset}
      & \multicolumn{2}{c}{Mean depth (mm)$\downarrow$} & Max & Frames \\
    \cmidrule(lr){2-3}
      & all frames & w/ pen. & (mm)$\downarrow$ & (\%)$\downarrow$ \\
    \midrule
    Hi4D~\cite{yin2023hi4d}                 & 3.99 & 8.06  & 81.64  & 49.56 \\
    CHI3D~\cite{fieraru2025chi3d}           & 2.63 & 9.62  & 133.59 & 27.37 \\
    Harmony4D~\cite{khirodkar2024harmony4d} & 3.89 & 12.18 & 134.82 & 31.97 \\
    \midrule
    \textbf{\cisyn{} (ours)}                & \textbf{1.36} & \textbf{5.41} & \textbf{78.05} & \textbf{25.12} \\
    \bottomrule
  \end{tabular}
\end{table}

We measure inter-person penetration on the annotated gendered SMPL meshes.
For each pair, sampled vertices of person~A are tested for inclusion inside person~B's full-resolution mesh using Kaolin's~\cite{kaolin} ray-based test, and then vice versa; distances from interior vertices to the other mesh are averaged across both directions.
In Table~\ref{tab:penetration}, the first mean assigns zero to frames without penetration and the second is conditional on penetration being present, while maximum depth is the largest point-to-mesh depth among all sampled penetrating vertices.
\cisyn{} has the lowest frequency, mean depth, and maximum depth, showing that its approximate collision geometry does not produce unusually severe inter-person penetration relative to captured datasets with fitted meshes.

\section{HMR Model Details}
\label{app:model}

\subsection{Architecture}

The encoder is a DINOv2~\cite{oquab2023dinov2} ViT-B/14 with embedding dimension 768; images are padded to 1288\,px resolution before encoding. The Scale-Adaptive Token (SAT) mechanism operates at three patch scales (14, 28, 56\,px). The scale classifier branches off at Block~3 with three additional transformer blocks and is supervised by the scale-map loss described in Sec.~\ref{app:model:losses}.
Concretely, the scale head predicts two values per patch: a presence logit $c\in\{0,1\}$ indicating whether the patch overlaps a person, and a normalized scale $s\in[0,1]$ representing the relative size of that person in the image.
These two outputs jointly determine the patch resolution used in the SAT mechanism: human patches with a small predicted scale ($c{=}1$, small $s$) are tokenized at 14\,px to capture fine detail for distant or small subjects; human patches with a large predicted scale ($c{=}1$, large $s$) use 28\,px tokens; and non-human background patches ($c{=}0$) are processed at the coarsest 56\,px resolution, reducing computation where pose information is absent.
The decoder is a 6-layer cross-attention transformer following the DAB-DETR~\cite{liu2022dabdetr} design with 50 learned queries; each layer uses $d_\text{model}=768$, 4 attention heads, a 2048-dim feed-forward network, and zero dropout. Table~\ref{tab:arch_hparams} lists the MLP specifications for all prediction heads.

\begin{table}[t]
  \centering
  \caption{Prediction head MLP specifications for \cisyn{}-HMR ($d{=}768$). Input features are described in the text.}
  \label{tab:arch_hparams}
  \small
  \setlength{\tabcolsep}{4pt}
  \begin{tabular}{ll}
    \toprule
    Head & Architecture \\
    \midrule
    \multicolumn{2}{l}{\textit{Encoder heads}} \\
    Scale Map            & MLP($d$, $d$, $2$; 4 layers), sigmoid \\
    FOV                  & MLP($d$, $d$, $1$; 4 layers) \\
    \midrule
    \multicolumn{2}{l}{\textit{Decoder heads (per layer)}} \\
    Pose (6D)            & MLP($d$, $d$, $144$; 6 layers) \\
    Shape                & MLP($d$, $d$, $11$; 5 layers) \\
    Translation          & MLP($d$, $d/2$, $3$; 3 layers) \\
    BBox                 & MLP($d$, $d$, $4$; 3 layers) \\
    Detection conf.      & Linear($d$, 1) \\
    Uncertainty $\sigma$ & MLP($d$, $d/2$, $45$; 3 layers), clamp$[-6,6]$ \\
    \bottomrule
  \end{tabular}
\end{table}

All MLP heads attached to the transformer decoder take the per-query $d$-dimensional decoder feature as input, except the FOV head and Scale Map head, which use the encoder's features (CLS token and per-patch tokens, respectively).
The shape head predicts 11 SMPL~\cite{2015smpl} coefficients ($\boldsymbol{\beta}\in\mathbb{R}^{10}$ plus one child-body offset); during training the loss applies per-coefficient weights $[2.56, 1.28, 0.64, 0.64, 0.32{\times}6, 10.0]$, giving the kid-offset dimension a substantially higher weight to encourage correct child-vs-adult classification.
The uncertainty head outputs 45 per-joint log-standard-deviations, one per output joint: the 24 standard SMPL~\cite{2015smpl} body joints (indices 0--23) concatenated with 21 extra regressor joints (indices 24--44) derived from mesh vertices via \texttt{J\_regressor\_extra.npy}, matching the joint set used to compute $\mathcal{L}_\text{j3d}$.
Its final layer is zero-initialized, so $\exp(\sigma_j){=}1$ at the start of training and the Laplacian NLL reduces to the occlusion-weighted L1 in the first gradient step. Outputs are clamped to $[-6, 6]$ to prevent numerical instability.

The pose, shape, and bounding-box heads are replicated independently for each decoder layer (weights are \emph{not} shared across layers). Bounding-box reference points, pose predictions, and shape predictions are iteratively refined layer-by-layer using the per-layer head outputs. Losses supervise the prediction at every layer, so the uncertainty head is likewise replicated and outputs an uncertainty for each layer.

\subsection{Model Size}

\cisyn{}-HMR has 226\,M parameters in total: the DINOv2 ViT-B/14 encoder contributes 106\,M, the 6-layer cross-attention decoder 73\,M, and the prediction heads plus embeddings the remaining 47\,M.
The uncertainty head introduced in this work adds only 2.8\,M parameters (six replicated MLPs, one per decoder layer), representing less than 1.3\% of the total model size.
The model is therefore a lightweight extension of the MA-HMR backbone: the entire overhead of our contribution is negligible relative to the inherited architecture.

\subsection{Loss Definitions}
\label{app:model:losses}

$\mathcal{L}_\text{j3d}$ and $\mathcal{L}_\text{j2d}$ are defined in Eq.~\ref{eq:uncertainty_loss} of the main text.
The remaining terms in Eq.~\ref{eq:total_loss} are inherited from
SAT-HMR~\cite{su2025sathmr} and MA-HMR~\cite{wang2025mahmr}; we reproduce
their definitions here for completeness.

\paragraph{Pose and shape losses (SAT-HMR).}
\begin{equation}
  \mathcal{L}_\text{pose} = \bigl\|\hat{\boldsymbol{\theta}} - \boldsymbol{\theta}\bigr\|_1,
  \qquad
  \mathcal{L}_\text{shape} = \sum_{k} v_k \bigl|\hat{\beta}_k - \beta_k\bigr|,
\end{equation}
where $\boldsymbol{\theta}\in\mathbb{R}^{24\times6}$ are the 6D-rotation pose parameters and
$v = [2.56, 1.28, 0.64, 0.64, 0.32{\times}6, 10.0]$ are per-coefficient shape weights
(the last entry targets the child-body offset of the extended SMPL model).

\paragraph{Bounding-box and detection losses (SAT-HMR).}
\begin{equation}
  \mathcal{L}_\text{box} = \bigl\|\hat{b} - b\bigr\|_1 + \bigl(1 - \mathrm{GIoU}(\hat{b}, b)\bigr),
\end{equation}
where $b = (c_x, c_y, w, h)$ is the normalized bounding box.
$\mathcal{L}_\text{detect}$ is a focal loss with $\alpha{=}0.25$,
$\gamma{=}2$ on per-query detection confidence scores.

\paragraph{Scale-map loss (SAT-HMR).}
Each image patch is assigned a presence label $c\in\{0,1\}$ and a
normalized scale $s = d_\text{bb}/\max(H,W)\in[0,1]$, where $d_\text{bb}$ is
the bounding-box diagonal of the overlapping person.
\begin{equation}
  \mathcal{L}_\text{scale} = \mathrm{FocalLoss}(\hat{c}, c)
  + \frac{1}{N_+}\sum_{c=1} |\hat{s} - s|,
\end{equation}
where $N_+$ is the number of positive patches.

\paragraph{Relative metric depth loss (MA-HMR).}
Let $\hat{z}_i$ and $z_i$ denote the predicted and ground-truth root depth of
person $i$. For each image, define the set of valid pairs
$\mathcal{P} = \{(i,j)\colon i<j,\;|z_i - z_j| \le 1\,\text{m}\}$.
\begin{equation}
  \mathcal{L}_\text{depth} = \frac{1}{|\mathcal{P}|}\sum_{(i,j)\in\mathcal{P}}
  \log\!\bigl(1 + \bigl|(\hat{z}_i - \hat{z}_j) - (z_i - z_j)\bigr|\bigr).
\end{equation}
The 1\,m threshold discards pairs that are far apart in depth, where exact
relative ordering is less important for appearance-based disambiguation.
The log penalty provides stable gradients across a wide range of depth errors.

\paragraph{Field-of-view loss (MA-HMR).}
Let $\phi$ denote the predicted vertical FOV and $\phi^*$ the ground-truth
value (both in radians).
\begin{equation}
  \mathcal{L}_\text{fov} = \sum_i w_i\,(\phi_i - \phi^*_i)^2,
  \quad
  w_i = \begin{cases} 3 & \phi_i > \phi^*_i \\ 1 & \text{otherwise,}\end{cases}
\end{equation}
where the asymmetric weight penalizes FOV over-estimation more heavily, since
over-estimating FOV compresses the predicted metric scale.

\subsection{Loss Weights}
\label{app:model:lossweights}
All weights are taken unchanged from MA-HMR~\cite{wang2025mahmr}; we tuned none of them.
Table~\ref{tab:loss_weights} reports the scalar weights $\lambda_{(\cdot)}$
in Eq.~\ref{eq:total_loss}. Auxiliary decoder losses (one set per intermediate
decoder layer) use the same weights. The Hungarian matching costs are:
confidence~1, bounding-box L1~1, GIoU~1, keypoint L1~20.

\begin{table}[ht]
  \centering
  \caption{Loss weights $\lambda_{(\cdot)}$ for \cisyn{}-HMR training.}
  \label{tab:loss_weights}
  \small
  \setlength{\tabcolsep}{4pt}
  \begin{tabular}{lll}
    \toprule
    Symbol & Loss & Weight \\
    \midrule
    $\lambda_\text{j3d}$    & 3D joint Laplacian NLL (root-aligned) & 8.0  \\
    $\lambda_\text{j2d}$    & 2D joint Laplacian NLL (projected)    & 40.0 \\
    $\lambda_\text{pose}$   & SMPL 6D pose L1                       & 5.0  \\
    $\lambda_\text{shape}$  & SMPL shape L1 (per-coeff.\ weighted)  & 3.0  \\
    $\lambda_\text{box}$    & Bounding-box L1 + GIoU                 & 2.0  \\
    $\lambda_\text{detect}$ & Detection confidence (focal loss)     & 4.0  \\
    $\lambda_\text{depth}$  & Relative metric depth                 & 0.5  \\
    $\lambda_\text{fov}$    & Field-of-view                         & 0.5  \\
    $\lambda_\text{scale}$  & SAT scale-map (conf.\ + scale)         & 4.0  \\
    \bottomrule
  \end{tabular}
\end{table}

\section{Experiment Details}
\label{app:exp}

\subsection{Training Details for \cisyn{}-HMR}

\paragraph{Initialization.}
\cisyn{}-HMR is initialized from MA-HMR's~\cite{wang2025mahmr} final checkpoint, except for the uncertainty head $\sigma$, which is freshly added with zero-initialized weights so that the Laplacian NLL reduces to the standard occlusion-weighted L1 at the start of training.

\paragraph{Data augmentation.}
Each training image undergoes random in-plane rotation ($\pm15^\circ$), scale jitter ($[0.8, 1.8]$), horizontal flip (50\%), and a random crop retaining at least 80\% of the image content.

\paragraph{Datasets.}
All experiments use Base, the three datasets on which MA-HMR~\cite{wang2025mahmr} is trained: AGORA~\cite{patel2021agora} ($\sim$14K frames), BEDLAM~\cite{black2023bedlam} ($\sim$286K frames at 6\,fps), and DTO-Humans~\cite{wang2025mahmr} ($\sim$558K frames, comprising COCO 8K~\cite{lin2014coco}, MPII 1.4K~\cite{andriluka2014mpii}, AI-Challenger 70K~\cite{wu2017aic}, and InstaVariety 478K~\cite{kanazawa2019insta}), optionally augmented with one close-interaction training set. At each epoch, samples from all active datasets are mixed and shuffled uniformly before batching. The \cisyn{} training split contains $\sim$288K frames; the comparison interaction sets in Table~\ref{tab:dataset_results} are Hi4D~\cite{yin2023hi4d} ($\sim$80K training frames; $\sim$15K test frames), CHI3D~\cite{fieraru2025chi3d} ($\sim$259K training frames; the test split is obtained by 16$\times$ temporal downsampling of the training sequences, yielding $\sim$16K frames), and Harmony4D~\cite{khirodkar2024harmony4d} ($\sim$311K training frames after 4$\times$ temporal downsampling; $\sim$18K test frames after 16$\times$ temporal downsampling), each of which also serves as an evaluation benchmark for the rows that do not train on it. The motion-source ablation in Table~\ref{tab:ablation} additionally uses a non-simulated variant rendered from all 1{,}925 released DualHuman~\cite{fang2024dualhuman} two-person sequences (every 4th frame, $\sim$92K frames), with body shapes matched to our assets, through the same render pipeline (Sec.~\ref{sec:method:dataset:rendering}).

\paragraph{Optimizer and schedule.}
We use AdamW with learning rate $10^{-5}$ applied uniformly to all parameters (including the ViT-B/14 encoder), weight decay $10^{-4}$, and a cosine learning-rate schedule with no warm-up, training for 6 epochs at a batch size of 16 per GPU.

\paragraph{Training computation.}
All experiments are run on 4 GPUs with 48\,GB VRAM each (NVIDIA A6000, A40, or L40S), giving a total batch size of 64. Wall-clock training time is approximately 3 days on A6000 and A40 hardware, and approximately 2 days on L40S.

\subsection{Evaluation Metrics Details}

\textbf{Instance matching.}
Because all methods produce an unordered set of detections, predictions must first be associated with ground-truth persons before any per-person error is measured.
We follow the standard greedy 2D keypoint matching used in~\cite{wang2025prompthmr,wang2025mahmr}: for each image, predicted and ground-truth instances are matched greedily by 2D keypoint bounding-box IoU (threshold 0.1). All three metrics below are then computed over the matched pairs. Following~\cite{wang2025mahmr}, ground-truth persons with no matching prediction are treated as missed and excluded from the MPJPE/PA-MPJPE/PVE averages.

All three metrics are computed in a root-relative frame.
Following PromptHMR~\cite{wang2025prompthmr}, the pelvis is defined as the midpoint of the left and right hip joints in the H36M skeleton (indices 1 and 4), which are obtained by applying a linear H36M regressor to the predicted SMPL mesh vertices.

\textbf{MPJPE.}
The model predicts SMPL parameters, from which 3D joint positions are derived via the H36M regressor applied to the SMPL mesh.
Let $\hat{\mathbf{j}}_{i,k}\in\mathbb{R}^3$ and $\mathbf{j}_{i,k}\in\mathbb{R}^3$ denote the predicted and ground-truth pelvis-relative 3D positions of joint $k$ for person $i$.
\begin{equation}
  \mathrm{MPJPE} = \frac{1}{NJ}\sum_{i=1}^{N}\sum_{k=1}^{J}
  \bigl\|\hat{\mathbf{j}}_{i,k} - \mathbf{j}_{i,k}\bigr\|_2,
\end{equation}
where $J{=}14$ joints from the H36M skeleton (indices 1--6, 8, 10--16) are used for evaluation and results are reported in millimeters.

\textbf{PA-MPJPE.}
Before computing the per-joint error, the predicted joint set $\hat{\mathbf{J}}_i\in\mathbb{R}^{J\times3}$ is rigidly aligned to the ground-truth joint set $\mathbf{J}_i\in\mathbb{R}^{J\times3}$ via Procrustes analysis (rotation, translation, and isotropic scaling), yielding the aligned prediction $\tilde{\mathbf{J}}_i = \mathcal{P}(\hat{\mathbf{J}}_i,\mathbf{J}_i)\in\mathbb{R}^{J\times3}$:
\begin{equation}
  \mathrm{PA\text{-}MPJPE} = \frac{1}{NJ}\sum_{i=1}^{N}\sum_{k=1}^{J}
  \bigl\|\tilde{\mathbf{j}}_{i,k} - \mathbf{j}_{i,k}\bigr\|_2,
\end{equation}
where $\tilde{\mathbf{j}}_{i,k}$ is the $k$-th row of $\tilde{\mathbf{J}}_i$. Because of this alignment, which is not available at deployment time, PA-MPJPE measures pose accuracy independently of global positioning.

\textbf{PVE.}
Per-Vertex Error is computed over all $V{=}6890$ SMPL mesh vertices rather than the sparse joint set:
\begin{equation}
  \mathrm{PVE} = \frac{1}{NV}\sum_{i=1}^{N}\sum_{v=1}^{V}
  \bigl\|\hat{\mathbf{v}}_{i,v} - \mathbf{v}_{i,v}\bigr\|_2,
\end{equation}
where $\hat{\mathbf{v}}_{i,v}$ and $\mathbf{v}_{i,v}$ are the predicted and ground-truth vertex positions in the pelvis-relative frame.

\section{Additional Results}
\label{app:results}
\begin{table*}[t]
  \caption{
    \textbf{Quantitative results on the \cisyn{} test set and ExPI.}
    All metrics in mm; lower is better; \textbf{bold} marks the best value per column within each block.
    Top block: dataset-comparison rows from Table~\ref{tab:dataset_results}.
    Bottom block: settings follow Table~\ref{tab:main_results}.
    \textsuperscript{\textdagger}ExPI numbers are a proxy due to skeleton
    mismatch.
    \textsuperscript{\ddag}Trained on the \cisyn{} training split, so these \cisyn{} test-set
    numbers are in-domain and are listed for reference only.
  }
  \label{tab:quantitative}
  \centering
  \small
  \begin{tabular}{lcccccc}
    \toprule
    \multirow{2}{*}{Method}
      & \multicolumn{3}{c}{\cisyn{} test set}
      & \multicolumn{2}{c}{ExPI\textsuperscript{\textdagger}} \\
    \cmidrule(lr){2-4}\cmidrule(lr){5-6}
      & MPJPE$\downarrow$ & PA-MPJPE$\downarrow$ & PVE$\downarrow$
      & MPJPE$\downarrow$ & PA-MPJPE$\downarrow$ \\
    \midrule
    \multicolumn{6}{l}{\emph{\cisyn{}-HMR \quad (Base = AGORA + BEDLAM + DTO-Humans)}} \\
    Base+Hi4D                            & 265.5          & 162.8          & 286.5          & 114.5          & 80.7          \\
    Base+CHI3D                           & 260.0          & 160.3          & 280.6          & 111.6          & 80.9          \\
    Base+Harmony4D                       & 298.5          & 171.2          & 322.4          & 113.6          & 82.0          \\
    Base+\cisyn{}                        & \textbf{133.2}\textsuperscript{\ddag} & \textbf{103.2}\textsuperscript{\ddag} & \textbf{149.8}\textsuperscript{\ddag} & \textbf{110.3} & \textbf{80.2} \\
    \midrule
    MA-HMR (Base)   & 270.4          & 163.5          & 292.8          & \textbf{113.5} & \textbf{81.5} \\
    \cisyn{}-HMR (Base)                  & \textbf{264.2} & \textbf{161.4} & \textbf{285.9} & 113.9          & 82.5          \\
    \bottomrule
  \end{tabular}
\end{table*}

\begin{table*}[t]
  \caption{%
    \textbf{Per-category comparison of close-interaction training datasets.}
    Table~\ref{tab:dataset_results} broken down by interaction category, with ExPI~\cite{guo2022expi} added; settings follow that table and ``--'' again marks the in-domain split excluded from evaluation.
    All values are MPJPE in mm.
    Raw labels are unified by semantics: \emph{Casual} = pose/posing/talk; \emph{Combat} = fight/grappling/hit/kick/MMA/push/sword; \emph{Social} = backhug/cheers/grab/handshake/high-five/holding hands/hug/kiss/sidehug; \emph{Dance} = ballroom/dance; \emph{Sports} = basketball; \emph{Acrobatics} covers all ExPI actions.
    \textsuperscript{\textdagger}ExPI numbers are a proxy due to skeleton mismatch (Sec.~\ref{app:results:quantitative}).
  }
  \label{tab:percate}
  \centering
  \small
  \setlength{\tabcolsep}{3pt}
  \begin{tabular}{lccccccccccccc}
    \toprule
    \multirow{2}{*}{Training data}
      & \multicolumn{5}{c}{Hi4D}
      & \multicolumn{3}{c}{CHI3D}
      & \multicolumn{1}{c}{ExPI\textsuperscript{\textdagger}}
      & \multicolumn{3}{c}{Harmony4D} \\
    \cmidrule(lr){2-6} \cmidrule(lr){7-9} \cmidrule(lr){10-10} \cmidrule(lr){11-13}
      & Casual & Combat & Dance & Social & Sports
      & Casual & Combat & Social
      & Acrobatics
      & Combat & Dance & Social \\
    \midrule
    Base+Hi4D      & --   & --   & --   & --   & --   & 56.0 & 46.4 & 47.1 & 114.5 & 81.4 & 72.8 & 65.7 \\
    Base+CHI3D     & 61.7 & 55.8 & 52.7 & 58.7 & 46.9 & --   & --   & --   & 111.6 & 81.5 & 73.4 & 65.9 \\
    Base+Harmony4D & 64.2 & 56.7 & 53.4 & 60.3 & 47.8 & 56.5 & 46.6 & 47.9 & 113.6 & --   & --   & --   \\
    Base+\cisyn{}  & \textbf{56.3} & \textbf{51.1} & \textbf{47.5} & \textbf{52.6} & \textbf{45.3} & \textbf{55.5} & \textbf{45.6} & \textbf{46.8} & \textbf{110.3} & \textbf{80.3} & \textbf{71.3} & \textbf{64.5} \\
    \bottomrule
  \end{tabular}
\end{table*}

\subsection{Quantitative Results on the \cisyn{} Test Set and ExPI}
\label{app:results:quantitative}

Table~\ref{tab:quantitative} extends Tables~\ref{tab:dataset_results} and~\ref{tab:main_results}
to two additional benchmarks; method settings follow those tables exactly.

\paragraph{\cisyn{} test set.}
Errors are much higher than on Hi4D, CHI3D and Harmony4D for every model not trained on
\cisyn{}, reflecting the difficulty of the physics-simulated contact scenarios.
Measured against the base-only model in the last row, adding any of the three real
close-interaction datasets does not close this gap, whereas \cisyn{} transfers to all three
of them (Table~\ref{tab:dataset_results}).
Our uncertainty-aware extensions give a small but consistent improvement over the
MA-HMR~\cite{wang2025mahmr} baseline under identical training data.
The \cisyn{}-trained row is in-domain and is listed for reference only.

\paragraph{ExPI.}
ExPI~\cite{guo2022expi} provides annotations in an 18-joint skeleton that does not
align with the SMPL or H36M joint set; we match each ExPI joint to the most
anatomically corresponding joint and report the result as a proxy metric,
which should be interpreted cautiously given this mismatch.
Unlike on \cisyn{}, the uncertainty-aware extensions yield no clear gain on ExPI, and none
of the three real close-interaction datasets helps as much as \cisyn{}.
Comparing the last row of each block, adding \cisyn{} improves both metrics, consistent
with Table~\ref{tab:main_results}.

\subsection{Per-Category Results}
\label{app:results:percate}

Table~\ref{tab:percate} breaks the dataset comparison of Table~\ref{tab:dataset_results} down by interaction category and adds ExPI~\cite{guo2022expi}.
Adding \cisyn{} gives the lowest error in every category.
The four benchmarks together cover social contact, combat, sports, casual interaction, dance, and acrobatics, indicating that the benefit is not confined to a single dataset or interaction type.
The gain also extends to goal-directed categories such as dance and casual interaction, whose motion the collisions in \cisyn{} do not imitate.
Two properties of the setting explain this.
First, single-image HMR is supervised per frame, so the model never observes a trajectory and learns from the distribution of two-body configurations rather than from the motion that produced them.
Second, collisions sample that configuration space broadly, so the support of \cisyn{} covers the configurations these categories contain (Sec.~\ref{app:dataset_analysis:pose}), leaving the model to interpolate within it.

\subsection{Detection Recall and Inter-Person Accuracy}
\label{app:results:pair}
\begin{figure*}[p]
  \centering
  \setlength{\tabcolsep}{1pt}
  \renewcommand{\arraystretch}{1.0}
  \newcommand{\sqw}{0.1875\linewidth}
  \newcommand{\sqhead}[1]{\scriptsize #1}
  \newcommand{\sqrowlabel}[1]{\rotatebox{90}{\scriptsize #1}}
  \newcommand{\sqna}[1]{\raisebox{#1\linewidth}{\textcolor{black!40}{\scriptsize n/a}}}
  \begin{tabular}{@{}c@{\hspace{2pt}}ccccc@{}}
    & \sqhead{Input} & \sqhead{GT} & \sqhead{PromptHMR~\cite{wang2025prompthmr}} & \sqhead{MA-HMR~\cite{wang2025mahmr}} & \sqhead{\cisyn{}-HMR (Ours)} \\
    \sqrowlabel{Hi4D}      & \includegraphics[width=\sqw]{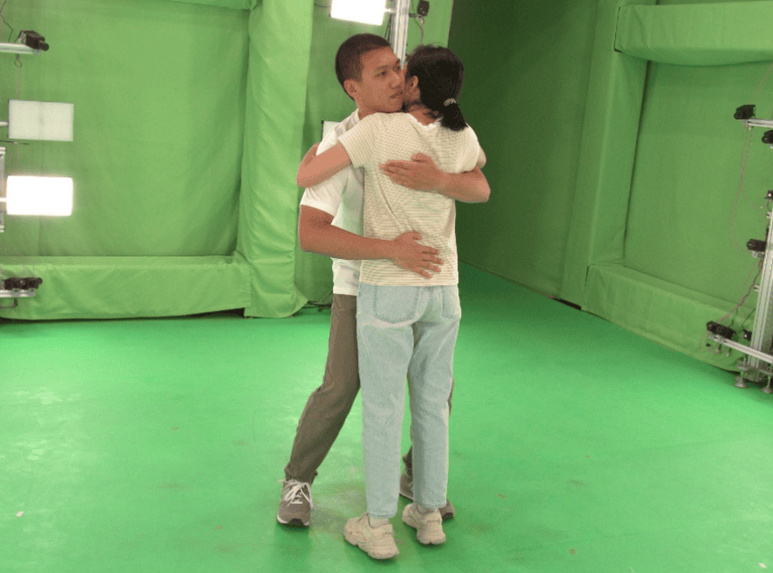}      & \includegraphics[width=\sqw]{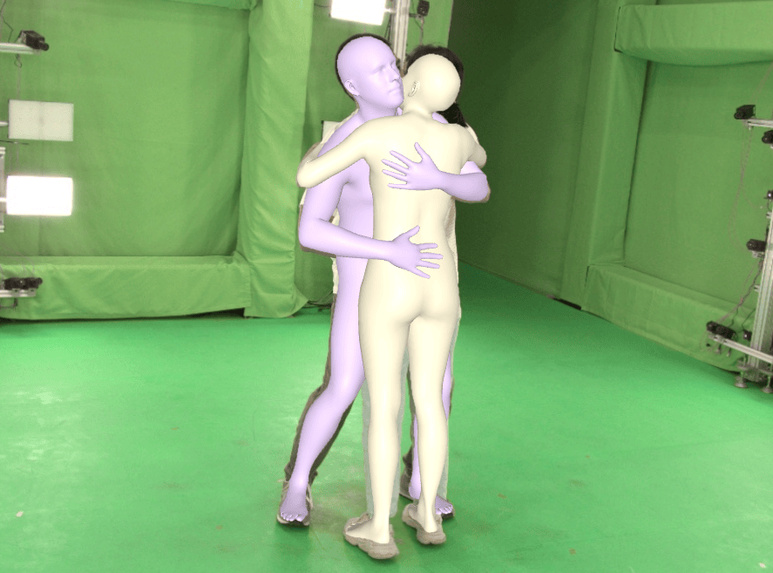}      & \sqna{0.066}            & \includegraphics[width=\sqw]{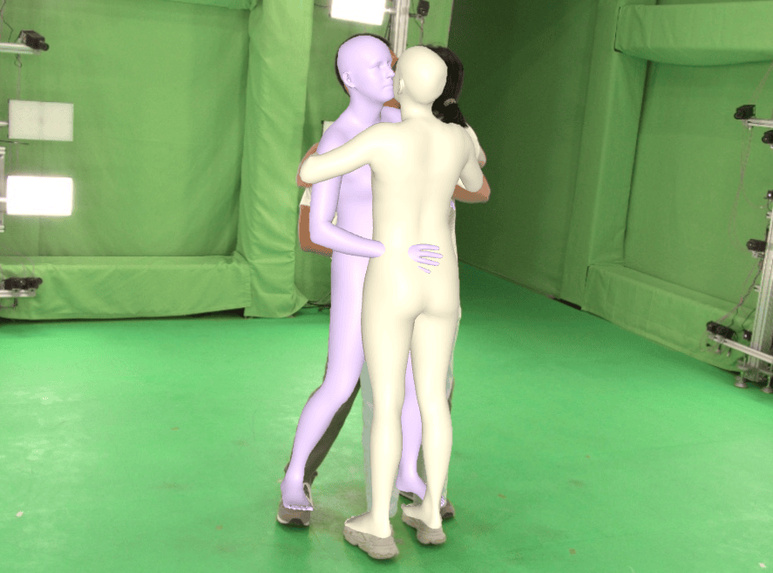}      & \includegraphics[width=\sqw]{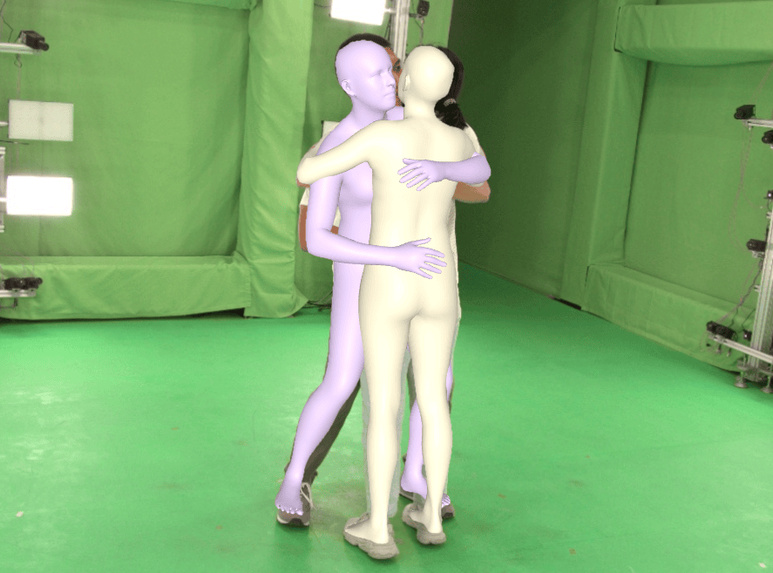}      \\
    \sqrowlabel{Hi4D}      & \includegraphics[width=\sqw]{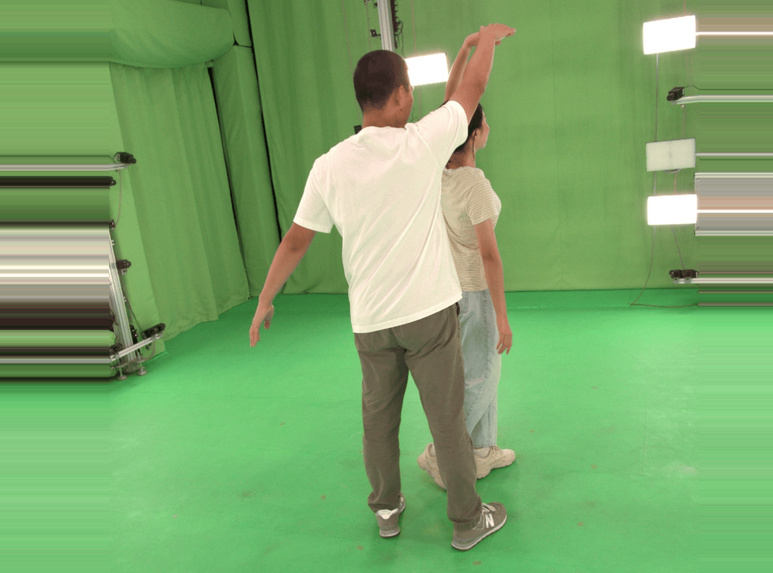}     & \includegraphics[width=\sqw]{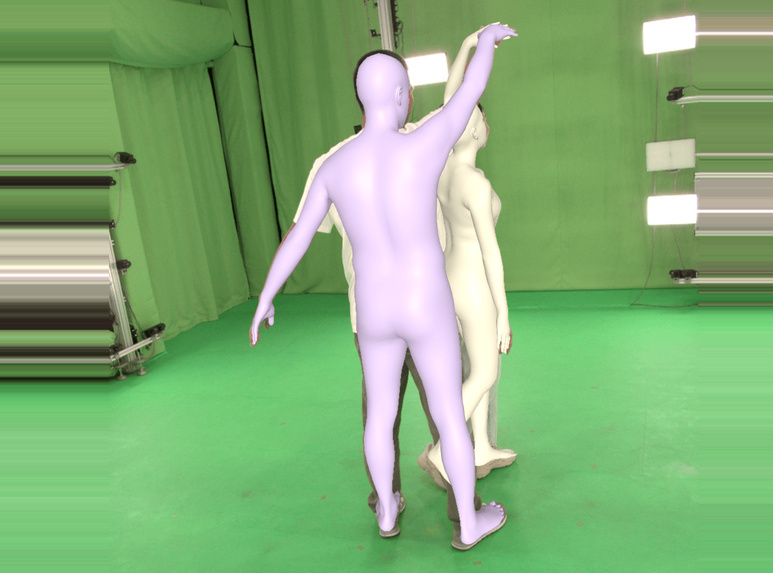}     & \sqna{0.066}            & \includegraphics[width=\sqw]{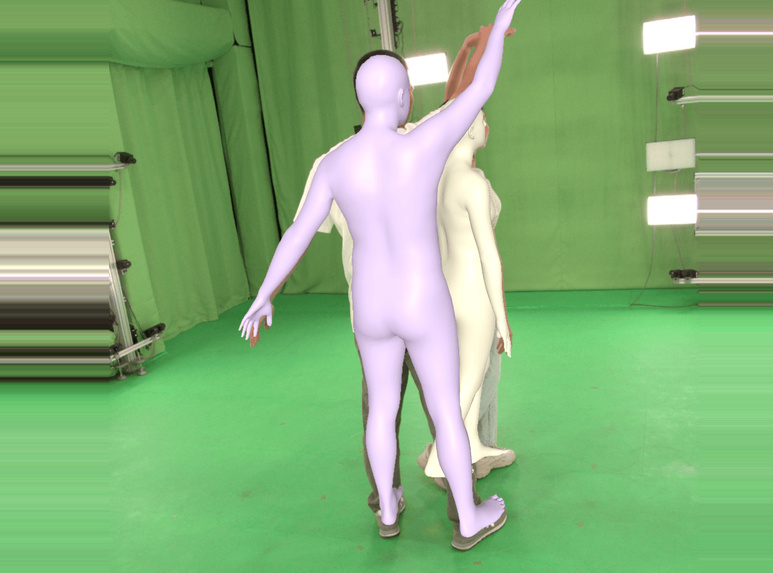}     & \includegraphics[width=\sqw]{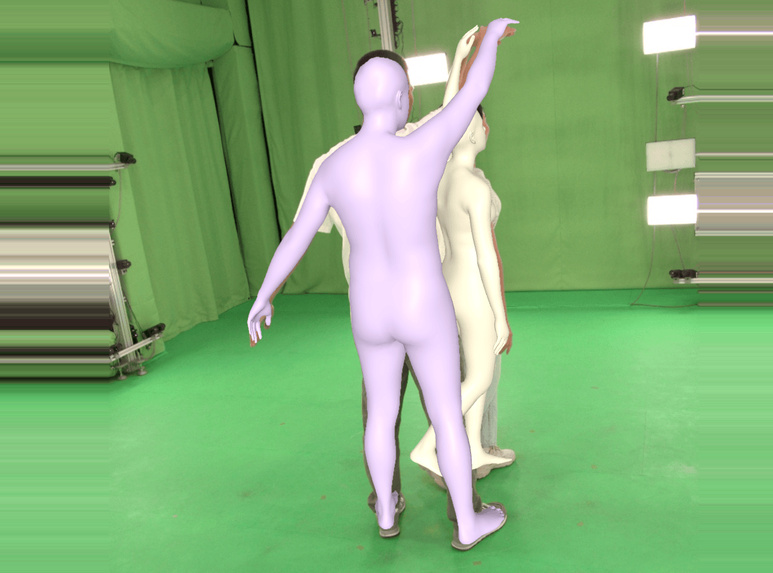}     \\
    \sqrowlabel{CHI3D}     & \includegraphics[width=\sqw]{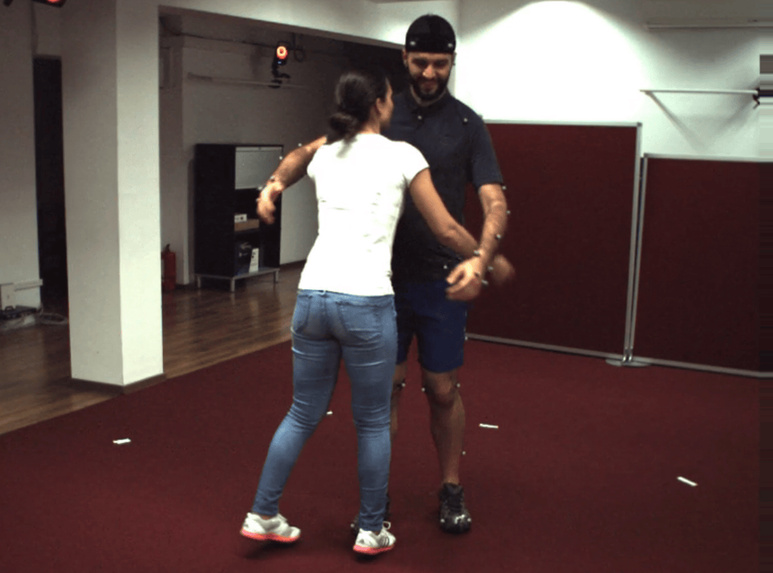}     & \includegraphics[width=\sqw]{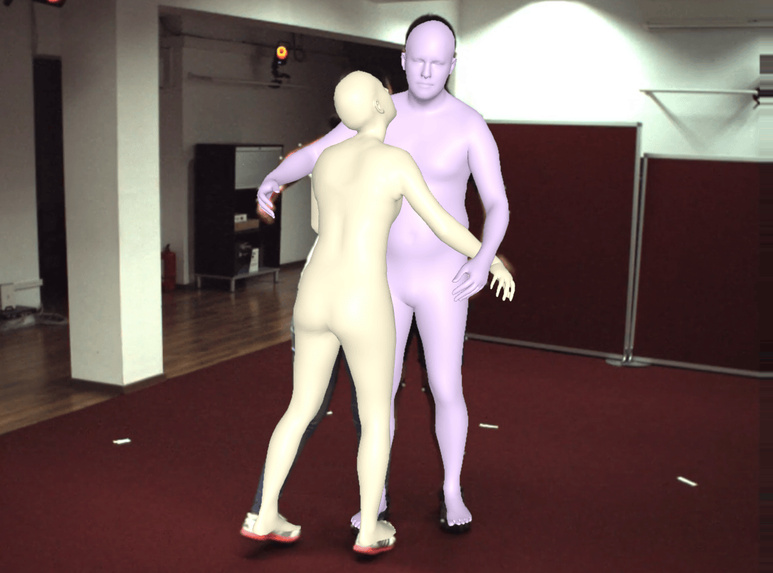}     & \sqna{0.066}            & \includegraphics[width=\sqw]{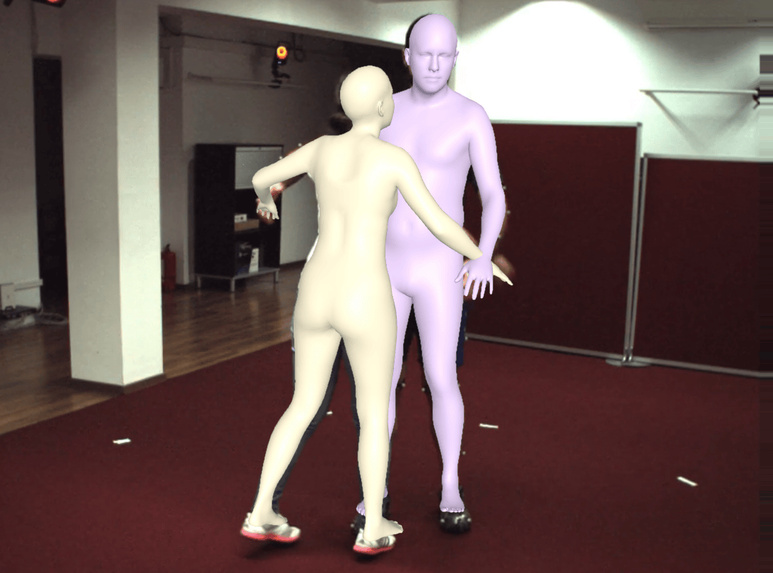}     & \includegraphics[width=\sqw]{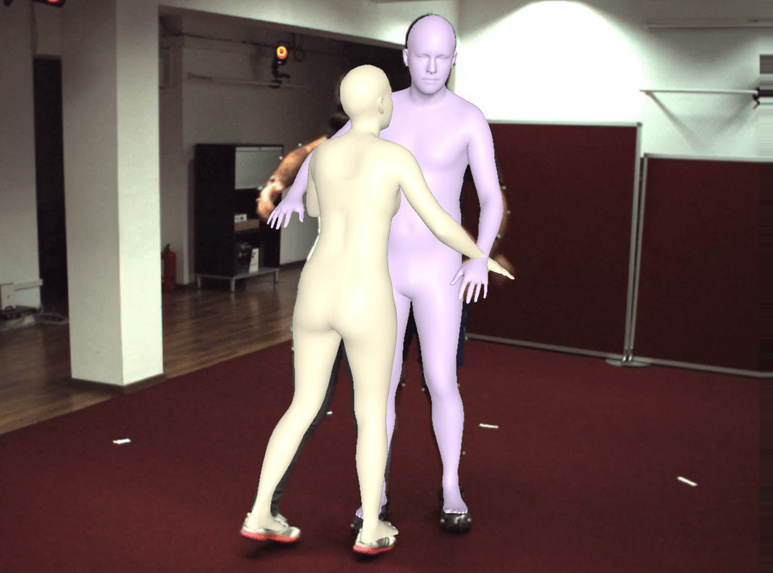}     \\
    \sqrowlabel{CHI3D}     & \includegraphics[width=\sqw]{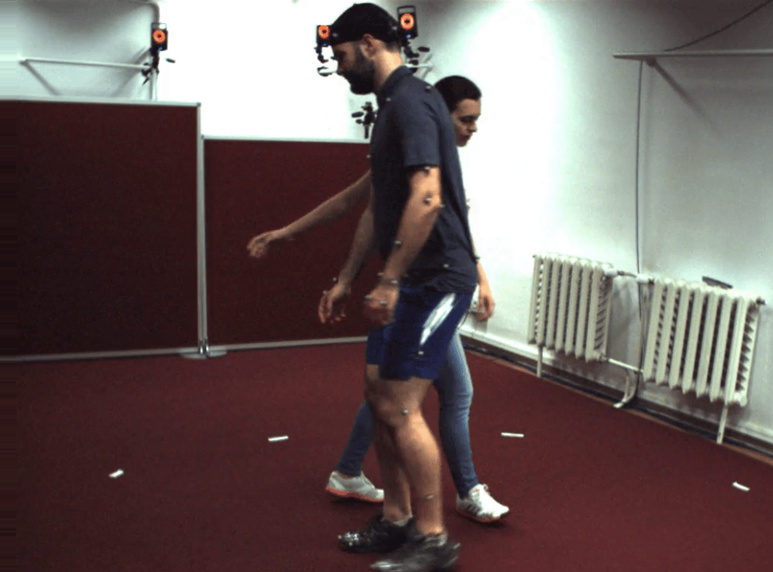}    & \includegraphics[width=\sqw]{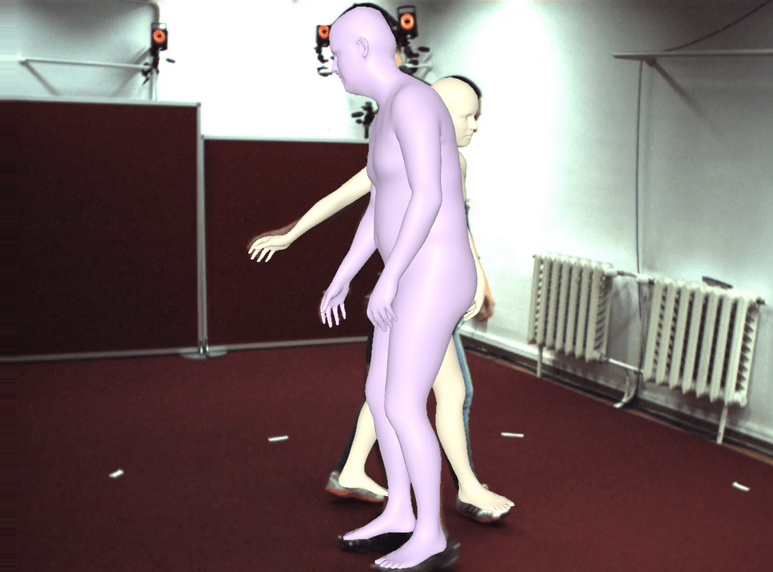}    & \sqna{0.066}            & \includegraphics[width=\sqw]{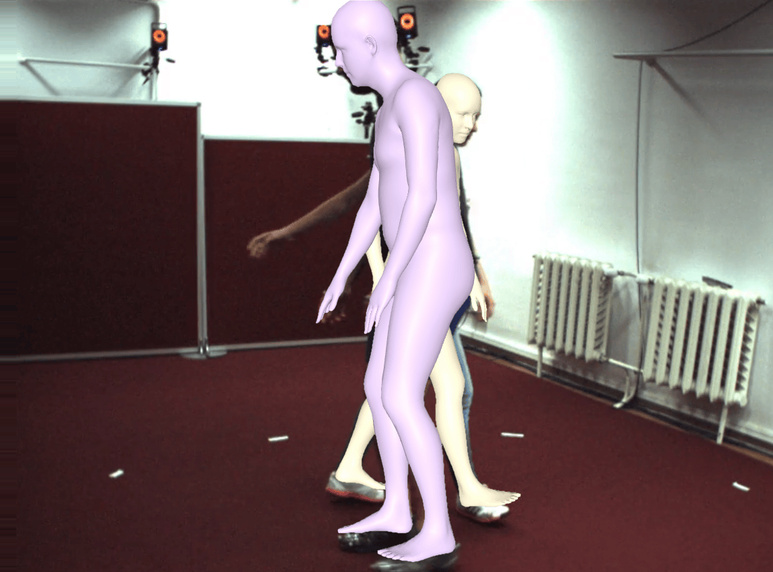}    & \includegraphics[width=\sqw]{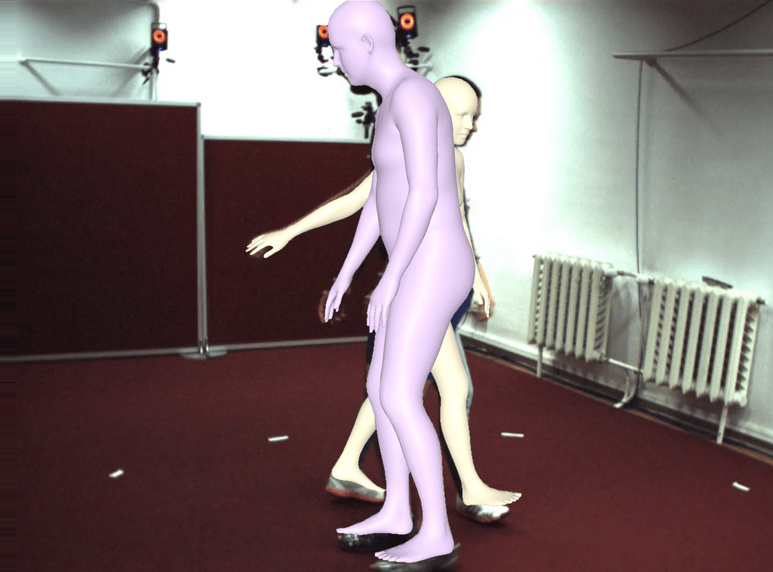}    \\
    \sqrowlabel{Harmony4D} & \includegraphics[width=\sqw]{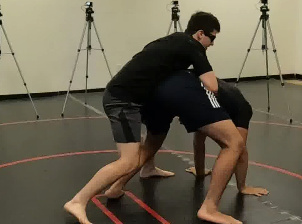} & \includegraphics[width=\sqw]{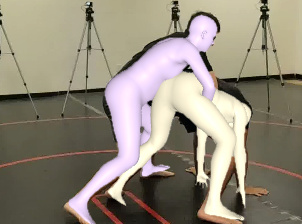} & \includegraphics[width=\sqw]{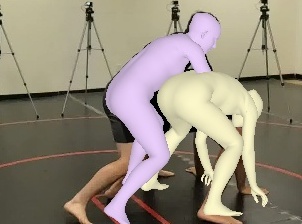} & \includegraphics[width=\sqw]{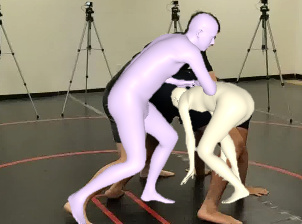} & \includegraphics[width=\sqw]{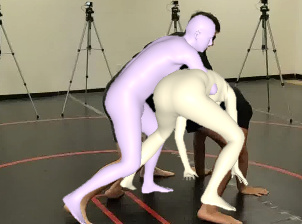} \\
    \sqrowlabel{ExPI}      & \includegraphics[width=\sqw]{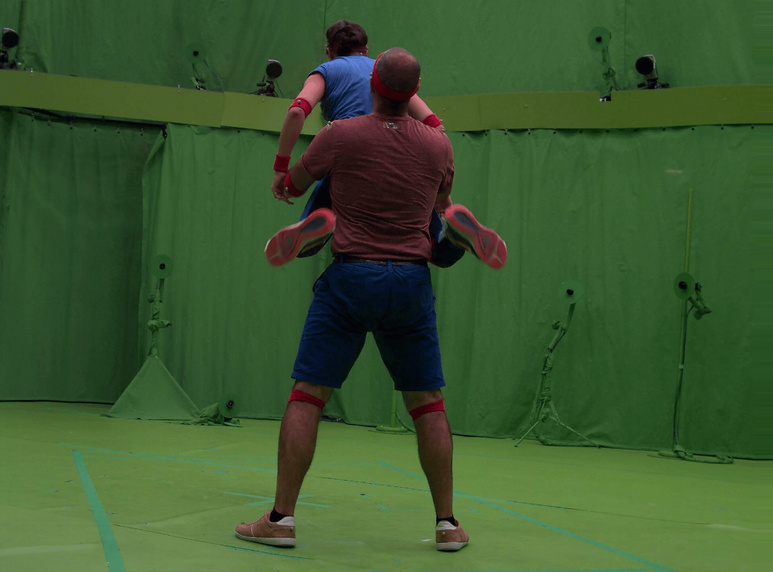}      & \sqna{0.066}          & \includegraphics[width=\sqw]{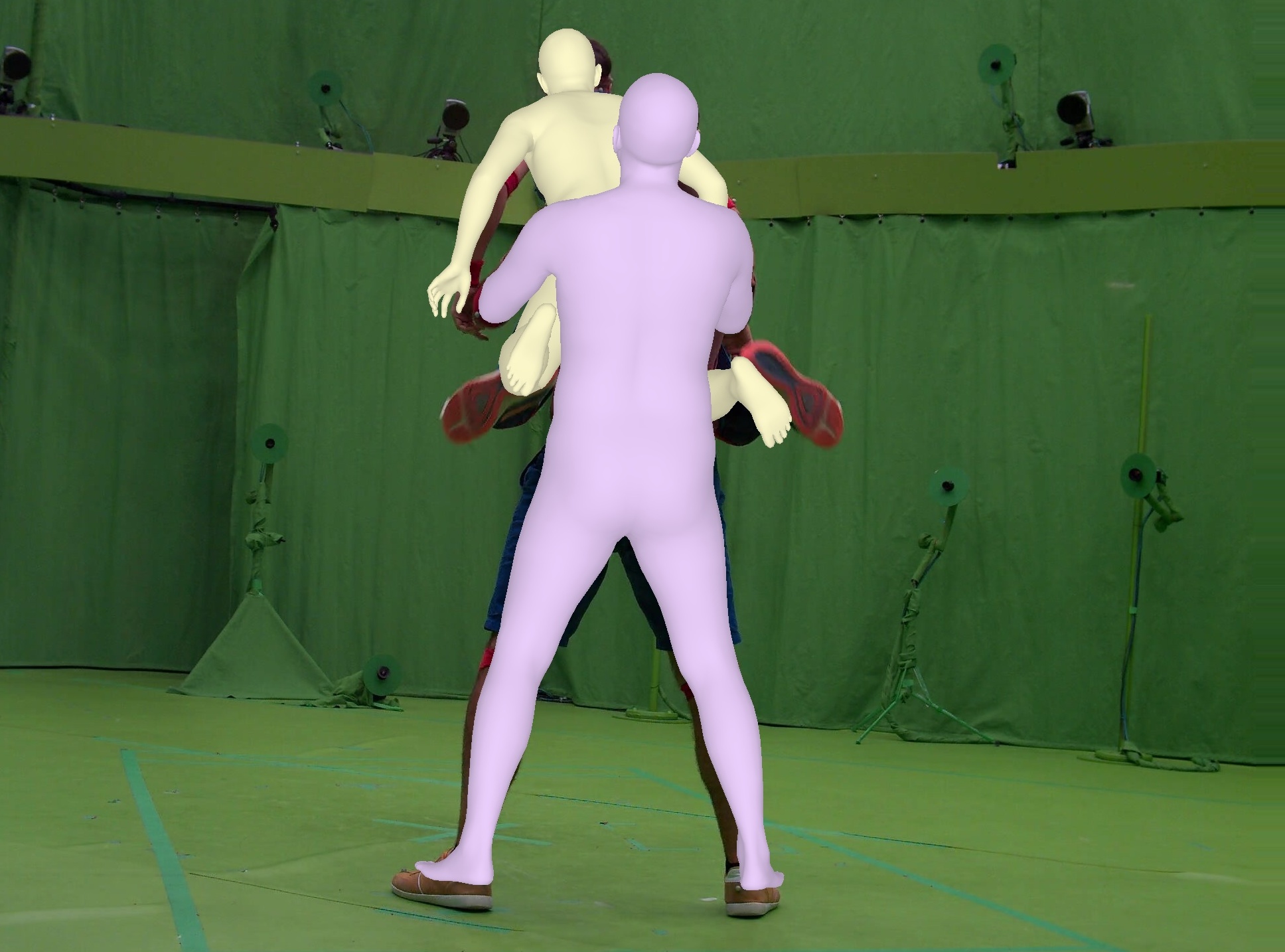}      & \includegraphics[width=\sqw]{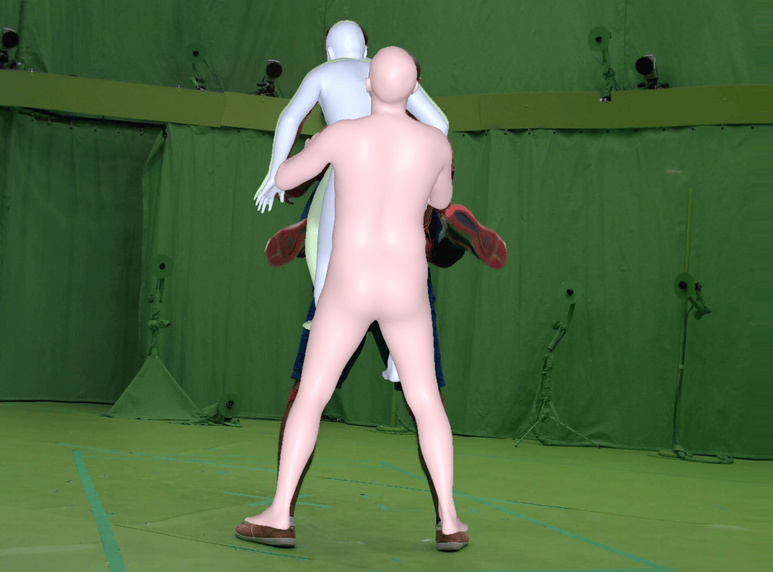}      & \includegraphics[width=\sqw]{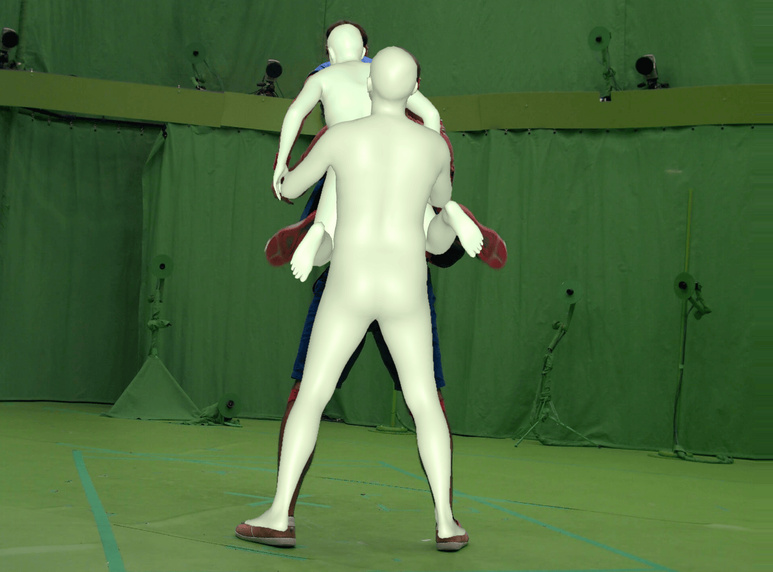}      \\
    \sqrowlabel{ExPI}      & \includegraphics[width=\sqw]{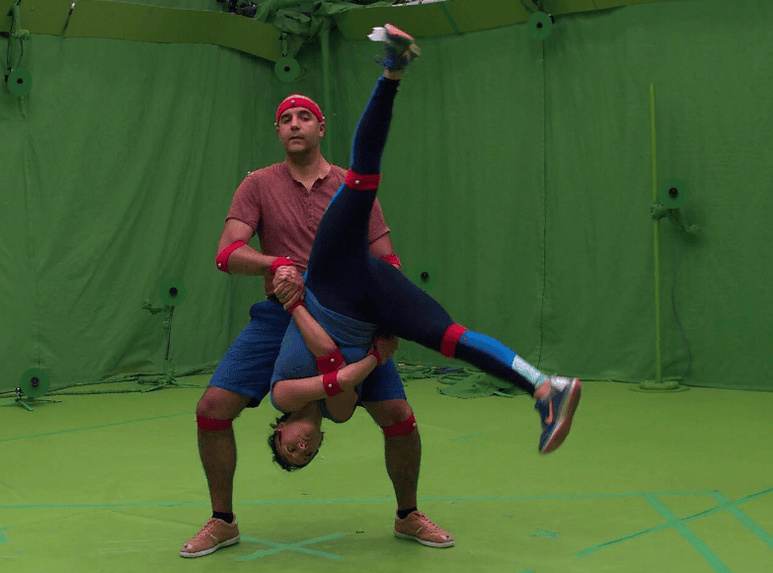}     & \sqna{0.066}          & \includegraphics[width=\sqw]{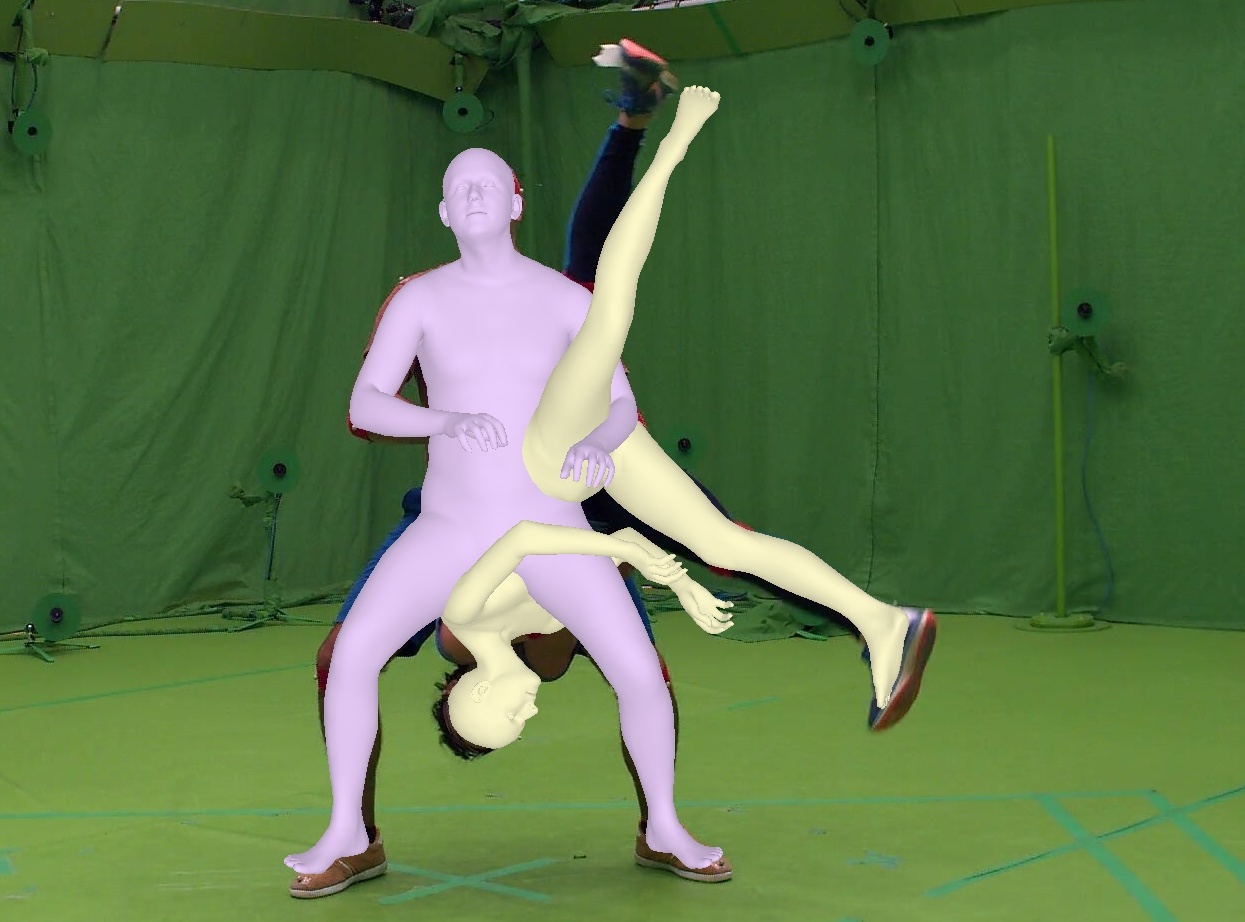}     & \includegraphics[width=\sqw]{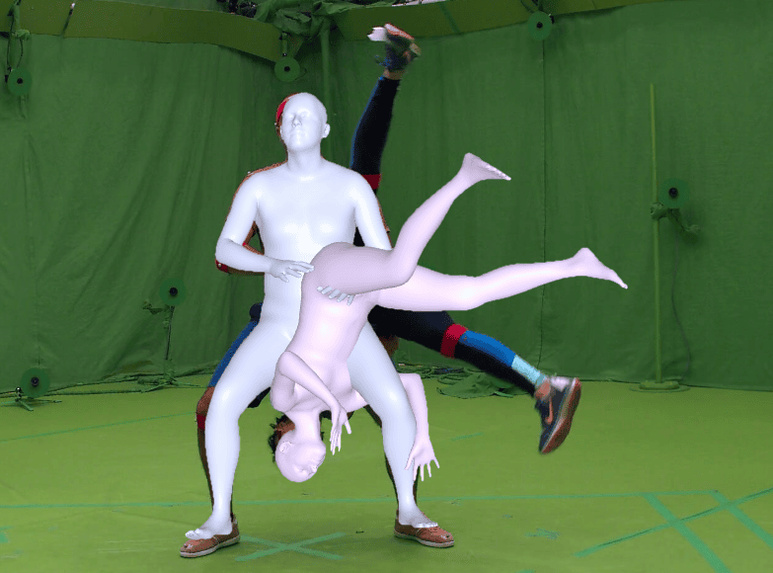}     & \includegraphics[width=\sqw]{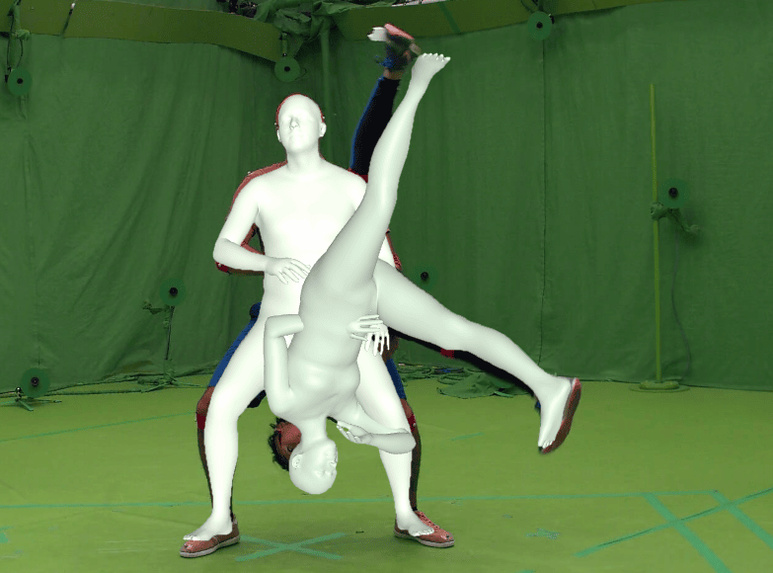}     \\
    \sqrowlabel{ITW}       & \includegraphics[width=\sqw]{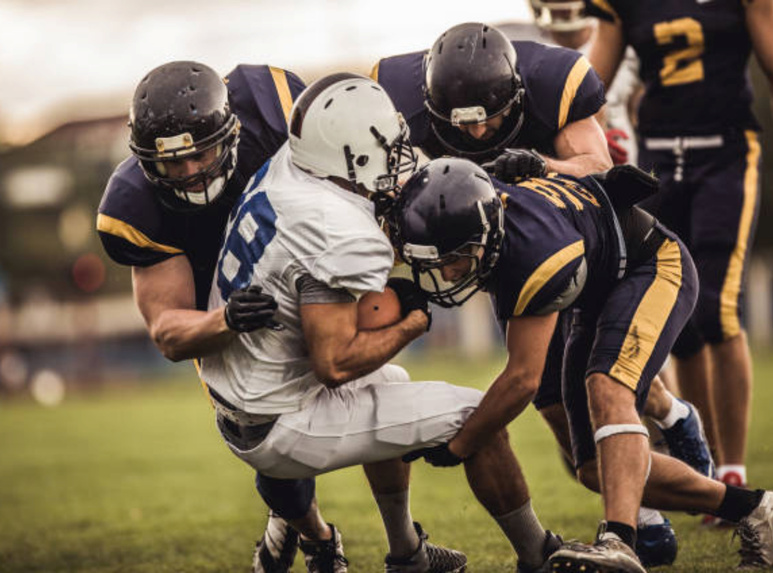}       & \sqna{0.066}          & \includegraphics[width=\sqw]{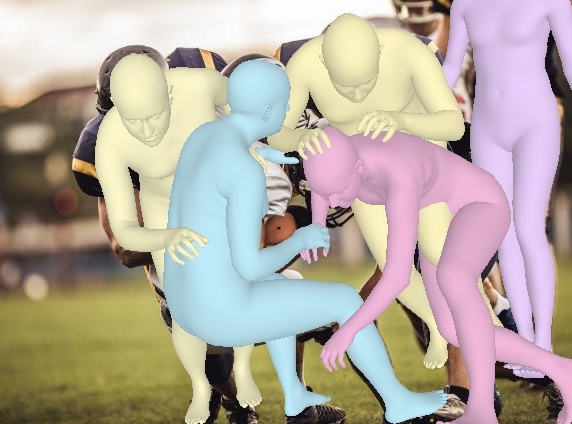}       & \includegraphics[width=\sqw]{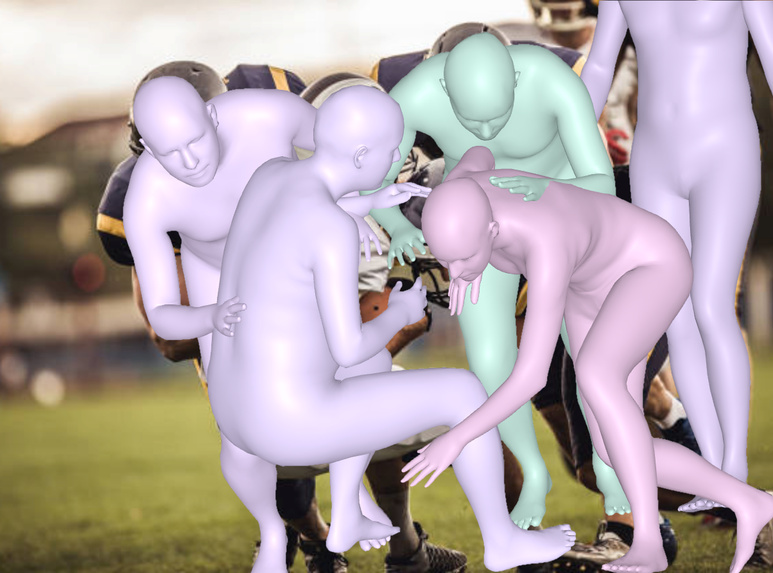}       & \includegraphics[width=\sqw]{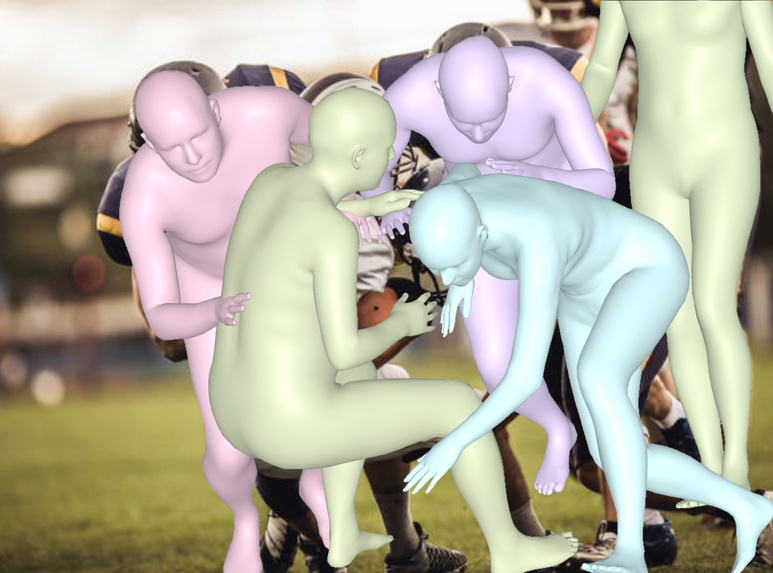}       \\
  \end{tabular}
  \caption{%
    \textbf{Additional qualitative comparisons.}
    Top five rows (Hi4D, CHI3D, Harmony4D) have ground-truth meshes; the bottom three rows (ExPI and an in-the-wild image, ITW) have no 3D ground truth.
    PromptHMR is shown only on Harmony4D, ExPI and ITW, because its publicly released checkpoint is trained on Hi4D and CHI3D, making these the only held-out settings.
  }
  \label{fig:supp_qualitative}
\end{figure*}

\begin{figure*}[t]
  \centering
  \includegraphics[width=0.96\linewidth]{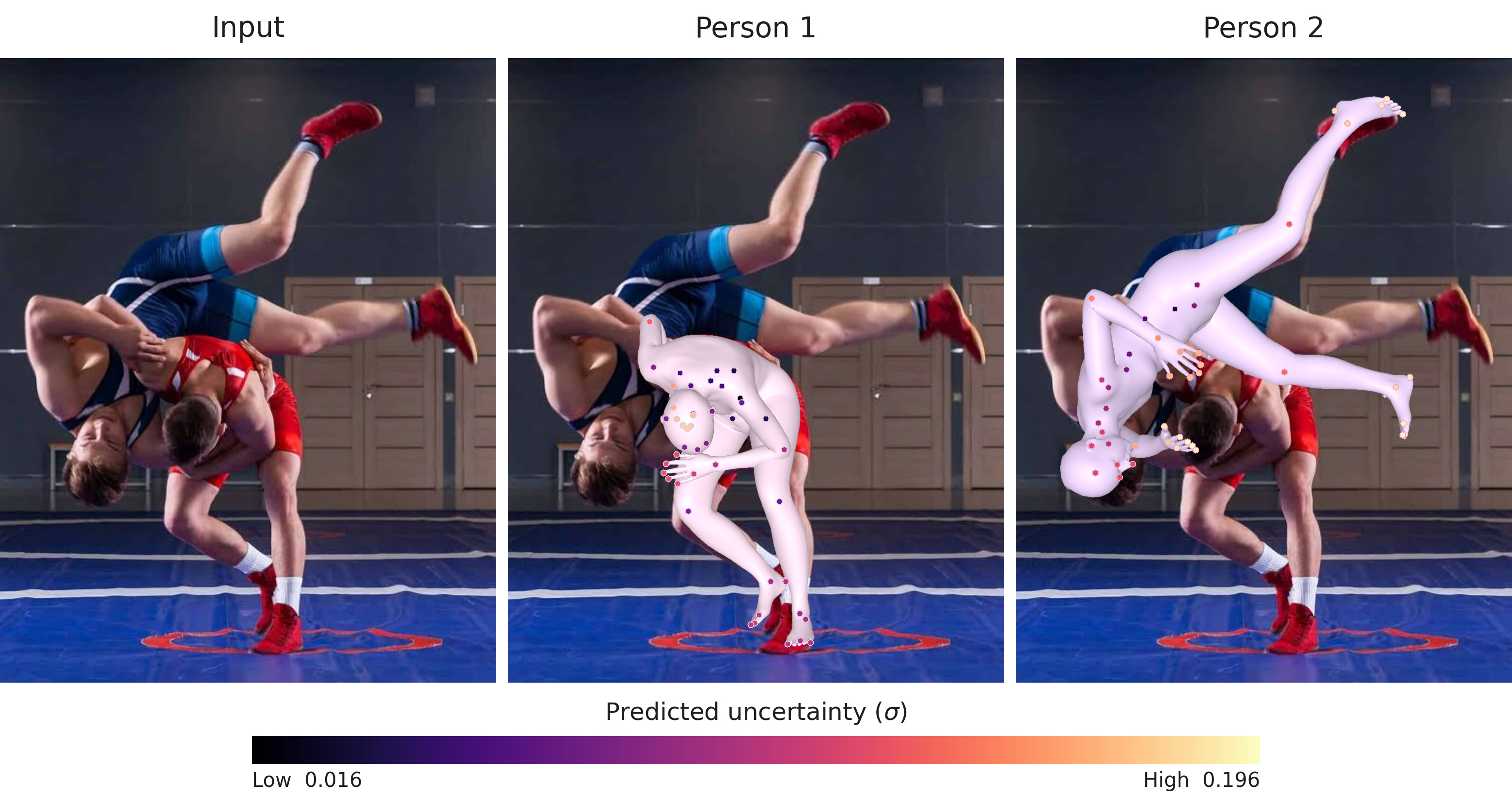}
  \caption{%
    \textbf{Predicted per-joint uncertainty on an in-the-wild image.}
    \emph{Left}: input. \emph{Middle, right}: the \cisyn{}-HMR mesh for each of the two interacting people, with the 45 output joints projected onto the image and colored by the predicted standard deviation $\exp(\sigma_j)$ (shared color scale, brighter is more uncertain).
    Uncertainty concentrates on the joints that are occluded by the other person or that lie on the least accurately reconstructed limbs, while the visible torso and pelvis receive low uncertainty.
  }
  \label{fig:supp_uncertainty}
\end{figure*}

The metrics of Table~\ref{tab:main_results} are root-relative and averaged over matched people.
As Table~\ref{tab:recall} shows, every model detects over 97\% of the ground-truth people on each benchmark, so these averages cover nearly all people and the few missed ones would change them only slightly.
Adding \cisyn{} also raises recall for both models on all three benchmarks.

Table~\ref{tab:pair} evaluates the two people jointly.
Adding \cisyn{} reduces the inter-person distance error for both models on all three benchmarks.
On Hi4D and CHI3D, all models penetrate more deeply than the ground truth, as no term in the training objective penalizes inter-person penetration (Sec.~\ref{app:failure}).
Since \cisyn{} consists of close interactions, adding it also brings the predicted people closer together in most cases, so penetration becomes more frequent, approaching the ground-truth frequency, and slightly deeper.

\begin{table}[htbp]
  \caption{%
    \textbf{Detection recall.}
    Recall (\%) and number of missed ground-truth people for the controlled models of Table~\ref{tab:main_results}.
  }
  \label{tab:recall}
  \centering
  \small
  \setlength{\tabcolsep}{3pt}
  \begin{tabular}{lcccccc}
    \toprule
    \multirow{2}{*}{Method} & \multicolumn{2}{c}{Hi4D} & \multicolumn{2}{c}{CHI3D} & \multicolumn{2}{c}{Harmony4D} \\
    \cmidrule(lr){2-3} \cmidrule(lr){4-5} \cmidrule(lr){6-7}
    & Rec.$\uparrow$ & Miss$\downarrow$ & Rec.$\uparrow$ & Miss$\downarrow$ & Rec.$\uparrow$ & Miss$\downarrow$ \\
    \midrule
    \multicolumn{7}{l}{\emph{Base}} \\
    MA-HMR       & 99.18 & 242 & 98.92 & 349 & 97.47 & 912 \\
    \cisyn{}-HMR & 99.35 & 191 & 98.98 & 330 & 98.11 & 680 \\
    \midrule
    \multicolumn{7}{l}{\emph{Base+\cisyn{}}} \\
    MA-HMR       & \textbf{99.98} & 7 & \textbf{99.29} & 231 & 98.00 & 718 \\
    \cisyn{}-HMR & \textbf{99.98} & 5 & 99.25 & 242 & \textbf{98.17} & 658 \\
    \bottomrule
  \end{tabular}
\end{table}

\begin{table*}[t]
  \caption{%
    \textbf{Inter-person accuracy.}
    \emph{Dist.}: image-plane error of the pelvis-to-pelvis vector (mm).
    \emph{Pen.}: mean penetration depth (mm) and share of frames with penetration (\%), following Table~\ref{tab:penetration}; closer to the ground truth is better.
    Evaluated on frames where all models detect both people.
  }
  \label{tab:pair}
  \centering
  \small
  \setlength{\tabcolsep}{4pt}
  \begin{tabular}{lccccccccc}
    \toprule
    \multirow{2}{*}{Method} & \multicolumn{3}{c}{Hi4D} & \multicolumn{3}{c}{CHI3D} & \multicolumn{3}{c}{Harmony4D} \\
    \cmidrule(lr){2-4} \cmidrule(lr){5-7} \cmidrule(lr){8-10}
    & Dist.$\downarrow$ & Pen.\ (mm) & Pen.\ (\%) & Dist.$\downarrow$ & Pen.\ (mm) & Pen.\ (\%) & Dist.$\downarrow$ & Pen.\ (mm) & Pen.\ (\%) \\
    \midrule
    Ground truth                   & --    & 4.9  & 56.6 & --    & 2.6 & 27.3 & --    & 3.6 & 33.1 \\
    \midrule
    MA-HMR (Base)                  & 29.7  & 11.0 & 49.8 & 38.4  & 3.2 & 21.2 & 87.2  & 2.7 & 18.2 \\
    \cisyn{}-HMR (Base)            & 27.5  & 10.1 & 45.9 & 38.8  & 2.8 & 18.9 & 78.9  & 2.4 & 16.2 \\
    \midrule
    MA-HMR (Base+\cisyn{})         & 24.9  & 11.6 & 53.3 & \textbf{37.4} & 3.3 & 22.0 & 86.7  & 3.1 & 21.4 \\
    \cisyn{}-HMR (Base+\cisyn{})   & \textbf{24.0} & 10.9 & 50.3 & 37.7  & 2.7 & 18.5 & \textbf{71.8} & 2.6 & 17.6 \\
    \bottomrule
  \end{tabular}
\end{table*}

\subsection{General-Domain Evaluation on 3DPW}
\label{app:results:3dpw}

To check that \cisyn{} does not harm general HMR, we evaluate the controlled models of Table~\ref{tab:main_results}, trained from the released MA-HMR~\cite{wang2025mahmr} checkpoint on Base and on Base+\cisyn{} under an identical schedule, on the single-person, in-the-wild 3DPW~\cite{von20183dpw} benchmark.
As shown in Table~\ref{tab:3dpw}, adding \cisyn{} does not degrade general-domain performance for either model.

\begin{table}[htbp]
  \caption{%
    \textbf{General-domain evaluation on 3DPW.}
    Within each block, both rows continue training the released MA-HMR~\cite{wang2025mahmr} checkpoint under an identical schedule and differ only in whether \cisyn{} is added to Base (Sec.~\ref{app:exp}), evaluated on 3DPW~\cite{von20183dpw}; all metrics in mm.
  }
  \label{tab:3dpw}
  \centering
  \small
  \setlength{\tabcolsep}{3.5pt}
  \begin{tabular}{lccc}
    \toprule
    Training & MPJPE$\downarrow$ & PA-MPJPE$\downarrow$ & PVE$\downarrow$ \\
    \midrule
    MA-HMR (Base)       & 62.9 & 39.9 & 73.5 \\
    MA-HMR (Base+\cisyn{})                 & \textbf{62.2} & \textbf{39.8} & \textbf{73.0} \\
    \midrule
    \cisyn{}-HMR (Base)       & \textbf{62.6} & 39.6 & \textbf{73.0} \\
    \cisyn{}-HMR (Base+\cisyn{})                 & \textbf{62.6} & \textbf{39.5} & \textbf{73.0} \\
    \bottomrule
  \end{tabular}
\end{table}

\subsection{Additional Qualitative Results}
\label{app:results:qualitative}

Figure~\ref{fig:supp_qualitative} extends the qualitative evaluation of Sec.~\ref{sec:exp:qualitative} with two further examples each for Hi4D, CHI3D and ExPI, one for Harmony4D, and an additional in-the-wild image. The Hi4D, CHI3D and Harmony4D rows include ground-truth meshes for reference, and PromptHMR is again shown only on the rows held out from its released checkpoint (Harmony4D, ExPI and the in-the-wild image); \cisyn{}-HMR consistently produces tighter contact and more accurate limb alignment than the baselines across all sequences, including the unconstrained in-the-wild inputs.

\subsection{Visualizing the Predicted Uncertainty}
\label{app:results:uncertainty}

The uncertainty head is used only during training and is discarded at inference, but inspecting what it has learned to be uncertain about is informative, since the loss never specifies \emph{which} joints should be uncertain.
Figure~\ref{fig:supp_uncertainty} visualizes the predicted $\exp(\sigma_j)$ on an in-the-wild image.
The predicted uncertainty is low on the visible torso and pelvis and high on the joints occluded by the other person and on the limbs that are reconstructed least accurately, indicating that the head does reflect where the model itself is unsure.
The occlusion mask and the uncertainty head do not aim to reconstruct heavily occluded joints more accurately. Because a single image leaves the 3D position of such joints ambiguous, both instead down-weight the supervision on these joints, so that ambiguous targets add less noise to training.

\subsection{Qualitative Results on the \cisyn{} Test Set}
\label{app:results:cisyn_qual}

\begin{figure}[ht]
  \centering
  \setlength{\tabcolsep}{2pt}
  \renewcommand{\arraystretch}{0.5}

  \begin{tabular}{@{}cc@{}}
    \footnotesize Input & \footnotesize GT \\[2pt]
    \includegraphics[width=0.4\linewidth]{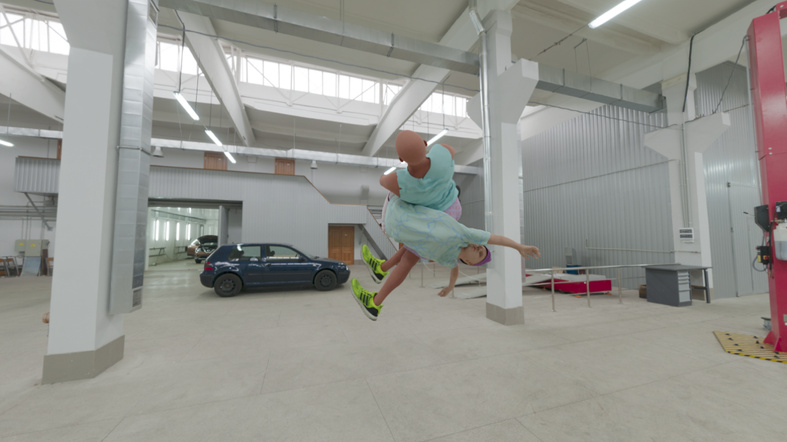} & \includegraphics[width=0.4\linewidth]{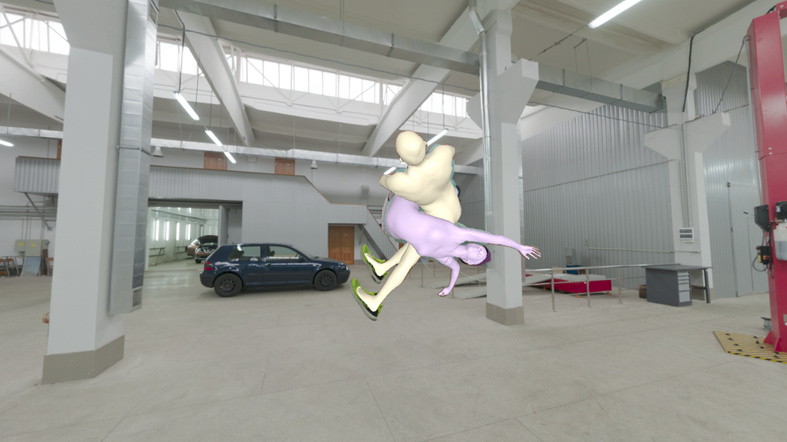} \\[6pt]
    \footnotesize Base+Hi4D & \footnotesize Base+CHI3D \\[2pt]
    \includegraphics[width=0.4\linewidth]{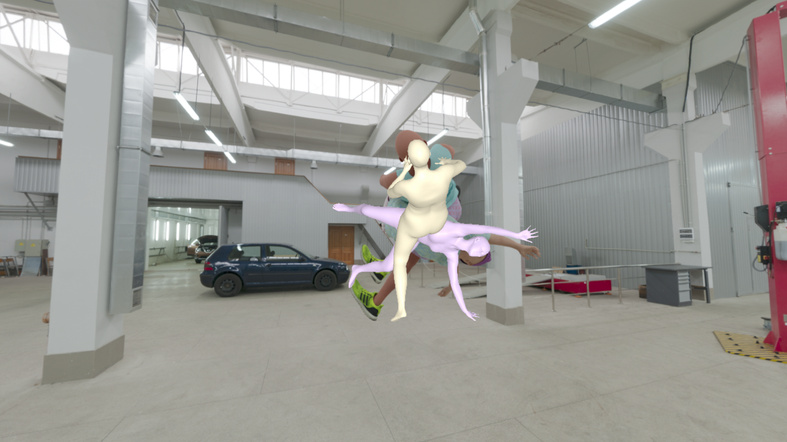} & \includegraphics[width=0.4\linewidth]{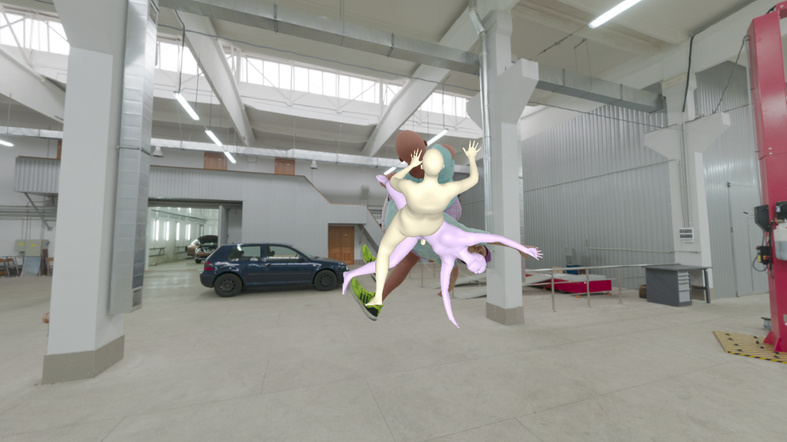} \\[6pt]
    \footnotesize MA-HMR (Base) & \footnotesize \cisyn{}-HMR (Base) \\[2pt]
    \includegraphics[width=0.4\linewidth]{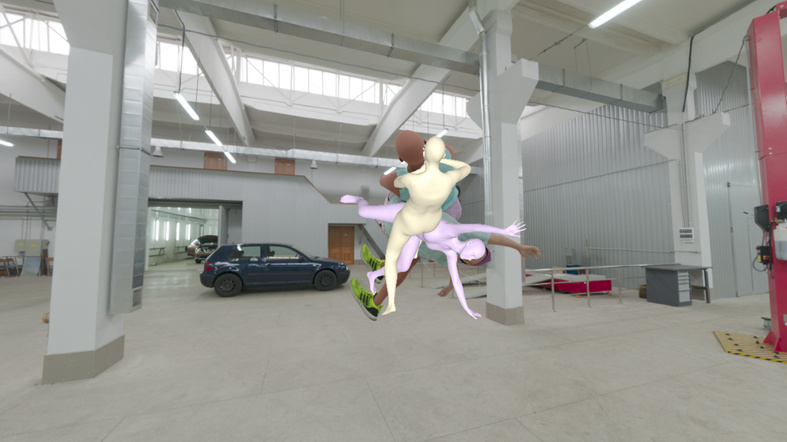} & \includegraphics[width=0.4\linewidth]{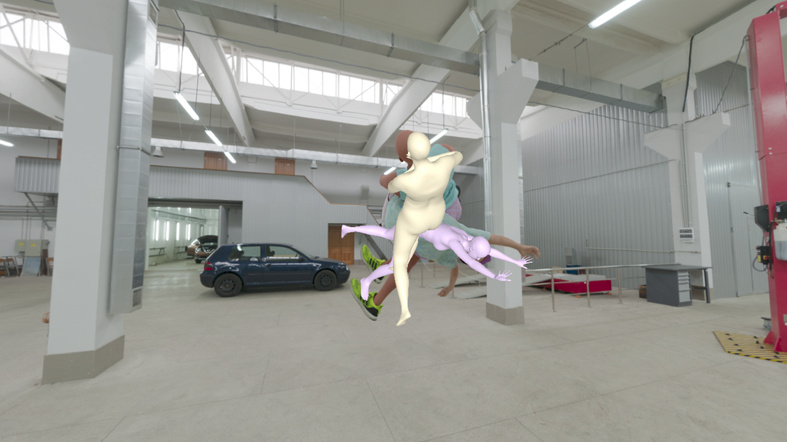} \\
  \end{tabular}

  \caption{%
    \textbf{Qualitative results on the \cisyn{} test set.}
    Model settings match Table~\ref{tab:quantitative}:
    \emph{top}~input image and ground-truth mesh;
    \emph{middle}~dataset-comparison rows from Table~\ref{tab:dataset_results};
    \emph{bottom}~settings follow Table~\ref{tab:main_results}.
  }
  \label{fig:supp_cisyn_qual}
\end{figure}

Figure~\ref{fig:supp_cisyn_qual} shows a representative example from the \cisyn{} test set under the same model settings as Table~\ref{tab:quantitative}.

\section{Limitations and Future Work}
\label{app:failure}
\begin{figure}[t]
  \centering
  \setlength{\tabcolsep}{2pt}
  \renewcommand{\arraystretch}{1.0}

  \begin{tabular}{@{}cccc@{}}
    \footnotesize Input & \footnotesize \cisyn{}-HMR Prediction\\[2pt]
    \includegraphics[width=0.47\linewidth]{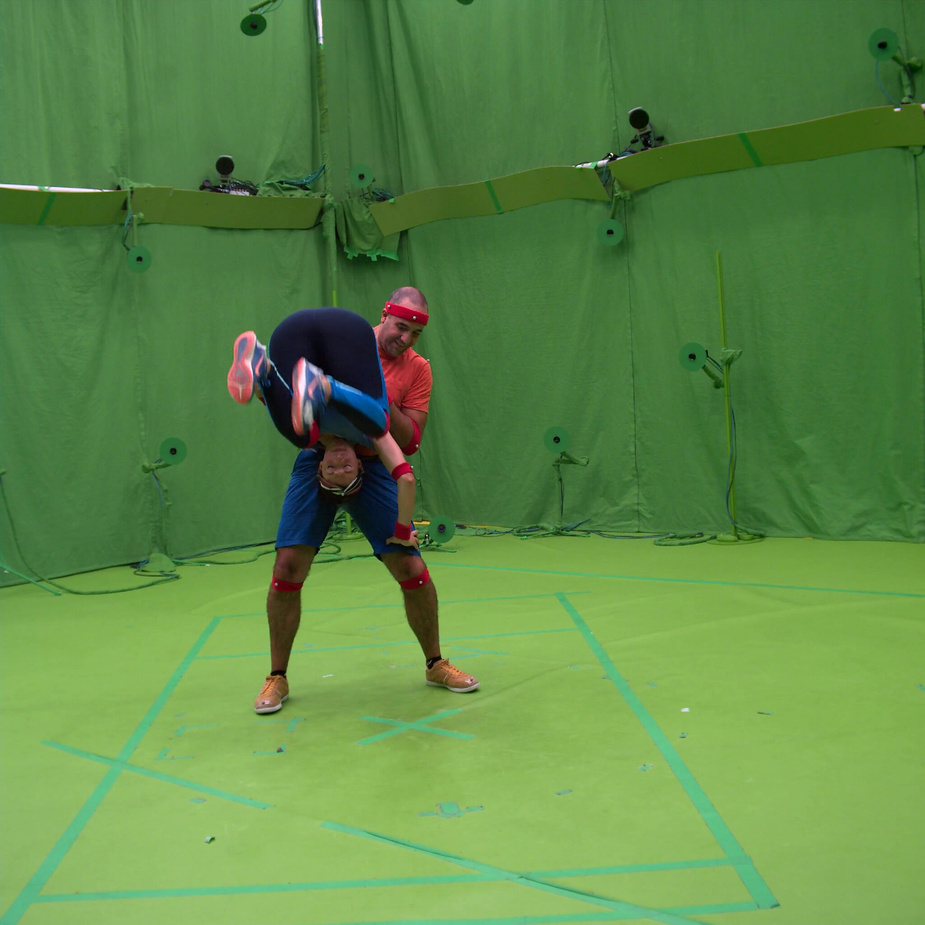} & \includegraphics[width=0.47\linewidth]{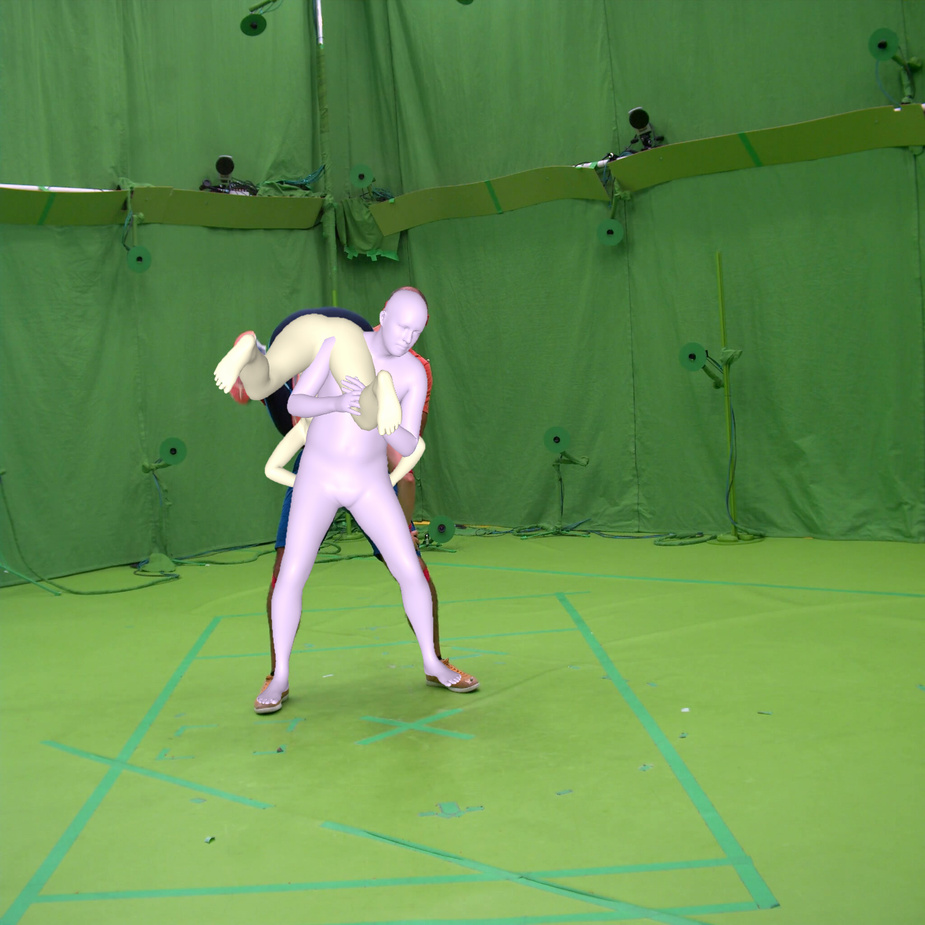} \\
    \includegraphics[width=0.47\linewidth]{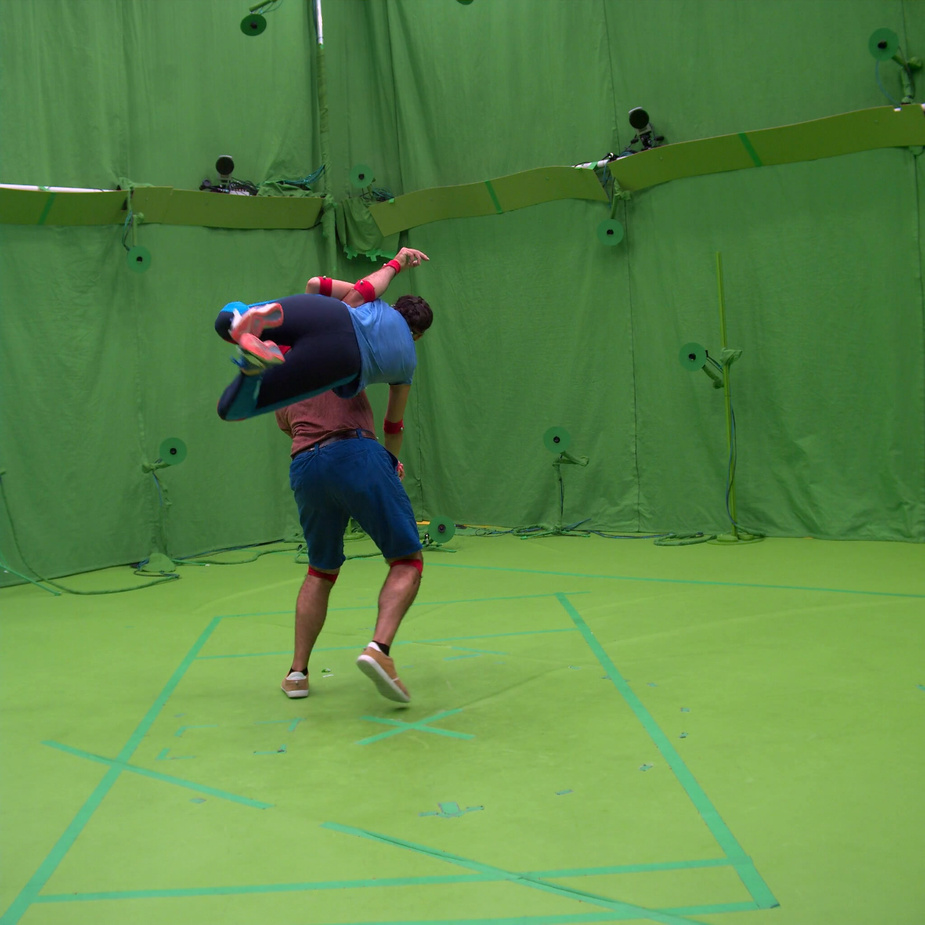} & \includegraphics[width=0.47\linewidth]{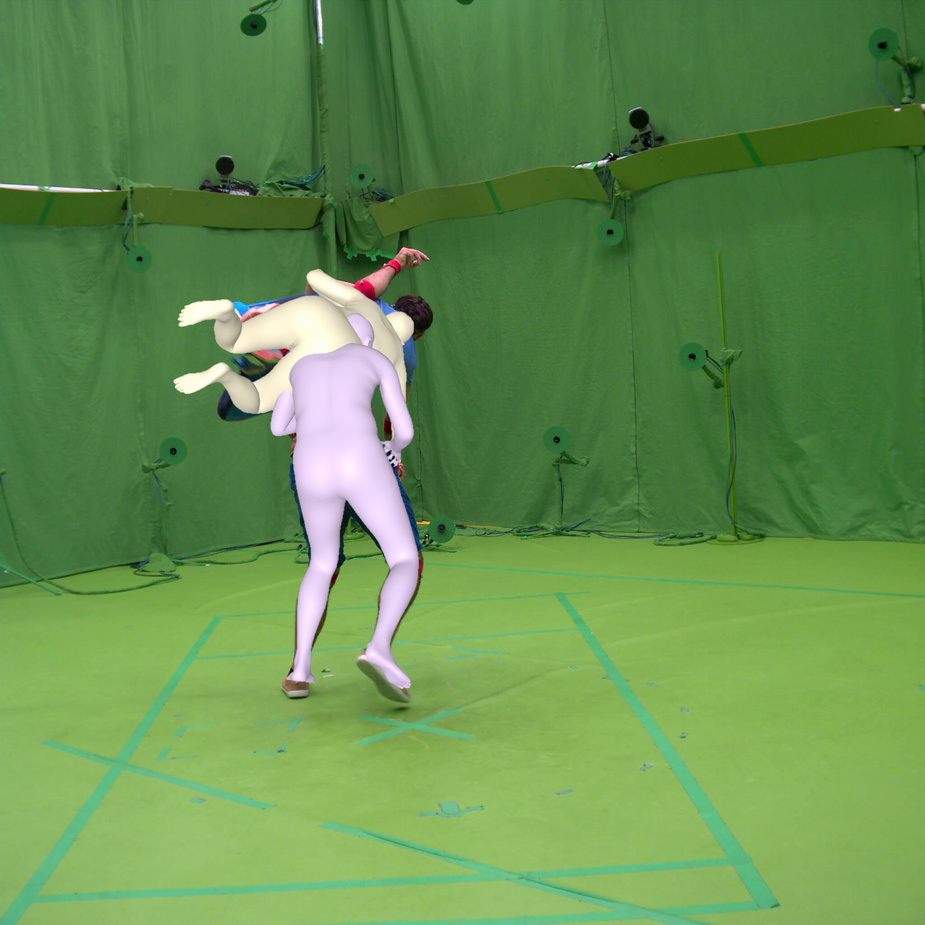} \\
  \end{tabular}

  \caption{%
    \textbf{Failure cases on ExPI.}
    Predictions from \cisyn{}-HMR on two acrobatic ExPI sequences.
  }
  \label{fig:supp_failure}
\end{figure}

\paragraph{Limitations.}
Figure~\ref{fig:supp_failure} shows failure cases on ExPI, where acrobatic interactions produce mesh interpenetration in the predictions.
Note that this is not inherited from the training data: the simulator resolves collisions at generation time, and Sec.~\ref{app:dataset_analysis:penetration} shows that \cisyn{} has a lower penetration frequency and depth than the captured close-interaction datasets.
The limitation is on the model side, where nothing enforces the same property: no term in the training objective penalizes overlap between the two predicted meshes, and no post-processing step resolves it at inference.
We experimented with an interpenetration loss but it did not improve results in practice.

\paragraph{Future work.}
Beyond the dataset and model presented here, the generation pipeline supports several extensions that we have not yet exploited and that the community can build on directly:
(1)~\textbf{Generator-in-the-loop training}: the generator is efficient and trivially parallel (Sec.~\ref{app:pipeline:cost}), so hard samples identified at training time can be regenerated on-the-fly to close the distribution gap without manual annotation;
(2)~\textbf{Action-involved simulation}: the current actors are driven by collision dynamics alone, and adding motion controllers or action priors on top of the same simulator would extend the same generator to goal-directed actions, widening the range of contact scenarios it covers;
(3)~\textbf{Scaling up}: nothing in the pipeline is specific to the released sequence count or to two interactors, and Figure~\ref{fig:supp_datasample} already includes three- and four-person samples generated with minimal modification;
(4)~\textbf{Contact annotations}: the simulator natively produces contact normals and impulse data, which we currently discard but which could be exported as ground-truth labels to benefit contact prediction tasks.

\end{document}